\documentclass[
    paper = a4,
    fontsize = 12pt,
    headinclude = true,
    open = right,
    twoside = true,
    BCOR = 10mm,
    toc = listofnumbered,
    toc = bibnumbered,
    numbers = noendperiod
]{scrreprt}

\usepackage{natbib}
\usepackage[utf8]{inputenc}
\usepackage[final]{pdfpages}
\usepackage{pdflscape}
\usepackage{listings}
\usepackage{xcolor}
\usepackage[english]{babel}
\usepackage{setspace}
\usepackage[hidelinks]{hyperref}
\usepackage[acronym]{glossaries}
\usepackage[german, algochapter, linesnumbered]{algorithm2e}
\usepackage{pgfplots}
\usepackage{tikz}
\usetikzlibrary{positioning}
\usetikzlibrary{shapes.geometric, arrows}
\usetikzlibrary{datavisualization.formats.functions}
\usetikzlibrary{decorations.pathreplacing,calligraphy}
\usepgfplotslibrary{external}
\pgfplotsset{compat=1.18} 

\usepackage{subcaption}
\usepackage{wrapfig}
\usepackage{adjustbox}
\usepackage{aligned-overset}
\usepackage{amsfonts}
\usepackage{amsmath}
\usepackage{amssymb}
\usepackage{amsthm}
\usepackage{colonequals}
\usepackage{comment}
\usepackage{csquotes}
\usepackage{graphicx}
\usepackage{mathtools}

\newtheorem{definition}{Definition}
\newtheorem{theorem}{Theorem}
\newtheorem{lemma}{Lemma}

\def\renderpapers{1}

\newacronym{nic}{NIC}{Network Interface Controller}

\SetAlCapSty{}
\SetNlSty{footnotesize}{}{}

\definecolor{lst_comments}{rgb}{0, 0.75, 0}
\definecolor{lst_linenumbers}{rgb}{0.2, 0.2, 0.2}
\definecolor{lst_keywords}{rgb}{0, 0, 0.75}
\definecolor{lst_strings}{rgb}{0.75, 0, 0}
\definecolor{lst_background}{rgb}{0.97, 0.97, 0.97}

\lstdefinestyle{msqc_style}{
    backgroundcolor=\color{lst_background},   
    commentstyle=\color{lst_comments},
    keywordstyle=\color{lst_keywords},
    stringstyle=\color{lst_strings},
    numberstyle=\tiny\color{lst_linenumbers},
    basicstyle=\footnotesize,
    xleftmargin=2em,
    breaklines = true,
    captionpos = b,
    keepspaces = true,
    numbers = left,
    numbersep = 7pt,
    showspaces = false,
    showstringspaces = false,
    showtabs = false,
    tabsize = 4
}
\begin{document}
    \pagenumbering{Roman}
    \begin{titlepage}
    \begin{center}
        \LARGE
        \textbf{A Novel Framework for Uncertainty Quantification via Proper Scores \\ for Classification and Beyond}
        \vfill
        
        \Large
        Dissertation \\
zur Erlangung des Doktorgrades \\
der Naturwissenschaften\\

        \vfill
        \Large
vorgelegt beim Fachbereich Informatik und Mathematik\\
der Johann Wolfgang Goethe-Universität\\
in Frankfurt am Main \\
            
        \vspace{0.8cm}
        von \\
        Sebastian G. Gruber \\
        aus Landshut \\
        \vfill
        Frankfurt am Main 2024\\
        (D 30)
    \end{center}

    \Large

    \newpage

    \vspace{1.75cm}
    \begin{tabbing}
    Vom Fachbereich Informatik und Mathematik \\
    der Johann Wolfgang Goethe-Universität \\ als Dissertation angenommen. \\
    \end{tabbing}

    \vspace*{8cm}
    \Large
    
    \begin{tabbing}
    
    \hspace*{0.5cm}\= \hspace*{6cm} \= \kill
    \textbf{Dekan:}\\
        Prof. Dr. Bastian von Harrach-Sammet \\
    \hspace*{0.5cm}\= \hspace*{6cm} \= \kill
    \textbf{Gutachter:}\\
        Prof. Dr. Florian B\"uttner \\
        Prof. Dr. Matthias Kaschube \\
        Prof. Dr. David R\"ugamer \\
    \hspace*{0.5cm}\= \hspace*{6cm} \= \kill
    \textbf{Datum der Disputation:} 17.07.2025 \\
    \end{tabbing}
\end{titlepage}
\cleardoublepage



\begin{minipage}[0.49\textheight]{0.95\textwidth}
    \chapter*{Abstract}
    In this PhD thesis, we propose a novel framework for uncertainty quantification
in machine learning, which is based on proper scores.
Uncertainty quantification is an important cornerstone for trustworthy and reliable machine learning applications in practice.
Usually, approaches to uncertainty quantification are problem-specific, and solutions and insights cannot be readily transferred from one task to another.
Proper scores are loss functions minimized by predicting the target distribution.
Due to their very general definition, proper scores apply to regression, classification, or even generative modeling tasks.
We contribute several theoretical results, that connect epistemic uncertainty, aleatoric uncertainty, and model calibration with proper scores, resulting in a general and widely applicable framework.
We achieve this by introducing a general bias-variance decomposition for strictly proper scores via functional Bregman divergences.
Specifically, we use the kernel score, a kernel-based proper score, for evaluating sample-based generative models in various domains, like image, audio, and natural language generation.
This includes a novel approach for uncertainty estimation of large language models, which outperforms state-of-the-art baselines.
Further, we generalize the calibration-sharpness decomposition beyond classification, which motivates the definition of proper calibration errors.
We then introduce a novel estimator for proper calibration errors in classification, and a novel risk-based approach to compare different estimators for squared calibration errors.
Last, we offer a decomposition of the kernel spherical score, another kernel-based proper score, allowing a more fine-grained and interpretable evaluation of generative image models.
Our theoretical contributions are supported via empirical evaluations of modern neural networks to highlight the relevance of our framework in practice.
\end{minipage}
    
    \tableofcontents
    
    \cleardoublepage
    \pagenumbering{arabic}
    
    
    \addcontentsline{toc}{chapter}{Deutsche Zusammenfassung - German Summary}
    \chapter*{Deutsche Zusammenfassung - German Summary}

In dieser Arbeit stellen wir einen theoretischen Rahmen vor, welcher es erlaubt Größen von Unsicherheiten im maschinellen Lernen herzuleiten.
Dieser Rahmen basiert auf Proper Scores, auch bekannt als Proper Scoring Rules, welche als Verlustfunktionen für prädiktive Verteilungen verwendet werden \citep{gneitingscores}.
Sie sind vor allem für die Klassifizierung bekannt, ihre axiomatische Definition erlaubt jedoch eine einfache Erweiterung auf Nicht-Klassifizierungsaufgaben, wie Regression oder allgemeine Verteilungen, wie sie im Bild-, Audio- oder natürlichsprachlichen Bereich vorkommen.
Diese Flexibilität erlaubt ein Ableiten neuer Unsicherheitsgrö\ss en für verschiedenste Anwendungen und dient als übergreifende Theorie für Unsicherheit.
Die Forschung im Bereich des maschinellen Lernens schreitet schnell voran und erstreckt sich auf immer komplexere Vorhersagebereiche.
Ursprünglich waren Klassifizierung und Regression von grundlegendem Interesse, aber in den letzten Jahren haben sich neue, wirkungsvolle Aufgaben etabliert.
Zu den wichtigsten gehören generative Aufgaben, bei denen maschinelle Lernmodelle den Zweck haben, Daten zu generieren, die den Trainingsdaten sehr ähnlich sind.
Der Aufstieg der generativen Modelle, insbesondere der großen Sprachmodelle, geschah schnell und unvorhersehbar.
Aufgrund dieser schnellen Entwicklung blieben theoretisches Verständnis und Intuition hinter den praktischen Ergebnissen zurück, da die meisten Theorien mit Blick auf Klassifizierung und Regression als Anwendungen entwickelt wurden.
Besonders Proper Scores erfuhren in ihrer über 70-jährigen Geschichte ein großes Interesse von verschiedenen wissenschaftlichen Gemeinschaften außerhalb des Bereichs des maschinellen Lernens \citep{VERIFICATIONOFFORECASTSEXPRESSEDINTERMSOFPROBABILITY, mccarthy1956measures, winkler1969scoring, savage1971elicitation, gneitingscores}.
Auch wenn sie oft nicht in ihrer allgemeinsten Form betrachtet werden, werden aufgrund ihrer axiomatischen Natur in den meisten Theorien und Anwendungen des maschinellen Lernens spezielle Fälle von Proper Scores verwendet.
Ihre definierende Eigenschaft ist, dass sie durch die Vorhersage der Zielverteilung minimiert werden \citep{gneitingscores}.
Aufgrund dieser sehr allgemeinen Definition sind Proper Scores in der Praxis fast überall zu finden, selbst wenn sich jemand nicht bewusst ist, dass er einen Proper Score verwendet.
Beispiele für Proper Scores sind der mittlere quadratische Fehler in der Regression und der Kreuzentropie-Verlust in der Klassifizierung.
Allgemeiner gesagt, ist die Log-Likelihood jeder Exponentialfamilie auch ein Proper Score \citep{dawid2007geometry}.
Weitere prominente Beispiele sind der Continuous Ranked Probability Score (CRPS) \citep{zamo2018estimation} für Zeitreihenprognosen, wie z. B. Wettervorhersagen, und der Hyv\"arinen Score \citep{hyvarinen2005estimation}, der vor allem für Diffusionsmodelle verwendet wird \citep{ho2020denoising}.
Die maximale mittlere Diskrepanz (MMD), die für nichtparametrische Hypothesentests \citep{JMLR:v13:gretton12a} und das Training \citep{ren2021improving} und die Evaluierung \citep{binkowski2018demystifying} von Bildgeneratoren verwendet wird, ist streng genommen kein Proper Score, da sie mehrere Stichproben aus der Zielverteilung benötigt.
Sie ist jedoch bis auf eine zielabhängige Konstante äquivalent zum sogenannten Kernel Score, der ein Proper Score ist.
Genauer gesagt ist die MMD eine funktionale Bregman-Divergenz. Diese Beziehung zwischen MMD und Kernel Score kann auf eine Beziehung zwischen funktionalen Bregman-Divergenzen und Proper Scores verallgemeinert werden.
Daraus folgt, dass jede Anwendung, die die MMD oder eine andere (funktionale) Bregman-Divergenz zur Optimierung oder Bewertung von Verteilungen verwendet, implizit auch einen Proper Score verwendet, da sie das gleiche Optimierungsziel haben.\footnote{Dies gilt nur für das nicht-konvexe Argument von Bregman-Divergenzen, vgl. Abschnitt~\ref{sec:bg_breg_div}.}

Folglich reichen Proper Scores in ihrer Anwendung bereits weit über Klassifizierung und Regression hinaus, auch wenn viele theoretische Fortschritte für Proper Scores auf solche beschränkt sind \citep{Br_cker_2009, williamson2014geometry, williamson2023geometry}.
Theoretische Arbeiten, die einen allgemeineren Begriff von Proper Scores verwenden, existieren, beschränken sich aber in der Regel auf mathematische Ergebnisse, ohne empirische Auswertungen oder Erkenntnisse anzubieten, die die Relevanz und Anwendbarkeit von Proper Scores für modernes maschinelles Lernen hervorheben \citep{hendrickson1971proper, dawid2007geometry}.

In unserem täglichen Leben existieren verschiedene Formen von Unsicherheit.
Statistisch lassen sie sich in zwei Kategorien einteilen, wie das folgende Zitat von Ronald Fisher veranschaulicht \citep{hullermeier2021aleatoric}:
\begin{displayquote}
``Die Wahrscheinlichkeiten von Ereignissen nicht zu wissen und die Ereignisse als gleich wahrscheinlich zu wissen, sind zwei ganz verschiedene Wissenszustände.'' \end{displayquote} Das \emph{``Nicht-Zu-Wissen''} wird oft als epistemische Unsicherheit bezeichnet \citep{hullermeier2021aleatoric}.
Diese Art von Unsicherheit entsteht durch einen Mangel an Wissen oder Informationen über ein System oder einen Prozess.
Sie wird oft als systematische Unsicherheit bezeichnet, weil sie mit Lücken in unserem Verständnis oder Einschränkungen in unseren Modellen zusammenhängt.
Das Hauptmerkmal der epistemischen Unsicherheit ist ihre Reduzierbarkeit.
Indem wir mehr Daten sammeln, unsere Modelle verbessern oder unser Wissen verfeinern, können wir diese Art von Unsicherheit potenziell verringern.
Ein Beispiel ist die Wettervorhersage.
Unsere derzeitigen Modelle sind zwar ausgeklügelt, aber immer noch unvollkommen.
Es gibt Faktoren, die wir nicht vollständig verstehen, und Daten, auf die wir möglicherweise keinen Zugriff haben. Dies führt zu epistemischer Unsicherheit in Wettervorhersagen.
Der zweite Teil in Fishers Zitat über das \emph{``Gleich-Wahrscheinlich-Zu-Wissen''} wird typischerweise als aleatorische Unsicherheit bezeichnet \citep{hullermeier2021aleatoric}. Diese Art von Unsicherheit rührt von der inhärenten Zufälligkeit oder Stochastizität her, die in einem System oder Prozess vorhanden ist. Sie wird oft als statistische Unsicherheit bezeichnet, weil sie sich mit der Variabilität der Ergebnisse befasst, die selbst dann auftreten, wenn alle Eingaben und Bedingungen bekannt sind.
Ihr Hauptmerkmal ist ihre Irreduzierbarkeit.
Selbst mit perfektem Wissen und unendlich vielen Daten können wir die Variabilität, die mit wirklich zufälligen Ereignissen verbunden ist, nicht eliminieren.
Ein Beispiel ist das Würfeln mit einem fairen Würfel.
Auch wenn wir alle möglichen Ergebnisse (1 bis 6) und die Wahrscheinlichkeit für jedes einzelne kennen, können wir das Ergebnis eines einzelnen Wurfs nicht mit Sicherheit vorhersagen.
Diese inhärente Unvorhersehbarkeit ist die aleatorische Unsicherheit.

In Abschnitt~\ref{sec:summary_uncertainties_via_bvd} fassen wir Kapitel~\ref{ch:uncertainties_via_bvd} dieser Arbeit zusammen, in dem wir die Bias-Varianz-Zerlegung als zentrales theoretisches Werkzeug verwenden, um die Unsicherheit von Vorhersagen im maschinellen Lernen zu quantifizieren und zu verbessern.
Die Bias-Varianz-Zerlegung ist ein grundlegendes Konzept in der statistischen Lerntheorie, das hilft, den Generalisierungsfehler einer Vorhersage zu analysieren, indem die Zielvorhersage in einen Bias-Term und einen Varianz-Term zerlegt wird \citep{hastie2009elements}.
Der Bias-Term stellt die Abweichung zwischen der durchschnittlichen Vorhersage und dem erwarteten Ziel dar, während der Varianz-Term die Variabilität der Vorhersage aufgrund der Zufälligkeit in den Trainingsdaten oder Modellparametern darstellt.
Diese Zerlegung ist klassischerweise für den mittleren quadratischen Fehler oder andere Klassifizierungsfehler bekannt.
Als Teil unserer Beiträge erweitern wir dies auf die breitere Klasse der Proper Scores.
Der Hauptvorteil dieser Verallgemeinerung besteht darin, dass wir entwickelte Theorien und Algorithmen für Proper Scores für Vorhersagen von Wahrscheinlichkeitsvektoren in der Klassifizierung übertragen können zu beliebigen Verteilungen in generativen Modellen.
Darüber hinaus ist die Bias-Varianz-Zerlegung eng mit der Modellmittelung verwandt, einem Spezialfall des Ensemble-Lernens, das eine robuste Technik zur Verbesserung der Vorhersageleistung und zur Minderung von Überanpassung ist \citep{hastie2009elements}.
Ensemble-Lernen basiert auf dem Prinzip, mehrere Vorhersagen einer Dateninstanz, die oft mit verschiedenen Algorithmen konstruiert oder auf verschiedenen Teilmengen der Daten trainiert werden, zu kombinieren, um eine kollektive Vorhersage zu erhalten, die die Genauigkeit jeder einzelnen Vorhersage übertrifft.
Modellmittelung bezieht sich auf die Berechnung des Durchschnitts einer Menge von Vorhersagen, um eine Fehlerquelle der epistemischen Unsicherheit ``herauszuintegrieren'' \citep{hullermeier2021aleatoric}.
Dies kann unter bestimmten Annahmen über die Bias-Varianz-Zerlegung theoretisch bewiesen werden \citep{ueda1996generalization}.
Eine Annahme ist jedoch, dass die einzelnen Vorhersagen unkorreliert sind, was in der Praxis oft nicht zutrifft.
Diese Annahme ist für die Bias-Varianz-Kovarianz-Zerlegung, die von \citep{ueda1996generalization} für den mittleren quadratischen Fehler eingeführt wurde, nicht erforderlich, welche wir auf Kernel Scores, eine spezielle Klasse von Proper Scores, verallgemeinern.
Zusätzlich führen wir praktische Schätzverfahren und experimentelle Auswertungen für die generative Modellierung in verschiedenen Bereichen, wie Bilder, Audio und natürliche Sprache, durch.

Ein probabilistischer Klassifikator wird üblicherweise auf Trainingsdaten optimiert, um die aleatorische Unsicherheit vorherzusagen, indem er Klassenwahrscheinlichkeiten für eine Eingabe zurückgibt.
Im Allgemeinen beinhalten Klassifizierungsaufgaben die Vorhersage diskreter Klassenlabels für gegebene Instanzen \citep{bishop2006pattern}.
Da diese Modelle zunehmend in kritischen Anwendungen wie dem Gesundheitswesen \citep{HAGGENMULLER2021202}, dem autonomen Fahren \citep{feng2020deep}, der Wettervorhersage \citep{Gneiting2005WeatherFW} und der Finanzentscheidung \citep{frydman1985introducing} eingesetzt werden, ist der Bedarf an zuverlässigen und interpretierbaren Vorhersagen von entscheidender Bedeutung geworden.
Ein wichtiger Aspekt der Zuverlässigkeit von Klassifikationsmodellen ist die Kalibrierung ihrer vorhergesagten Wahrscheinlichkeiten \citep{10.2307/2346866, hekler2023test}.
Kalibrierung bezieht sich auf die Übereinstimmung zwischen vorhergesagten Wahrscheinlichkeiten und den tatsächlichen Wahrscheinlichkeiten von Ergebnissen bei einer bestimmten Vorhersage, wodurch sichergestellt wird, dass Vorhersagen in Bezug auf ihre Konfidenzwerte zuverlässig sind \citep{ANewVectorPartitionoftheProbabilityScore}. \
Trotz der Fortschritte in den Modellarchitekturen und Lernalgorithmen neigen viele moderne Klassifikatoren, wie z. B. tiefe neuronale Netze, dazu, übermäßig selbstsichere Vorhersagen zu treffen \citep{minderer2021revisiting}.
Diese Selbstüberschätzung kann auf mehrere Faktoren zurückgeführt werden, darunter die Komplexität des Modells, die Beschränkungen der Trainingsdaten und inhärente Verzerrungen in Lernprozessen \citep{guo2017calibration}.
Folglich können selbst Modelle, die eine hohe Genauigkeit erreichen, unter einer schlechten Kalibrierung leiden, was zu möglichen Fehlinterpretationen und Entscheidungen ohne angemessenes Risikobewusstsein führen kann.
Die Quantifizierung der Kalibrierung eines Modells ist ein herausforderndes Problem, das wir in unseren Beiträgen angehen.
In Abschnitt~\ref{sec:summary_uncertainty_cal_sharp} fassen wir unsere Beiträge aus Kapitel~\ref{ch:uncertainties_via_cal_sharp} zusammen, das sich um Kalibrierungsfehler, ihre Schätzer und ihre Verallgemeinerung über die Klassifizierung hinaus dreht.
Insbesondere motivieren wir die Definition von \emph{Proper Calibration Errors} basierend auf einer Verallgemeinerung der Kalibrierungs-Schärfe-Zerlegung von Proper Scores über die Klassifizierung hinaus.
Es folgt ein allgemeiner Schätzer für Proper Calibration Errors in der Klassifizierung und ein risikobasiertes Optimierungsverfahren für Schätzer von quadratischen Kalibrierungsfehlern.

Zuletzt fassen wir Kapitel~\ref{ch:disentangling_mean_embeddings} in Abschnitt~\ref{sec:summary_neurips_ws_24} zusammen, wo wir ein Verfahren zur Entflechtung der Kosinus-Ähnlichkeit zwischen mittleren Einbettungen in einem reproduzierenden Kernel-Hilbertraum vorstellen, das sich leicht auf den Kernel Spherical Score, ein Proper Score, übertragen lässt.
Dies ermöglicht eine feinkörnigere Diagnose von Fehlverhalten des Modells bei der Bilderzeugung.
Die praktische Umsetzbarkeit der theoretischen Beiträge wird durch verschiedene reale Experimente gestützt, die effektiv neue Einblicke in das Training von Bildgeneratoren geben.

Unsere \textbf{Kernbeiträge} lassen sich wie folgt zusammenfassen:
\begin{itemize}
 \item Wir bieten eine allgemeine Bias-Varianz-Zerlegung für Proper Scores an und leiten Sonderfälle von Interesse mit Anwendung auf die Unsicherheitsquantifizierung ab.
 Insbesondere wird die Bias-Varianz-Kovarianz-Zerlegung für Kernel Scores auf die generative Modellierung in einer Vielzahl von Domänen angewendet, darunter Bilder, Audio und natürliche Sprache (siehe Abschnitt~\ref{sec:summary_uncertainties_via_bvd} und Kapitel~\ref{ch:uncertainties_via_bvd}).
 \item Wir führen das Konzept der Proper Calibration Errors für zuverlässige Unsicherheiten jenseits der Klassifizierung ein, definieren einen allgemeinen Schätzer in der Klassifizierung und etablieren ein Optimierungsverfahren für Schätzer von quadratischen Kalibrierungsfehlern (siehe Abschnitt~\ref{sec:summary_uncertainty_cal_sharp} und Kapitel~\ref{ch:uncertainties_via_cal_sharp}).
 \item Wir entflechten die Kosinus-Ähnlichkeit von mittleren Einbettungen, die mit dem Kernel Spherical Score zusammenhängt, für eine feinkörnigere Bewertung von Bildgenerierungsmodellen (siehe Abschnitt~\ref{sec:summary_neurips_ws_24} und Kapitel~\ref{ch:disentangling_mean_embeddings}).
\end{itemize}

    \chapter{Introduction}
\label{ch:intro}

\section{Motivation}
\label{sec:motivation}

The scientific field of machine learning has been consistently growing in size and impact throughout the last decades, partly due to the breakthroughs in the design and training of neural networks \citep{kelleher2019deep}.
Its influence reaches well beyond academia, into industry, and our daily lives \citep{kasneci2023chatgpt}.
Due to the flexibility of machine learning methods, solutions for novel tasks, which we may also refer to as applications in this thesis, arise in a regular yet unpredictable manner.
Breakthroughs in the past often revolved around regression \citep{friedman2001greedy} and decision-making via classification \citep{krizhevsky2012imagenet}.
In recent years, a multitude of additional tasks have become viable due to improvements in generative modeling.
Prominent cases are image generation \citep{ho2020denoising}, text-to-speech audio generation \citep{kim2020glow}, and natural language generation \citep{openai2023gpt4}.
The rapid improvement in generating content, which is indistinguishable from human-generated content, brought forth wide-scale deployment of such methods for daily consumers \citep{team2023gemini}.
Even though these machine learning tasks may seem unrelated, they are often used in sensitive domains with real-world risks \citep{yurtsever2020survey, HAGGENMULLER2021202, kasneci2023chatgpt}.
It is therefore of little surprise that a significant effort is put into researching various approaches for increasing the trustworthiness and reliability of machine learning methods \citep{abdar2021review, kasneci2023chatgpt, chang2024survey}.
One such approach is uncertainty quantification, which approximates how much we may trust a prediction \citep{hullermeier2021aleatoric}.
However, due to the seemingly disconnected applications, solutions in each domain are often ad hoc, and insights are rarely transferred from one domain to another.
This becomes apparent when comparing the advances in quantifying uncertainty in classification with the more modern data generation tasks.
Various approaches to uncertainty have been proposed in classification; however, results and insights have mostly not been transferred to data generation.
We argue that one possible reason for this is the lack of theoretical generalizations.
For example, model outputs in classification are usually represented via finite probability vectors, which are relatively easy to understand compared to generative models, which may only return samples of an implicit distribution of arbitrary complexity.
That such an implicit distribution exists, at least in theory, is supported by the assumption that we may sample infinitely often from a generative model.
Further, we argue that a general theory requires a mathematically rigorous framework to formulate meaningful quantities, which apply to simple cases, like classification, but also complicated cases, like generative modeling.
Only then may we be able to transfer concepts, methods, and results across this vast landscape of machine learning applications.
In this PhD thesis, we provide such a framework for uncertainty quantification for classification and beyond, which is based on the flexible and broad definition of proper scores.
Proper scores are a general class of loss functions for distribution predictions, covering a wide range of possible applications.
In the following, we will derive various quantities of uncertainty based on a given proper score, and demonstrate their effectiveness across various real-world evaluations of the aforementioned tasks.
This will allow an understanding of the concepts of uncertainty quantification beyond each application, preparing for new, currently unknown machine learning tasks in the future.

\section{Overview}
\label{sec:overview}

In this thesis, we establish a connection between various quantities of uncertainties used in machine learning via proper scores.
Proper scores, also known as proper scoring rules, are loss functions used for predictive distributions \citep{gneitingscores}.
They are mostly prominent for classification, however, their axiomatic definition allows a straightforward extension to non-classification tasks, like regression or general distributions, like those encountered in the image, audio, or natural language domain.
Machine learning research progresses fast and extends to increasingly more complex forecasting domains.
Originally, classification and regression have been of fundamental interest, but new, impactful tasks have been established in recent years.
Among the most prominent are generative tasks, where machine learning models have the purpose of generating data resembling closely the training domain.
The rise of generative models, especially large language models, happened quickly and unpredictably.
Due to this quick development, theoretical understanding and intuition fell behind in providing guidance since most theory was developed with classification and regression as applications in mind.
Especially proper scores received a large interest from various forecasting communities beyond the field of machine learning throughout their over 70 years of history \citep{VERIFICATIONOFFORECASTSEXPRESSEDINTERMSOFPROBABILITY, mccarthy1956measures, winkler1969scoring, savage1971elicitation, gneitingscores}.
Even though they are often not considered in their most general form, specific cases are used in most machine learning theories and applications due to their axiomatic nature.
Their defining property is that they are minimized by predicting the target distribution \citep{gneitingscores}.
Due to this very general definition, proper scores can be found almost everywhere in practice, even when someone may not be aware they are using a proper score.
Examples of proper scores include the mean-squared error in regression and cross-entropy loss in classification.
More generally, the log-likelihood of any exponential family is also a proper score \citep{dawid2007geometry}.
Other prominent examples include the Continuous Ranked Probability Score (CRPS) \citep{zamo2018estimation} for time series forecasts, like weather forecasts, and the Hyv\"arinen Score \citep{hyvarinen2005estimation}, which is prominently used for diffusion models \citep{ho2020denoising}.
The maximum mean discrepancy (MMD), which is used for non-parametric hypothesis testing \citep{JMLR:v13:gretton12a} and image generator training \citep{ren2021improving} and evaluation \citep{binkowski2018demystifying}, is, strictly speaking, not a proper score since it requires multiple samples from the target distribution.
However, it is up to a target-dependent constant equivalent to the so-called kernel score, which is a proper score.
Specifically, the MMD is a functional Bregman divergence; the relation between MMD and kernel score can be generalized to a relation between functional Bregman divergences and proper scores.
It follows that any application using the MMD or any other (functional) Bregman divergence for optimization or evaluation of distributions implicitly also uses a proper score because they share the same optimization objective.\footnote{This only holds for the non-convex argument of Bregman divergences, c.f. Section~\ref{sec:bg_breg_div}.}

Consequently, proper scores have already reached far beyond classification and regression, even though a lot of theoretical progress for proper scores is restricted to such tasks \citep{Br_cker_2009, williamson2014geometry, williamson2023geometry}.
Theoretical works, which use a more general notion of proper scores, exist but are usually restricted to mathematical results without offering empirical evaluations or insights for highlighting the relevancy and applicability of proper scores \citep{hendrickson1971proper, dawid2007geometry}. \\

In our daily existence, various forms of uncertainty exist.
Statistically, they can be put into two categories as illustrated by the following quote of Ronald Fisher \citep{hullermeier2021aleatoric}:
\begin{displayquote}
``Not knowing the chance of mutually exclusive events and knowing the chance to be equal are two quite different states of knowledge.''
\end{displayquote}
The \emph{``not knowing''} is often referred to as epistemic uncertainty \citep{hullermeier2021aleatoric}.
This type of uncertainty arises from a lack of knowledge or information about a system or process.
It's often referred to as systematic uncertainty because it's related to gaps in our understanding or limitations in our models.
The key characteristic of epistemic uncertainty is its reducibility.
By gathering more data, improving our models, or refining our knowledge, we can potentially decrease this type of uncertainty.
One example is predicting the weather.
Our current models, while sophisticated, are still imperfect.
There are factors we don't fully understand, and data we might not have access to. This leads to epistemic uncertainty in weather forecasts.
The second part in Fisher's quote about \emph{``knowing the chance''} is typically referred to as aleatoric uncertainty \citep{hullermeier2021aleatoric}.
This type of uncertainty stems from the inherent randomness or stochasticity present in a system or process.
It's often referred to as ``statistical uncertainty'' because it deals with the variability in outcomes that occur even when all inputs and conditions are known.
Its key characteristic is its irreducibility.
Even with perfect knowledge and infinite data, we can't eliminate the variability associated with truly random events.
One example is the rolling of a fair die.
Even though we know all the possible outcomes (1 through 6) and the probability of each, we can't predict with certainty the outcome of any single roll.
This inherent unpredictability is the aleatoric uncertainty.

In Section~\ref{sec:summary_uncertainties_via_bvd} we summarize Chapter~\ref{ch:uncertainties_via_bvd} of this thesis, where we use the bias-variance decomposition as a core theoretical tool to quantify and improve the uncertainty of predictions in machine learning.
The bias-variance decomposition is a fundamental concept in statistical learning theory that helps analyze the generalization error of a prediction by separating the target prediction into a bias term and a variance term \citep{hastie2009elements}.
The bias term represents the difference between the average prediction and the expected target, while the variance term represents the variability of the prediction due to the randomness in the training data or model parameters.
This decomposition is classically known for the mean squared error or other classification errors.
As part of our contributions, we extend this to the broader class of proper scores.
The main advantage of this generalization is that we can decompose proper scores for predictions beyond probability vectors in classification to arbitrary distributions in generative models.
Further, the bias-variance decomposition is closely related to model averaging, a special case of ensemble learning, which is a robust technique to enhance predictive performance and mitigate overfitting \citep{hastie2009elements}.
Ensemble learning operates on the principle of combining the predictions of multiple base learners, often constructed using different algorithms or trained on diverse subsets of the data, to yield a collective prediction that surpasses the capabilities of any individual model.
Model averaging refers to computing the average of a set of predictions to ``marginalize out'' an error source of epistemic uncertainty \citep{hullermeier2021aleatoric}.
This can be theoretically proven under certain assumptions via the bias-variance decomposition \citep{ueda1996generalization}.
However, one assumption is that the individual predictions are uncorrelated, which does often not hold in practice.
This assumption is not required for the bias-variance-covariance decomposition introduced by \citep{ueda1996generalization} for the mean-squared error, which we generalize to kernel scores, a special class of proper scores.
We contribute practical estimation procedures and experimental evaluations for generative modeling in various domains, like images, audio, and natural language.

\begin{figure}
\centering
    \includegraphics[width=\textwidth]{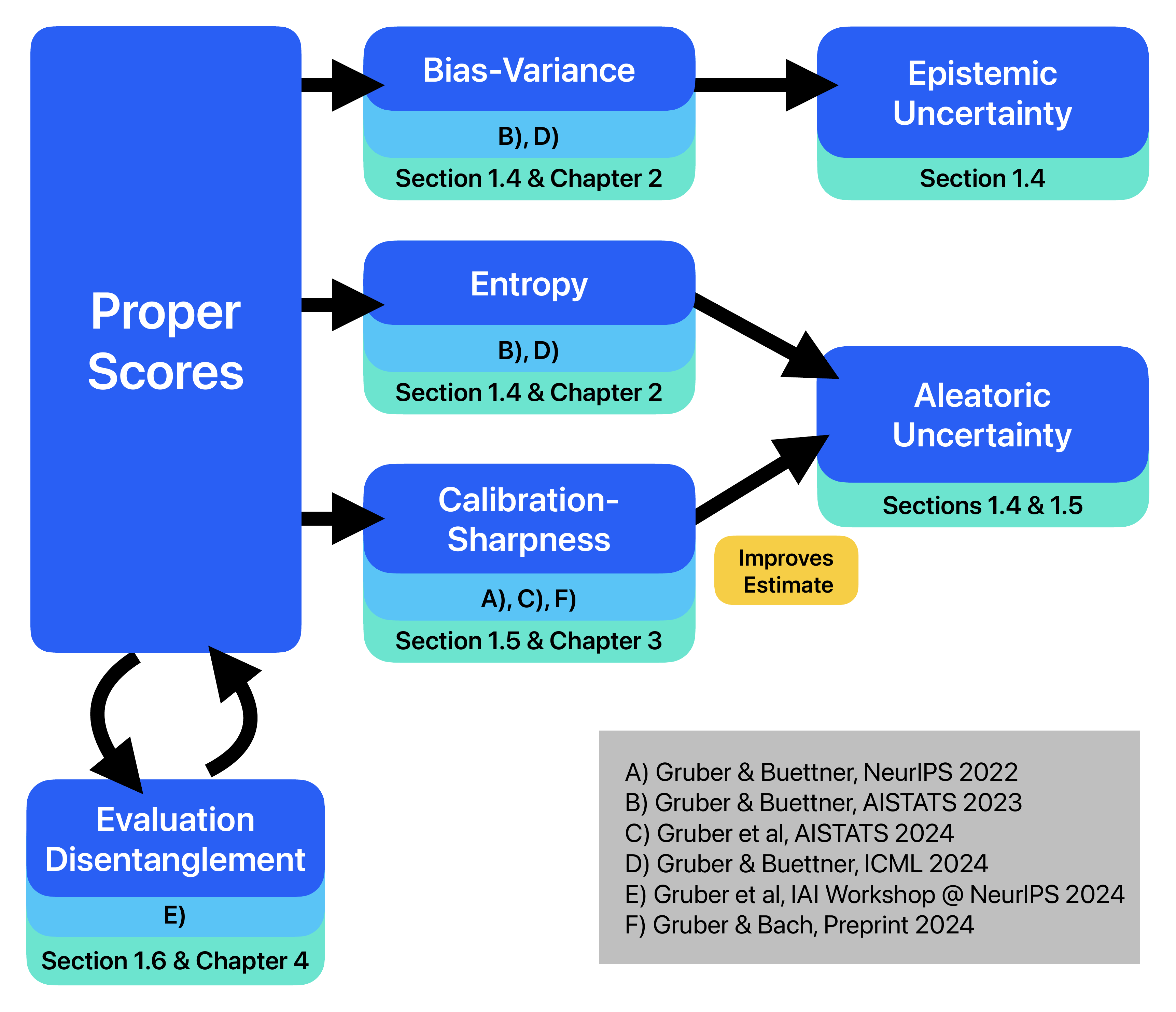}
\caption{
The framework of uncertainty quantification, which we propose in this thesis, is based on proper scores.
Proper scores are a class of loss functions defined in an axiomatic manner and applicable to arbitrary distribution predictions.
Within the framework, a (strictly) proper score induces a bias-variance decomposition, an entropy function, and a calibration-sharpness decomposition.
The variance term can be used as a measure of epistemic uncertainty and to quantify the performance improvement of ensemble predictions.
The entropy function, which also appears in the bias-variance decomposition, is a measure of the aleatoric uncertainty as estimated by the model.
The calibration term in the calibration-sharpness decomposition is referred to as proper calibration error and specifies a way to quantify the reliability of the aleatoric uncertainty estimates.
Further, specific choices of kernel-based proper scores for a target domain can be disentangled into proper scores of target sub-domains, which allows a more fine-grained evaluation of model performance.
}
\label{fig:overview_illus}
\end{figure}

A probabilistic classifier is usually optimized on training data to predict the aleatoric uncertainty by returning class probabilities for an input.
In general, classification tasks involve predicting discrete class labels for given instances \citep{bishop2006pattern}.
As these models are increasingly being employed in critical applications such as healthcare \citep{HAGGENMULLER2021202}, autonomous driving \citep{feng2020deep}, weather forecasting \citep{Gneiting2005WeatherFW}, and financial decision-making \citep{frydman1985introducing}, the need for reliable and interpretable predictions has become of critical importance.
A key aspect of reliability in classification models is the calibration of their predicted probabilities \citep{10.2307/2346866, hekler2023test}.
Calibration refers to the alignment between predicted probabilities and the true likelihoods of outcomes given a specific prediction, ensuring that predictions are reliable regarding their confidence scores \citep{ANewVectorPartitionoftheProbabilityScore}. \\
Despite the advancements in model architectures and learning algorithms, many modern classifiers, such as deep neural networks, are prone to producing overconfident predictions \citep{minderer2021revisiting}.
This overconfidence can be attributed to several factors, including the model's complexity, training data limitations, and inherent biases in learning processes \citep{guo2017calibration}.
Consequently, even models that achieve high accuracy might suffer from poor calibration, leading to potential misinterpretations and decisions without appropriate risk awareness.
Quantifying the calibration of a model is a challenging problem, which we address in our contributions.
In Section~\ref{sec:summary_uncertainty_cal_sharp} we summarize our contributions of Chapter~\ref{ch:uncertainties_via_cal_sharp}, which revolves around calibration errors, their estimators, and their generalization beyond classification.
Specifically, we motivate the definition of proper calibration errors based on a generalization of the calibration-sharpness decomposition of proper scores beyond classification.
This is followed by a general estimator for proper calibration errors in classification and a risk-based optimization procedure for estimators of squared calibration errors.

Last, we summarize Chapter~\ref{ch:disentangling_mean_embeddings} in Section~\ref{sec:summary_neurips_ws_24}, where we contribute a procedure for disentangling the cosine similarity between mean embeddings in a reproducing kernel Hilbert space, which can be easily transferred to the kernel spherical score.
This allows a more fine-grained diagnosis of model misbehavior for image generation.
The practical viability of the theoretical contributions is supported via various real-world experiments, effectively giving novel insights into image generator training.

Our \textbf{core contributions} can be summarized as follows:
\begin{itemize}
    \item We offer a general bias-variance decomposition for proper scores and derive special cases of interest with application to uncertainty quantification.
    Specifically, the bias-variance-covariance decomposition for kernel scores is applied to generative modeling across a wide range of domains, including images, audio, and natural language (c.f. Section~\ref{sec:summary_uncertainties_via_bvd} and Chapter~\ref{ch:uncertainties_via_bvd}).
    \item We introduce the concept of proper calibration errors for reliable uncertainties beyond classification, define a general estimator in classification and establish an optimization procedure for estimators of squared calibration errors (c.f. Section~\ref{sec:summary_uncertainty_cal_sharp} and Chapter~\ref{ch:uncertainties_via_cal_sharp}).
    \item We disentangle the cosine similarity of mean embeddings, related to the kernel spherical score, for a more fine-grained evaluation of image generation models  (c.f. Section~\ref{sec:summary_neurips_ws_24} and Chapter~\ref{ch:disentangling_mean_embeddings}).
\end{itemize}

These core contributions are connected via proper scores and combined represent a novel framework for uncertainty quantification, which we illustrate in Figure~\ref{fig:overview_illus}.
In the next section, we introduce existing concepts and mathematical definitions in the scientific literature, which are required to give a more formal and extensive summary of our contributions.

\section{Preliminaries}
\label{sec:background}

In this section, we introduce necessary mathematical definitions to summarize and discuss the contributions of this thesis.
We first introduce the definition and some basic properties of proper scores in Section~\ref{sec:bg_proper_scores}, followed by (functional) Bregman divergences and their relation to proper scores in Section~\ref{sec:bg_breg_div}.
This is followed by kernel methods in Section~\ref{sec:bg_ps_kernels}, which are used to define some special cases of proper scores useable for distributions beyond classification.
Next, we offer some necessary background on the calibration of model uncertainties in Section~\ref{sec:bg_cal}.
Last, we discuss the notions of aleatoric and epistemic uncertainty more in-depth in Section~\ref{sec:bg_aleatoric_epistemic}, specifically how they are related to the bias-variance decomposition.

Throughout this thesis, we require the notion of convexity for functions and sets as defined in the following.
\begin{definition}[\cite{zalinescu2002convex}]
\label{def:convex_set}
    A set $\Omega$ is said to be \textbf{convex} if and only if for all $x,y \in \Omega$ and $\lambda \in [0,1]$ it holds $\lambda x + \left( 1 - \lambda \right) y \in \Omega$.
\end{definition}
\begin{definition}[\cite{zalinescu2002convex}]
\label{def:convex_func}
    Given a convex set $\Omega$, a function $g \colon \Omega \to \mathbb{R}$ is defined to be \textbf{convex} if and only if for all $x,y \in \Omega$ and $\lambda \in [0,1]$ it holds $g \left( \lambda x + \left( 1 - \lambda \right) y \right) \leq \lambda g \left( x \right) + \left( 1 - \lambda \right) g \left( y \right)$.
\end{definition}
Further, we assume all random variables throughout this thesis are based on a measure space $\left( \Omega, \mathcal{F}, \mu \right)$, where $\Omega$ is the set of outcomes, $\mathcal{F}$ a $\sigma$-algebra, i.e., a set of subsets of $\Omega$ referred to as events, and $\mu \colon \mathcal{F} \to \mathbb{R}$ a measure of events in $\mathcal{F}$ (c.f. \citep{capinski2004measure} for an introduction to measure theory).
Any random variable $X \colon \Omega \to \mathcal{X}$ with outcomes in a set $\mathcal{X}$ implies a probability space $\left( \mathcal{X}, \mathcal{F}_{\mathcal{X}}, \mathbb{P}_X \right)$ with $\mathcal{F}_{\mathcal{X}} \coloneqq \left\{ \left\{ X \left( \omega \right) \mid \omega \in F \right\} \mid F \in \mathcal{F} \right\}$ and $\mathbb{P}_X \coloneqq \mu \circ X^{-1}$.
We will often omit these measure theoretic technicalities and only write $X$ and mention its outcome space $\mathcal{X}$.
Further, sometimes a second random variable $Y$ is given based on which we then assume a joint probability space $\left( \mathcal{X} \times \mathcal{Y}, \mathcal{F}_{\mathcal{X}\mathcal{Y}}, \mathbb{P}_{XY} \right)$, where $\mathcal{F}_{XY} \coloneqq \sigma \left( \mathcal{F}_{\mathcal{X}} \times \mathcal{F}_{\mathcal{Y}} \right)$ is the generated $\sigma$-algebra of the product space $\mathcal{F}_{\mathcal{X}} \times \mathcal{F}_{\mathcal{Y}}$ (which is not a $\sigma$-algebra), and $\mathbb{P}_{XY} \colon \mathcal{F}_{\mathcal{X}\mathcal{Y}} \to \mathbb{R}$ the respective joint probability distribution.
We denote with $\mathbb{P}_{Y \mid X} \coloneqq \frac{\mathrm{d} \mathbb{P}_{XY}}{\mathrm{d} \mathbb{P}_{X}}$ the conditional distribution of $Y$ given $X$.
We may treat it as a function $\mathcal{X} \to \mathcal{P}$, where $\mathcal{P}$ is a set of distributions defined for the measureable space $\left( \mathcal{Y}, \mathcal{F}_{\mathcal{Y}} \right)$.

\subsection{Proper Scores}
\label{sec:bg_proper_scores}

Proper scores are loss functions with a prediction for an outcome's distribution as the first argument and the outcome as the second argument, and which are minimized via the outcome's distribution as the prediction.
This is formally defined as follows.
\begin{definition}[\citep{gneitingscores}]
\label{def:proper_score}
    Assume we are given a convex set $\mathcal{P}$ of distributions defined for a measurable set $\left( \mathcal{Y}, \mathcal{F}_{\mathcal{Y}} \right)$.
    A function $S \colon \mathcal{P} \times \mathcal{Y} \to \mathbb{R}$ is defined to be a \textbf{proper score} if and only if for all $P, Q \in \mathcal{P}$ it holds that
    \begin{equation}
        \mathbb{E}_{Y \sim Q} \left[ S \left( P, Y \right) \right] \geq \mathbb{E}_{Y \sim Q} \left[ S \left( Q, Y \right) \right].
    \end{equation}
\end{definition}
In other words, when we use a function $S$ as a loss function for a dataset $Y_1, \dots, Y_n \sim Q$ of $n$ i.i.d. random variables to quantify the performance of the prediction $P$ via the average loss
\begin{equation}
    \frac{1}{n} \sum_{i=1}^n S \left( P, Y_i \right)
\end{equation}
then the loss $S$ is only \emph{proper} if it is minimized by predicting $P=Q$.
If this minimum is unique, then we refer to the loss $S$ as a \textbf{strictly proper} score.

Classification with $d \in \mathbb{N}$ classes implies that $\mathcal{Y} = \left\{ 1, \dots, d \right\}$ and the set of distributions is the simplex $\mathcal{P} = \Delta^d \coloneqq \left\{ \left(p_1, \dots, p_d \right) \in \mathbb{R}^d \mid \sum_{i=1}^d p_i = 1 \right\}$.
Common examples of proper scores are given for $P \in \Delta^d$ and $y \in \mathcal{Y}$ in the following.
The Brier Score is defined via
\begin{equation}
    S_{\mathrm{BS}} \left(P, y \right) \coloneqq \sum_{i=1}^d P_i^2 - 2 P_y
\end{equation}
and can be interpreted as the mean-squared error for classification.
The log score is given by
\begin{equation}
    S_{\log} \left(P, y \right) \coloneqq - \log P_y,
\end{equation}
which is closely related to the cross-entropy and negative log-likelihood.
Further, the spherical score is another example via
\begin{equation}
    S_{\mathrm{spherical}} \left(P, y \right) \coloneqq - \frac{P_y}{\left\lVert P \right\rVert_2},
\end{equation}
where $\left\lVert x \right\rVert_2 \coloneqq \sqrt{\left\langle x, x \right\rangle_{\mathbb{R}^d}}$ is the euclidean distance based on the dot product $\left\langle x, y \right\rangle_{\mathbb{R}^d} \coloneqq \sum_{i=1}^d x_i y_i$ for vectors $x,y \in \mathbb{R}^d$.
Throughout this work, we will also explore proper scores beyond classification, for example, the log-likelihood of exponential families in Section~\ref{sec:aistats_23}, the kernel score for generative tasks in Section~\ref{sec:summary_icml_24}, and the kernel spherical score in Section~\ref{sec:summary_neurips_ws_24}, which we introduce after defining kernels.
Note that in some literature the sign of proper scores is flipped, i.e., they are maximized instead of minimized \citep{gneitingscores}.
Since this does not affect theoretical results, it is often left to each author and application what is preferred.
In this work, we mostly use minimization to match the convention of loss functions in machine learning.

In practice, we are usually given a dataset $\left(Y_1, X_1 \right), \dots, \left(Y_n, X_n \right) \sim \mathbb{P}_{XY}$ of $n$ i.i.d. instances of input-target tuples sampled from a joint distribution $\mathbb{P}_{XY}$.
In our theoretical framework, we assume that for all $x \in \mathcal{X}$ it holds $f^* \left( x \right) \coloneqq \mathbb{P}_{Y \mid X=x} \in \mathcal{P}$.
Then, to quantify the performance of a model $f \colon \mathcal{X} \to \mathcal{P}$ it is custom to compute the empirical risk based on $S$ via
\begin{equation}
    \hat{\mathcal{R}} \left( f \right) \coloneqq \frac{1}{n} \sum_{i=1}^n S \left( f \left( X_i \right), Y_i \right).
\label{eq:risk_min}
\end{equation}
The procedure to optimize $f$ according to the empirical risk is usually referred to as risk minimization \citep{bishop2006pattern, hastie2009elements}.
If the score $S$ is proper then it holds that
\begin{equation}
    \mathbb{E} \left[ \hat{\mathcal{R}} \left( f \right) \right] \geq \mathbb{E} \left[ \hat{\mathcal{R}} \left( f^* \right) \right],
\end{equation}
where the expectation is with respect to the given dataset.

So far, we have introduced proper scores as a class of loss functions for distribution predictions.
Furthermore, we want to emphasize that Definition~\ref{def:proper_score} may be interpreted as more than just a definition, and, instead, as an axiomatic criterion we would want to always hold for evaluating the performance of models.
It may be quite surprising that such a minimalistic definition will enable all of our contributions in the following thesis.
However, as we will see, by simply choosing a loss function which is a proper score, we receive a fixed measure of aleatoric uncertainty, epistemic uncertainty, calibration error, and a measure of information monotonicity.
These quantities are directly implied by the proper score and do not require further choices.
This simplicity is beneficial since the procedures for uncertainty quantification in the literature are often ad-hoc and without theoretical guidance.
According to our framework, the question of what uncertainty measure one chooses is directly linked to what proper score is reasonable.
Our contributions may provide better guidance for constructing uncertainty measures of new emerging tasks in the rich field of machine learning.

The critical link, which allows the use of a large toolbox of mathematical techniques for connecting proper scores with various uncertainty measures, is by showing how each proper score implicitly defines a convex function.
Specifically, we heavily make use of the following definition.
\begin{definition}[\citep{dawid2007geometry}]
    For a proper score $S \colon \mathcal{P} \times \mathcal{Y} \to \mathbb{R}$, we define the associated \textbf{entropy} function $H \colon \mathcal{P} \to \mathbb{R}$ by
    \begin{equation}
        H \left( P \right) \coloneqq - \mathbb{E}_{Y \sim P} \left[ S \left( P, Y \right) \right].
    \end{equation}
\end{definition}
Special cases for the previously defined proper scores in a classification setup with $P \in \Delta^d$ are the Shannon entropy $H_{\log} \left( P \right) = -\sum_{i=1}^d P_i \log P_i$ \citep{mackay2003information} induced by the log score, the Gini index $H_{\mathrm{BS}} \left( P \right) = - \sum_{i=1}^d P_i^2$ \citep{hastie2009elements} induced by the Brier Score, and the negative euclidean length $H_{\mathrm{spherical}} \left( P \right) = - \left\lVert P \right\rVert_2$ induced by the spherical score.
Note that the Gini index is sometimes also referred to as Gini impurity.
All entropies induced by proper scores share the following critical property.
\begin{lemma}[\citep{hendrickson1971proper}]
    The negative entropy function of a (strictly) proper score is (strictly) convex.
\end{lemma}
Not only is this fact of central importance, but it is also surprisingly simple to prove.
For completeness and emphasis, we offer proof in the following few lines.
\begin{proof}
We use the notation of \citep{10.3150/16-BEJ857} and write $f \cdot \mathbb{P} \coloneqq \int f \mathrm{d} \mathbb{P} = \mathbb{E}_{Y \sim \mathbb{P}} \left[ f \left( Y \right) \right]$ for a function $f$ and a distribution $\mathbb{P}$, and $\mathbf{S} \left( P \right) \left( y \right) \coloneqq S \left( P, y\right)$.
For all $\lambda \in \left( 0, 1 \right)$, and $P, Q \in \mathcal{P}$, and proper score $S$ with negative entropy $G=-H$, it holds
\begin{equation}
\begin{split}
    & G \left( \lambda P + \left( 1 - \lambda \right) Q \right) \\
    & = \mathbf{S} \left( \lambda P + \left( 1 - \lambda \right) Q \right) \cdot \left( \lambda P + \left( 1 - \lambda \right) Q \right) \\
    & = \lambda \mathbf{S} \left( \lambda P + \left( 1 - \lambda \right) Q \right) \cdot P + \left( 1 - \lambda \right) \mathbf{S} \left( \lambda P + \left( 1 - \lambda \right) Q \right) \cdot Q \\
    \overset{\text{Def.\ref{def:proper_score}}}&{\geq} \lambda \mathbf{S} \left( P \right) \cdot P + \left( 1 - \lambda \right) \mathbf{S} \left( Q \right) \cdot Q \\
    & = \lambda G \left( P \right) + \left( 1 - \lambda \right) G \left( Q \right).
\end{split}
\end{equation}
\end{proof}

We can now introduce an important connection of proper scores to the class of (functional) Bregman divergences.
Bregman divergences play a central role in formulating the bias-variance decomposition within Chapter~\ref{ch:uncertainties_via_bvd}.

\subsection{Bregman Divergences}
\label{sec:bg_breg_div}

Bregman divergences occur in a wide range of applications for quantifying the dissimilarity of two vectors \citep{banerjee2005clustering, frigyik2008functional, si2009bregman, gupta2022ensembles}.
They are usually defined as follows.
\begin{definition}[\cite{bregman1967relaxation}]
\label{def:RdBD}
    Let $g \colon U \to \mathbb{R}$ be a differentiable, convex function with $U \subset \mathbb{R}^d$.
    The \textbf{Bregman divergence} $D_g \colon U \times U \to \mathbb{R}$ generated by $g$ for $x,y \in U$ is defined by
    \begin{equation}
        D_g \left( x, y \right) \coloneqq g \left( y \right) - g \left( x \right) - \left\langle \nabla g \left( x \right), y - x \right\rangle_{\mathbb{R}^d}.
    \end{equation}
\end{definition}
In general, Bregman divergences are not symmetric in their arguments and are equal to zero whenever $x=y$.
However, our contributions require a more general definition, which we will refer to as functional Bregman divergences.
These are defined based on subgradients of convex functions for general vector spaces.
Subgradients are elements within the dual vector space of a given vector space (c.f. \citep{zalinescu2002convex} for an extensive overview) and are defined as follows.

\begin{definition}[\citep{10.3150/16-BEJ857}]
    Let $g \colon U \to \mathbb{R}$ be a convex function in a vector space $V \supseteq U$ with dual vector space $V^*$ and bilinear pairing $\left\langle x^*, x \right\rangle_V \coloneqq x \left( x ^* \right)$ for $x \in V$ and $x^* \in V^*$.
    The \textbf{subdifferential} of $g$ at a point $x \in U$ is defined as 
    \begin{equation}
        \partial g \left( x \right) = \left\{ x^\prime \in V^* \mid g \left( y \right) \geq g \left( x \right) + \left\langle x^\prime, y - x \right\rangle_V \text{ for all } y \in U \right\}.
    \end{equation}
    An element $x^\prime \in \partial g \left( x \right)$ is referred to as \textbf{subgradient} of $g$ at $x$.
    We call a function $g^\prime \colon U \to V^*$ a \textbf{selection of subgradients} of $g$ on $U$ if and only if for all $x \in U$ it holds $g^\prime \left( x \right) \in \partial g \left( x \right)$.
\label{def:subdiff}
\end{definition}
If the context is clear, we follow the convention to simply refer to a selection of subgradients as \emph{subgradient function} or just \emph{subgradient}.
Based on the previous definition, we can now define functional Bregman divergences for general vector spaces.
\begin{definition}[\cite{10.3150/16-BEJ857}]
\label{def:breg_div}
Assume a convex function $g \colon U \to \mathbb{R}$ and a selection of subgradients $g^\prime \colon U \to V^*$ for dual vector spaces $V \supseteq U$ and $V^*$ with natural pairing $\left\langle ., . \right\rangle_V \colon V \times V^* \to \mathbb{R}$.
The \textbf{functional Bregman divergence} $D_{g, g^\prime} \colon U \times U \to \mathbb{R}$ generated by $(g, g^\prime)$ is defined by
\begin{equation}
\label{eq:breg_div}
    D_{g,g^\prime} \left( x, y \right) \coloneqq g \left( y \right) - g \left( x \right) - \left\langle g^\prime \left( x \right), y - x \right\rangle_V.
\end{equation}
\end{definition}
A basic property of a functional Bregman divergence $D_{g,g^\prime}$ is that for all $x,y \in U$ it holds $D_{g,g^\prime} \left( x, y \right) \geq 0$ and $D_{g,g^\prime} \left( x, x \right) = 0$.
The minimum is unique if the function $g$ is strictly convex.
If $V = \mathbb{R}^d = V^*$ and the function $g$ is differentiable, then we recover the definition for Bregman divergences (c.f. Def~\ref{def:RdBD}), and we may write $D_g \coloneqq D_{g,g^\prime}$ since $g^\prime = \nabla g$ is then the unique subgradient function of $g$.
A geometric visualization of a Bregman divergence for $V = \mathbb{R} = V^*$ and $g \left( x \right) = x \ln x + \left( 1 - x \right) \ln \left( 1 - x \right)$ is given in Figure~\ref{fig:breg_div_illus_1}.
Further, for the case when the generating function is of the form $g \colon \mathbb{R} \to \mathbb{R}$ and differentiable, we can also give an alternative characterization only via the gradient $\nabla g$ since it holds
\begin{equation}
\label{eq:breg_div_2}
    D_g \left( x, y \right) = \int_{x}^y \nabla g \left( z \right) \mathrm{d} z - \left\langle \nabla g \left( x \right), y - x \right\rangle_{\mathbb{R}}.
\end{equation}
This gives a second geometric representation, which can also be visualized nicely as in Figure~\ref{fig:breg_div_illus_2}.
If the function $g$ is not only differentiable but also \emph{strictly} convex, then the inverse of the gradient $\nabla g$ exists \citep{rockafellar1970convex}.
By writing $x^\prime = \nabla g \left( x \right)$ and $y^\prime = \nabla g \left( y \right)$, this allows another representation of the Bregman divergence $D_g$ via
\begin{equation}
    D_g \left( x, y \right) = \int_{y^\prime}^{x^\prime} \left( \nabla g \right)^{-1} \left( z^\prime \right) \mathrm{d} z^\prime - \left\langle \left( \nabla g \right)^{-1} \left( y^\prime \right), x^\prime - y^\prime \right\rangle_{\mathbb{R}}.
\label{eq:dual_breg_div_2}
\end{equation}
This representation is illustrated in Figure~\ref{fig:dual_breg_div_illus_2}.
An informal visual proof of Equation~\eqref{eq:dual_breg_div_2} is given by noting that the red area in Figure~\ref{fig:breg_div_illus_2} represents Equation~\eqref{eq:breg_div_2} and is equal to the red area of Figure~\ref{fig:dual_breg_div_illus_2} (since they are the same just mirrored on the diagonal), which represents the right-hand side of Equation~\eqref{eq:dual_breg_div_2}.
By comparing Equation~\eqref{eq:breg_div_2} with Equation~\eqref{eq:dual_breg_div_2} we can see that both equations match in their form except that the point $x$ is replaced with $y^\prime$, $y$ with $x^\prime$, and $\nabla g$ with $\left( \nabla g \right)^{-1}$.
This hints at how we can exchange the arguments in a Bregman divergence.
To state this property in general, we require the following definition.

\begin{definition}[\citep{zalinescu2002convex}]
\label{def:convex_conj}
Assume the dual vector spaces $V$ and $V^*$ with pairing $\left\langle .,. \right\rangle_V \colon V^* \times V \to \mathbb{R}$.
The convex conjugate $g^* \colon V^* \to \mathbb{R}$ of a function $g \colon V \to \mathbb{R} \cup \left\{ - \infty, \infty \right\}$ is defined by
\begin{equation}
    g^* \left( x^\prime \right) \coloneqq \sup_{x \in V} \left\langle x^\prime, x \right\rangle_V - g \left( x \right).
\end{equation}    
\end{definition}
This definition is given in its most general form for arbitrary dual vector spaces, however, we can significantly simplify it under certain assumptions.
Based on \citep{banerjee2005clustering}, if the function $g$ is convex and differentiable with $V = \mathbb{R}^d = V^*$, it holds for all $x^\prime \in V^*$ that 
\begin{equation}
    g^* \left( x^\prime \right) = \left\langle x^\prime, \nabla g \left( x^\prime \right) \right\rangle_{\mathbb{R}^d} - g \left( \nabla g \left( x^\prime \right) \right).
\end{equation}
And, if the function $g$ is additionally \emph{strictly} convex, then
\begin{equation}
    \nabla g^* \left( x^\prime \right) = \left( \nabla g \right)^{-1} \left( x^\prime \right).
\end{equation}
Consequently, we can bring Equation~\eqref{eq:dual_breg_div_2} in the form of Equation~\eqref{eq:breg_div} via the convex conjugate of the generating convex function.
Specifically, we have
\begin{equation}
    D_g \left( x, y \right) = g^* \left( x^\prime \right) - g^* \left( y^\prime \right) - \left\langle \nabla g^* \left( y^\prime \right), x^\prime - y^\prime \right\rangle_{\mathbb{R}},
\end{equation}
which leads to the following Lemma.
\begin{lemma}[\citep{banerjee2005clustering}]
For a strictly convex and differentiable function $g \colon U \to \mathbb{R}$ with $U \subseteq \mathbb{R}^d$ and for all $x, y \in U$ it holds
\begin{equation}
     D_g \left( x, y \right) = D_{g^*} \left( y^\prime, x^\prime \right)
\end{equation}
with $y^\prime = \nabla g \left( y \right)$ and $x^\prime = \nabla g \left( x \right)$.
\label{le:breg_arg_flip}
\end{lemma}
A geometric representation for the Bregman divergence $D_{g^*}$ is given in Figure~\ref{fig:dual_breg_div_illus_1}, which relates to Figure~\ref{fig:dual_breg_div_illus_2} the same way as Figure~\ref{fig:breg_div_illus_1} relates to Figure~\ref{fig:breg_div_illus_2}.
Specifically, the red lines in Figure~\ref{fig:dual_breg_div_illus_1} and Figure~\ref{fig:breg_div_illus_1} have the same length (note the different scales), which is a visual proof for Lemma~\ref{le:breg_arg_flip}.
Informally, Lemma~\ref{le:breg_arg_flip} says that we have to change a Bregman divergence to a so-called \emph{dual} Bregman divergence, by exchanging the convex function with its convex conjugate and transforming the arguments into a dual space.
As part of our contribution, we generalize Lemma~\ref{le:breg_arg_flip} to functional Bregman divergences, which is a required intermediate step for our core contribution, a general bias-variance decomposition.

\newcommand\py{0.2}
\newcommand\px{0.7}
\newcommand\lscale{6}
\newcommand\DEscale{1.7}
\newcommand\entropyX{(\px*ln(\px)+(1-\px)*ln(1-\px))}
\newcommand\entropyY{(\py*ln(\py)+(1-\py)*ln(1-\py))}
\newcommand\gradX{(ln(\px)-ln(1-\px))}
\newcommand\gradY{(ln(\py)-ln(1-\py))}
\newcommand\dentropyX{(ln(1+exp(\gradX))/\lscale)}
\newcommand\dentropyY{(ln(1+exp(\gradY))/\lscale)}
\newcommand\igradX{\px}
\newcommand\igradY{\py}
\begin{figure}
\vskip-1in
\centering
\begin{subfigure}{0.5\textwidth}
\centering
\resizebox{0.95\columnwidth}{!}{
\begin{tikzpicture}[every node/.style={scale=.3}]
    \draw [->] (-0.1,0)--(1.2,0) node[right, above] {};
    \foreach \x in {0,1}
       \draw[xshift=\x cm] (0pt,-1pt)--(0pt,1pt);
    \foreach \x in {-1,0}
       \draw[yshift=\x cm] (-1pt,0pt)--(1pt,0pt);
    \draw node[left] at (0,-1) {$-1$};
    \draw node[below] at (-0.08,0.2) {$0$};
    \draw node[below] at (1,0.2) {$1$};
    
    \draw [->] (0,-1.1)--(0,0.4) node[right] {$\;g$};
    
    \draw [domain=0.001:0.999, variable=\x]
      plot ({\x}, {(\x)*ln(\x) + (1 - \x)*ln(1 - \x) )});

    \draw [brown] (1, {\gradX-(\gradX*\px)+\entropyX}) -- (\py, {\py*\gradX-\px*\gradX+\entropyX});
    \draw [red!80, line width=0.3mm] (\py, {\entropyY}) -- (\py, {\py*\gradX-\px*\gradX+\entropyX});

    \draw[black, dashed] (\px, 0)--(\px, {\entropyX}) node[below] at (\px, 0.15) {$x$};
    \draw[black, dashed] (\py, 0)--(\py, {\entropyY}) node[below] at (\py, 0.15) {$y$};
\end{tikzpicture}
    }
\caption{Via $g \left( x \right) = x \ln x + \left( 1 - x \right) \ln \left( 1 - x \right)$}
\label{fig:breg_div_illus_1}
\end{subfigure}%
\begin{subfigure}{0.5\textwidth}
\centering
\resizebox{0.9\columnwidth}{!}{%
\begin{tikzpicture}[every node/.style={scale=.28}]
    \draw [->] (-0.1,0)--(1.25,0) node[right, above] {};
    \foreach \x in {0,1}
       \draw[xshift=\x cm] (0pt, -1pt)--(0pt, 1pt);
    \foreach \x in {0,0.5}
       \draw[yshift=\x cm] (-1pt, 0pt)--(1pt, 0pt);
   \draw node[left] at (0,.5) {$3$};

   \draw node[below] at (-0.06,0) {$0$};
   \draw node[below] at (1,-0.02) {$1$};
    
    \draw [->] (0,-0.5)--(0,0.8) node[right] {$\; \nabla g$};
    
    \fill [red!80, domain=\py:\px, variable=\x]
      (\py, {\gradX/\lscale})
      -- plot ({\x}, {(ln(\x)-ln(1 - \x))/\lscale})
      -- (\px, {\gradX/\lscale})
      -- cycle;
    \draw [domain=0.05:0.99, variable=\x]
      plot ({\x}, {(ln(\x) - ln(1 - \x))/\lscale});

    \draw[black, dashed] (\px, 0)--(\px, {\gradX/\lscale}) node[below] at (\px, 0) {$x$};
    \draw[black, dashed] (\py, 0)--(\py, {\gradY/\lscale}) node[below] at (\py-.05, 0) {$y$};
\end{tikzpicture}
}
\caption{Via $\nabla g \left( x \right) = \ln \frac{x}{1-x}$}
\label{fig:breg_div_illus_2}
\end{subfigure} \\
\begin{subfigure}{0.5\textwidth}
\centering
\resizebox{0.9\columnwidth}{!}{%
\begin{tikzpicture}[every node/.style={scale=.25}]
    \draw [->] (0,-0.12)--(0,1.2) node[above, right] {$\;\left(\nabla g \right)^{-1} = \nabla g^*$};
    \fill [red!80, domain={\gradY/\lscale}:{\gradX/\lscale}, variable=\x]
      ({\gradY/\lscale}, \py)
      -- plot ({\x}, {(exp(\x*\lscale))/(exp(\x*\lscale)+1)})
      -- ({\gradX/\lscale}, \py)
      -- cycle;
    \foreach \x in {0,1}
       \draw[yshift=\x cm] (-1pt, 0pt)--(1pt, 0pt);
    \foreach \x in {-.5,.5}
       \draw[xshift=\x cm] (0pt, -1pt)--(0pt, 1pt);
    \draw node[below] at (.5, -0.02) {$3$};
    \draw node[below] at (-.5, -0.02) {$-3$};

    \draw node[left] at (0,-0.08) {$0$};
    \draw node[left] at (0,1) {$1$};
    
    \draw [->] (-\lscale*0.09,0)--(\lscale*0.11,0) node[above, right] {};
    
    \draw [domain=-0.5:0.6, variable=\x]
      plot ({\x}, {(exp(\x*\lscale))/(exp(\x*\lscale)+1)});

    \draw[black, dashed] ({\gradY/\lscale}, 0)--({\gradY/\lscale}, \py) node[below] at ({\gradY/\lscale}, 0) {$y^\prime$};
    \draw[black, dashed] ({\gradX/\lscale}, 0)--({\gradX/\lscale}, \px) node[below] at ({\gradX/\lscale}, 0) {$x^\prime$};

\end{tikzpicture}
}
\caption{Via $\left( \nabla g \right)^{-1} \left( x^\prime \right) = \frac{\exp x^\prime}{1 + \exp x^\prime}$}
\label{fig:dual_breg_div_illus_2}
\end{subfigure}%
\begin{subfigure}{0.5\textwidth}
\centering
\resizebox{0.9\columnwidth}{!}{
\begin{tikzpicture}[every node/.style={scale=.26}]
    \draw [->] (0,-0.1)--(0,.7*\DEscale) node[above, right] {$\;g^*$};
    \foreach \x in {0,.5*\DEscale}
       \draw[yshift=\x cm] (-1pt, 0pt)--(1pt, 0pt);
    \foreach \x in {0,.5}
       \draw[xshift=\x cm] (0pt, -1pt)--(0pt, 1pt);
    \draw node[below] at (.5,-0.02) {$3$};

    \draw node[left] at (0,-0.08) {$0$};
    \draw node[left] at (0,.5*\DEscale) {$3$};
    
    \draw [->] (-\lscale*0.09,0)--(\lscale*0.11,0) node[above, right] {};
    
    \draw [domain={-\lscale*0.09}:{\lscale*0.11}, variable=\x]
      plot ({\x}, {ln(1+exp(\x*\lscale))/\lscale*\DEscale});

    \draw[black, dashed] ({\gradX/\lscale}, 0)--({\gradX/\lscale}, {\dentropyX*\DEscale}) node[below] at ({\gradX/\lscale}, 0) {$x^\prime$};
    \draw[black, dashed] ({\gradY/\lscale}, 0)--({\gradY/\lscale}, {\dentropyY*\DEscale}) node[below] at ({\gradY/\lscale}, 0) {$y^\prime$};

    \draw [brown] ({-((-\gradY/\lscale)*\igradY+\dentropyY)/\igradY}, 0) -- ({\gradX/\lscale}, {\gradX/\lscale*\igradY*\DEscale+\dentropyY*\DEscale-\gradY/\lscale*\igradY*\DEscale});
    \draw [red!80, line width=0.3mm] ({\gradX/\lscale}, {\dentropyX*\DEscale}) -- ({\gradX/\lscale}, {\gradX/\lscale*\igradY*\DEscale+\dentropyY*\DEscale-\gradY/\lscale*\igradY*\DEscale});

\end{tikzpicture}
    }
\caption{Via $g^* \left( x^\prime \right) = \ln \left( 1 + \exp x^\prime \right)$}
\label{fig:dual_breg_div_illus_1}
\end{subfigure}
\caption{
Four geometric representations of the same Bregman divergence between points $x$ and $y$ (here: Kullback-Leibler divergence). 
\textbf{Top left:} The Bregman divergence is the distance between the gradient at $x$ (brown line) and the convex function itself at point $y$ (here: negative Shannon entropy).
\textbf{Top right:} The same numeric value is represented by the area of the ``triangle'' between points $x$ and $y$ according to the gradient function (here: logit function). Flipping the arguments of the Bregman divergence would result in the area below the curve.
\textbf{Bottom left:} Moving into the dual space by computing the inverse of the gradient function (here: softmax function) and defining $x^\prime = \nabla g \left( x \right)$ and $y^\prime = \nabla g \left( y \right)$ results in the same area.
\textbf{Bottom right:} The Bregman divergence based on the convex conjugate function (here: softplus) in the dual space is identical to the original Bregman divergence. The red lines have the same length under rescaling.
}
\label{fig:breg_div_illus}
\end{figure}
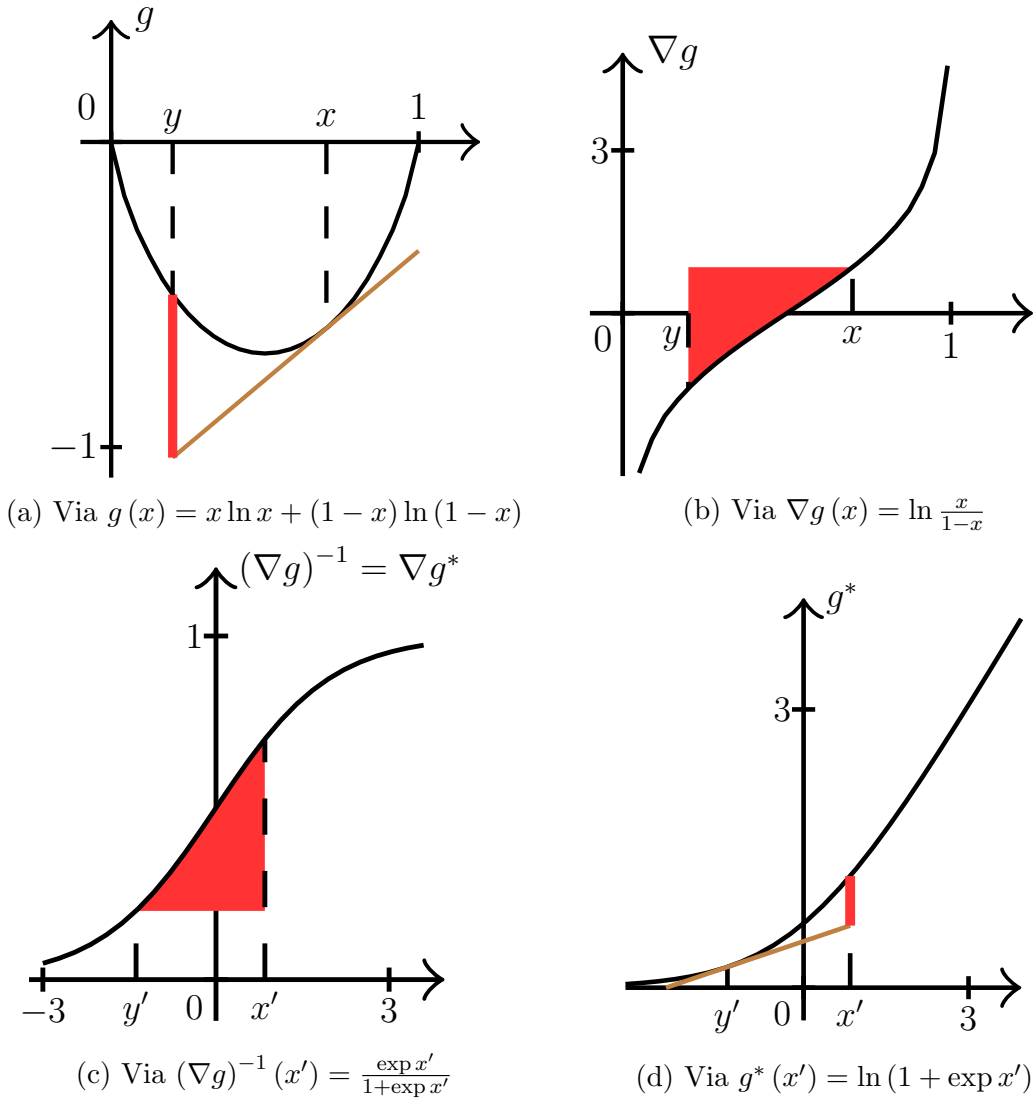

We now establish the connection between proper scores and functional Bregman divergences.
First of all, note that for a convex set $\mathcal{P}$ of distributions defined on a sample space $\mathcal{Y}$, the linear hull 
\begin{equation}
    \operatorname{span} \mathcal{P} \coloneqq \left\{ \sum_{i=1}^n a_i P_i \;\Bigg|\; n \in \mathbb{N}, a_1, \dots a_n \in \mathbb{R}, P_1, \dots P_n \in \mathcal{P} \right\}
\end{equation}
is a vector space and its dual is given by the space of all $\mathcal{P}$-integrable functions $\mathcal{L} \left( \mathcal{P} \right) = \left\{ f \colon \mathcal{Y} \to \mathbb{R} \mid -\infty < \int f \mathrm{d} P < \infty, \text{ for all } P \in \mathcal{P} \right\}$ \citep{10.3150/16-BEJ857}.
Based on these definitions, one can relate proper scores to functional Bregman divergences in the following.
\begin{lemma}[\citep{10.3150/16-BEJ857}]
    For a proper score $S \colon \mathcal{P} \times \mathcal{Y} \to \mathbb{R}$ with associated negative entropy $G$, and for all $P,Q \in \mathcal{P}$ it holds
    \begin{equation}
        \mathbb{E}_{Y \sim Q} \left[ S \left( P, Y \right) \right] = D_{G,\mathbf{S}} \left( P, Q \right) + G \left( Q \right),
    \end{equation}
    where $\mathbf{S}$ is a subgradient function of $G$ defined by $\mathbf{S} \left( P \right) \left( y \right) = - S \left(P, y \right)$ for $y \in \mathcal{Y}$.
\label{le:ps_breg_div}
\end{lemma}
This lemma is of central importance since it tells us that any proper score is connected to a respective (functional) Bregman divergence.
Throughout this thesis, we may refer to $D_{G,\mathbf{S}}$ as the associated divergence of a proper score $S$, and simply write $D \coloneqq D_{G,\mathbf{S}}$ in these cases.
Further, the only difference between a proper score and its associated divergence is a function of the target distribution (here: $Q$).
Since, by definition, we optimize the first argument in a proper score (the predicted distribution, here: $P$), optimizing an expected proper score is identical to optimizing its associated divergence in their first arguments.
A central part of our contributions is a bias-variance decomposition and a calibration-sharpness decomposition for proper scores, which we express via functional Bregman divergences.
Consequently, Lemma~\ref{le:ps_breg_div} will be used repeatedly throughout this thesis.

Bregman divergences are also deployed in the context of random variables.
Specifically, we can use Bregman divergences to generalize how we measure the variance of a random variable by using the definition of Bregman Information as follows.
\begin{definition}[\citep{banerjee2005clustering}]
For a random variable $X$ with outcomes in $\mathbb{R}^d$ and convex function $g \colon \mathbb{R}^d \to \mathbb{R}$ the \textbf{Bregman Information} is defined by
\begin{equation}
    \mathbb{B}_g \left( X \right) \coloneqq \min_{\mu \in \mathbb{R}^d} \mathbb{E} \left[ D_g \left( \mu, X \right) \right].
\end{equation}
\label{def:finite_BI}
\end{definition}
\cite{banerjee2005clustering} show that this definition is equal to the following two alternative representations.
First, we have
\begin{equation}
    \mathbb{B}_g \left( X \right) = \mathbb{E} \left[ D_g \left( \mathbb{E} \left[ X \right], X \right) \right],
\label{eq:finite_BI_repr_2}
\end{equation}
which turns the Bregman Information into the expected Bregman divergence between a random variable and its mean.
Second, we have
\begin{equation}
    \mathbb{B}_g \left( X \right) = \mathbb{E} \left[ g \left( X \right) \right] - g \left(  \mathbb{E} \left[ X \right] \right),
\label{eq:finite_BI_jensen_gap}
\end{equation}
which is the gap in Jensen's inequality for convex functions \citep{capinski2004measure}.
If $g \left( x \right) = x^2$, then we recover the definition of the variance of a random variable \citep{capinski2004measure}.
For our contribution, we are required to generalize the Bregman Information to general vector spaces, i.e., we need a \emph{functional} Bregman Information.
Note that the right-hand side of Equation~\eqref{eq:finite_BI_jensen_gap} does not involve the gradient of the generating function $g$, unlike Equation~\eqref{eq:finite_BI_repr_2} and Definition~\ref{def:finite_BI}.
Consequently, a generalization of the Bregman Information to the functional case is invariant to the choice of the subgradient function.
Due to this reason, we deviate from the original definition and, instead, use Equation~\eqref{eq:finite_BI_jensen_gap} for a generalized definition in the following.
\begin{definition}
\label{def:BI}
For a general vector space $V$, a random variable $X$ with outcomes in $U \subseteq V$, and a convex function $g \colon U \to \mathbb{R}$, we define the \textbf{functional Bregman Information} of $X$ generated by $g$ via
\begin{equation}
    \mathbb{B}_g \left( X \right) \coloneqq \mathbb{E} \left[ g \left( X \right) \right] - g \left(  \mathbb{E} \left[ X \right] \right).
\end{equation}
\end{definition}
The Bregman Information of a random variable appears in the following decomposition, which will be of central importance for our contribution.
Further, the following decomposition was proven for non-functional Bregman Information.
\begin{lemma}[\citep{banerjee2005clustering}]
\label{le:breg_decomp}
    Assume a random variable $Y$ with outcomes in $U \subseteq \mathbb{R}^d$, and a differentiable and convex function $g \colon U \to \mathbb{R}$.
    Under the assumption that all integrals are finite, it holds
    \begin{equation}
        \mathbb{E} \left[ D_{g} \left( x, Y \right) \right] = D_{g} \left( x, \mathbb{E} \left[ Y \right] \right) + \mathbb{B}_g \left( Y \right),
    \end{equation}
    for all $x \in U$.
\end{lemma}
As part of our contributions, we generalize Lemma~\ref{le:breg_arg_flip} and Lemma~\ref{le:breg_decomp} from the $\mathbb{R}^d$ case to the functional case.
This allows us to derive a bias-variance decomposition for strictly proper scores.

So far, we have given examples of proper scores for classification tasks.
However, we may also generalize the previous examples and make them applicable to sample-based generative models.
This requires the definition and basic properties of kernels as follows.

\subsection{Kernel Methods}
\label{sec:bg_ps_kernels}

In this section, we discuss various quantities based on positive-semi definite kernels.
This includes an introduction to reproducing kernel Hilbert spaces, examples of kernel-based proper scores, and a kernel-based measure of independence between random variables.
We first start off with an exemplary practical motivation.

The definition of proper scores is often motivated via the average score
\begin{equation}
    \frac{1}{n} \sum_{i=1}^n S \left(P_i, Y_i \right)
\end{equation}
of $n$ i.i.d. prediction-target pairs (c.f. risk minimization in Equation~\eqref{eq:risk_min}).
This works well in practice if our predictions are probability vectors in classification tasks.
However, in modern machine learning applications, like image or natural language generation, our models do not return distributions $P_1, \dots, P_n$ directly but only implicitly via $m$ i.i.d. generated samples $\hat{Y}_{i1}, \dots, \hat{Y}_{im} \sim P_i$ for $i=1,\dots, n$.
In theory, if we could sample $m=\infty$ amount of times, we would still recover the predicted distributions.
But since this is practically not feasible, we are interested in proper scores, which can be computed for finite $m$ via
\begin{equation}
\label{eq:emp_kernel_score}
    \frac{1}{n} \sum_{i=1}^n \hat{S} \left(\left\{ \hat{Y}_{i1}, \dots, \hat{Y}_{im} \right\}, Y_i \right),
\end{equation}
where $\hat{S}$ is an empirical estimator of a proper score $S$ such that 
\begin{equation}
    \hat{S} \left( \left\{ \hat{Y}_{i1}, \dots, \hat{Y}_{im} \right\}, Y_i \right) \to S \left( P_i, Y \right)
\end{equation}
in probability for $m \to \infty$.
Even though the very first example of such a score dates back to \citep{eaton1981method}, research in this direction is still in its infancy.
In this section, we give an overview of such scores based on kernels.

\subsubsection{Preliminaries for Kernels}
\label{sec:bg_kernels}

In essence, kernels are a class of functions, which compare the similarity of two inputs.
We make use of positive semi-definite (p.s.d.) kernels, i.e. functions of the form $k \colon \mathcal{X} \times \mathcal{X} \to \mathbb{R}$ defined on a set $\mathcal{X}$ for which holds for all $x_1, \dots, x_n \in \mathcal{X}$ and $a_1, \dots, a_n \in \mathbb{R}$ that $\sum_{i=1}^n \sum_{j=1}^n a_i k \left( x_i, x_j \right) a_j \geq 0$.
An extensive introduction is given in \citep{scholkopf2002learning}, which we summarize in the following.
For p.s.d. kernels holds that we can construct a Hilbert space $\mathcal{H}$, a function $\phi$, and an inner product $\left\langle .,. \right\rangle_{\mathcal{H}}$ such that $k \left( x, y \right) = \left\langle \phi \left( x \right), \phi \left( y \right) \right\rangle_{\mathcal{H}}$ for all $x,y \in \mathcal{X}$.
The space $\mathcal{H}$ is referred to as reproducing kernel Hilbert space (RKHS) since it has the so-called reproducing property, which states that for all $f \in \mathcal{H}$ and $x \in \mathcal{X}$ we have $\left\langle f, \phi \left( x \right) \right\rangle = f \left( x \right)$.
Further, the function $\phi$ is referred to as a feature map based on its use in kernel methods \citep{scholkopf2002learning}.
For a given $\phi$ the \textbf{mean embedding} of a distribution $P$ with support within $\mathcal{X}$ is defined as
\begin{equation}
    \mu_P \coloneqq \mathbb{E}_{X \sim P} \left[ \phi \left( X \right) \right] \in \mathcal{H}.
\end{equation}
If $\phi$ maps to an appropriately large RKHS and if $\mathcal{X} \subseteq \mathbb{R}^d$, then $\mu_P$ is an injective mapping with respect to $P$.
In that case, it does not lose information on the distribution and the embedding of two distinct distributions is different.
A kernel, which raises an injective mean embedding, is referred to as a \emph{characteristic} kernel \citep{scholkopf2002learning}.
Common examples of characteristic kernels with $\mathcal{X} = \mathbb{R}^d$ are the radial basis function (RBF) kernel defined by
\begin{equation}
    k_{\mathrm{rbf}} \left( x, y \right) \coloneqq \exp \left( - \gamma \left\lVert x - y \right\rVert_2^2 \right),
\end{equation}
or the Laplacian kernel
\begin{equation}
    k_{\mathrm{lap}} \left( x, y \right) \coloneqq \exp \left( - \gamma \left\lVert x - y \right\rVert_1 \right),
\end{equation}
where $\gamma$ is a bandwidth parameter, which controls the sensitivity of the kernels to differences in their arguments, and $\left\lVert . \right\rVert_p$ refers to the p-norm defined for vectors in $\mathbb{R}^d$ via $\left\lVert x \right\rVert_p \coloneqq \left( \left\lvert x_1 \right\rvert^p + \dots + \left\lvert x_d \right\rvert^p \right)^{\frac{1}{p}}$.
To illustrate, if $\mathcal{X} = \mathbb{R}$ and $\gamma=1$ the associated feature map of the RBF kernel is given by
\begin{equation}
    \phi_{\mathrm{rbf}} \left( x \right) = \exp \left( - x^2 \right) \left( 1, x, \frac{x^2}{2}, \frac{x^3}{6}, \dots \right)^\intercal = \exp \left( - x^2 \right) \left( \frac{x^n}{n!} \right)_{n=0 \dots \infty}^\intercal.
\end{equation}
Consequently, the respective mean embedding for a distribution is a vector of all of the distribution's non-central moments.
The emphasis is on the lower moments since the higher moments are increasingly scaled towards zero due to the rapidly diminishing factor $\frac{1}{n!}$.

Examples of non-characteristic kernels for $\mathcal{X} = \mathbb{R}^d$ are the cosine similarity defined by
\begin{equation}
    k_{\mathrm{cos}} \left( x, y \right) \coloneqq \frac{\left\langle x, y \right\rangle_{\mathbb{R}^d}}{\left\lVert x \right\rVert_2 \left\lVert y \right\rVert_2},
\end{equation}
or the polynomial kernel of degree $p \in \mathbb{R}_{>0}$ given by
\begin{equation}
    k_{\mathrm{pol}} \left( x, y \right) \coloneqq \left( \gamma \left\langle x, y \right\rangle_{\mathbb{R}^d} + c \right)^p
\end{equation}
with constants $\gamma,c \in \mathbb{R}_{>0}$ and dot product $\left\langle ., . \right\rangle_{\mathbb{R}^d}$.
For comparison with the RBF kernel, the feature map of the polynomial kernel with $\mathcal{X} = \mathbb{R}$ and $\gamma = 1$ is given by
\begin{equation}
    \phi_{\mathrm{rbf}} \left( x \right) = \left(c_0, c_1 x, \dots, c_p x^p \right)^\intercal
\end{equation}
with $c_i \coloneqq \binom{p}{i} c^{p-i}$.
As we can see, similar to the RBF feature map $\phi_{\mathrm{rbf}}$, the polynomial feature map also maps to non-central moments.
However, the highest moment is specified via the user-defined hyperparameter $p$.
Further, the moments are differently scaled and higher moments may not diminish to zero as quickly.
This demonstrates that a kernel does not necessarily have to be characteristic to provide meaningful mean embeddings.
The use of cosine similarity for semantic embeddings of natural language further supports this claim \citep{steinbach2000comparison}.

Based on the chosen kernel, the mean embedding $\mu_P$ may be infinite-dimensional and, thus, not computable in practice.
However, kernel methods usually aim to express $\mu_P$ only within the respective inner product $\left\langle .,. \right\rangle_{\mathcal{H}}$.
This allows to construct an empirical estimator via the identity $k \left( x, y \right) = \left\langle \phi \left( x \right), \phi \left( y \right) \right\rangle_{\mathcal{H}}$.
The most basic case is to estimate $\left\langle \mu_P, \phi \left( y \right) \right\rangle_{\mathcal{H}}$ for i.i.d. $X_1, \dots, X_n \sim P$ with $\frac{1}{n} \sum_{i=1}^n k \left( X_i, y \right)$ since it holds
\begin{equation}
    \left\langle \mu_P, \phi \left( y \right) \right\rangle_{\mathcal{H}} = \mathbb{E} \left[ \frac{1}{n} \sum_{i=1}^n \left\langle \phi \left( X_i \right), \phi \left( y \right) \right\rangle_{\mathcal{H}} \right] = \mathbb{E} \left[ \frac{1}{n} \sum_{i=1}^n k \left( X_i, y \right) \right]
\end{equation}
and $\frac{1}{n} \sum_{i=1}^n k \left( X_i, y \right) \to \left\langle \mu_P, \phi \left( y \right) \right\rangle_{\mathcal{H}}$ in distribution for $n \to \infty$.
Similarly, the quantities $\left\langle \mu_P, \mu_Q \right\rangle_{\mathcal{H}}$ and $\left\lVert \mu_P \right\rVert_{\mathcal{H}}^2$ can be estimated if we have samples for another distribution $Q$.
In consequence, we can express proper scores based on these quantities as desired in our preliminary discussion (c.f. Equation~\eqref{eq:emp_kernel_score}).

The definition of mean embeddings allows us to "kernelize" the Brier score and the spherical score, as in the next section.
However, it is not possible to compute the logarithm of mean embeddings in general, which restricts us from defining a kernel-based log score in the same manner.
Alternatively, in the realm of quantum information theory, the logarithm of positive semi-definite operators is well-defined \citep{watrous2018theory}.
Modifying mean embeddings to fulfill this property results in a kernel-based log score.
For this, given a distribution $P$ with support within $\mathcal{X}$, we can use the (non-central) covariance operator $\Sigma_P \colon \mathcal{H} \to \mathcal{H}$ defined by
\begin{equation}
\label{eq:non_c_cov_op}
    \Sigma_P \coloneqq \mathbb{E} \left[ \phi \left( X \right) \otimes \phi \left( X \right) \right],
\end{equation}
where $\otimes$ is defined such that $\left( f \otimes g \right) h = \left\langle g, h \right\rangle_{\mathcal{H}} f$ for all $f, g, h \in \mathcal{H}$ \citep{bach2022information}.
It holds that $\Sigma_P$ is a positive semi-definite operator for any $P$.

Last, we require the definition of spectral functions.
Taking a p.s.d. operator $\Sigma$ as argument, a spectral function $\Sigma \mapsto f \left( \Sigma \right) = Q \operatorname{diag} \left( f \left( \lambda_1 \right), f \left( \lambda_2 \right), \dots \right) Q^\intercal$ is defined based on a scalar function $f \colon \mathbb{R} \to \mathbb{R}$, which transforms the eigenvalues $\lambda_1, \lambda_2, \dots$ in the spectral decomposition $\Sigma = Q \operatorname{diag} \left( \lambda_1, \lambda_2, \dots \right) Q^\intercal$ \citep{bach2022information}.
We now have the necessary definitions to introduce kernel-based generalizations of the Brier score, the spherical score, and the log score.

\subsubsection{Kernel-based Proper Scores}
\label{sec:bg_ps_kernel_examples}

A kernel version of the Brier score was first introduced by \cite{eaton1981method}.
More recently, \cite{steinwart2021strictly} provides a more general definition, which we also use in this thesis.
For all generalizations in the following, we assume a convex set of distributions $\mathcal{P}$ defined on a set $\mathcal{Y}$, a sample $y \in \mathcal{Y}$, and a p.s.d. kernel $k \colon \mathcal{Y} \times \mathcal{Y} \to \mathbb{R}$.
Further, we imply that the RKHS $\mathcal{H}$ and all associated operators are with respect to the given kernel.

\begin{figure}
\centering
    \includegraphics[width=0.65\textwidth]{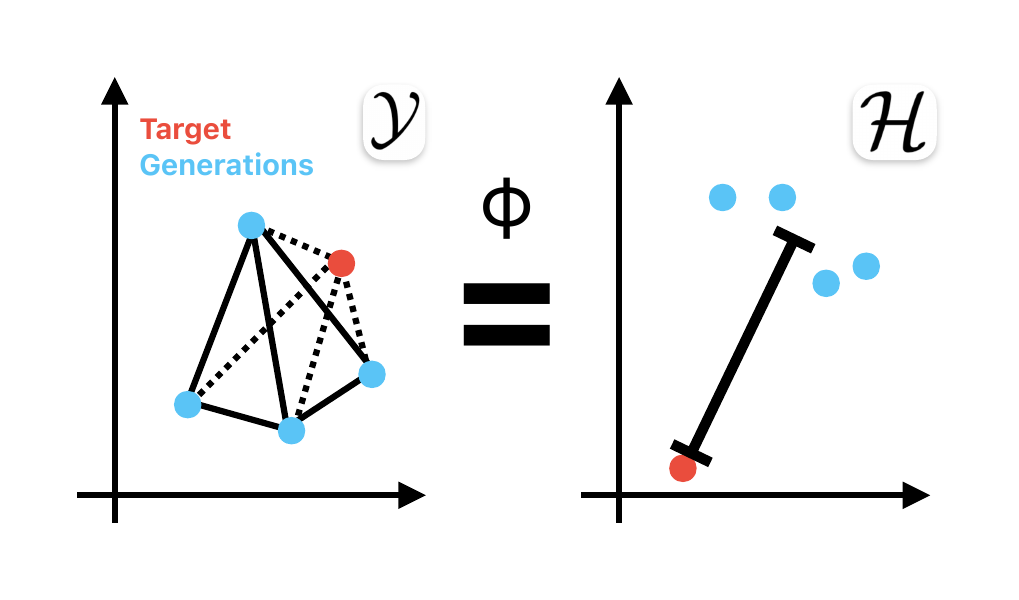}
\caption{
Illustration of kernel score between predicted generations (blue) and a target sample (red).
The kernel score is computed in practice by evaluating the kernel for all pairings of generations in the sample set $\mathcal{Y}$ (\textbf{left}), which is, in theory, equivalent to computing the squared distance between the mean embedding of the generations and the embedding of the target sample in the RKHS $\mathcal{H}$ associated with $k$ via $\phi$ (\textbf{right}).
}
\label{fig:ks_illus}
\end{figure}

Then, the \textbf{kernel score} is defined as
\begin{equation}
    S_{k} \left( P, y \right) \coloneqq \left\lVert \mu_P \right\rVert_{\mathcal{H}} - 2 \left\langle \mu_P, \phi \left( y \right) \right\rangle_{\mathcal{H}}.
\end{equation}
An illustration of kernel scores comparing its representation in the sample space and the RKHS is given in Figure~\ref{fig:ks_illus}.
Its associated negative entropy is given by $G_{k} \left( P \right) = \left\lVert \mu_P \right\rVert^2_{\mathcal{H}}$ and divergence by $D_{k} \left( P, Q \right) = \left\lVert \mu_P - \mu_Q \right\rVert^2_{\mathcal{H}}$.
The divergence $D_k$ matches the definition of the (squared) maximum mean discrepancy (MMD), which is prominently used for non-parametric hypothesis testing \citep{JMLR:v13:gretton12a} and image generator evaluation \citep{binkowski2018demystifying}.
The Brier score is recovered for $\mathcal{P} = \Delta^d$ and $k \left( x, y \right) = \mathbf{1}_{x=y}$, where $\mathbf{1}_{x=y}$ is equal to one if $x=y$ and zero otherwise.

A kernel-based generalization of the spherical score was introduced in \citep{huszar2013scoring}.
We define the \textbf{kernel spherical score} 
as
\begin{equation}
    S_{k\text{-spherical}} \left( P, y \right) \coloneqq -\frac{\left\langle \mu_P, \phi \left( y \right) \right\rangle_{\mathcal{H}}}{\left\lVert \mu_P \right\rVert_{\mathcal{H}}}.
\end{equation}
Its associated negative entropy is given by $G_{k\text{-spherical}} \left( P \right) = \left\lVert \mu_P \right\rVert_{\mathcal{H}}$ and divergence by $D_{k\text{-spherical}} \left( P, Q \right) = \left( 1 - \cos_{\mathcal{H}} \left( \mu_P, \mu_Q \right) \right) \left\lVert \mu_Q \right\rVert_{\mathcal{H}}$, where $\cos_{\mathcal{H}} \left( \mu_P, \mu_Q \right) \coloneqq \frac{\left\langle \mu_P, \phi \left( y \right) \right\rangle_{\mathcal{H}}}{\left\lVert \mu_P \right\rVert_{\mathcal{H}} \left\lVert \mu_Q \right\rVert_{\mathcal{H}}}$ is the cosine similarity in the RKHS $\mathcal{H}$.
The spherical score is recovered for $\mathcal{P} = \Delta^d$ and $k \left( x, y \right) = \mathbf{1}_{x=y}$.

The kernelized Brier and spherical score are part of our contributions in later sections.
However, for completeness and as an outlook for future work, we also discuss a kernel-based generalization of the log score.

\cite{bach2022information}, \cite{bach2024sum}, and \cite{chazal2024statistical} offer an extensive study of the Kullback-Leibler divergence generalized to the kernel-based covariance operator.
Since the expectation of the following kernel-based log score and the kernel-based Kullback-Leibler divergence only differ via the entropy term, similar to all the other score-divergence pairs, many results of these works can be readily translated.
Based on \citep{bach2022information}, we define the \textbf{kernel log score} as
\begin{equation}
    S_{k-\log} \left( P, y \right) \coloneqq - \left\langle \phi \left( y \right), \left( \log \Sigma_P \right) \phi \left( y \right) \right\rangle_{\mathcal{H}}.
\end{equation}
Its associated negative entropy is given by $G_{k-\mathrm{log}} \left( P \right) = \operatorname{tr} \Sigma_P \log \Sigma_P$ and divergence by $D_{k-\log} \left( P, Q \right) = \operatorname{tr} \Sigma_P \left( \log \Sigma_P - \log \Sigma_Q \right)$ under assumption that $\Sigma_P \log \Sigma_Q$ is well-defined \citep{bach2022information}.
The divergence $D_{k-\log}$ is also known as kernel Kullback-Leibler divergence \citep{chazal2024statisticalgeometricalpropertiesregularized}.
Like in the other cases, the log score is recovered for $\mathcal{P} = \Delta^d$ and $k \left( x, y \right) = \mathbf{1}_{x=y}$.
This can be seen by noting that it would follow $\Sigma_P = \operatorname{diag} \left( P_1, \dots, P_d \right)$, and $\phi \left( y \right) = e_y$, where $e_y$ denotes a one-hot encoded vector with a '1' at index $y$, and $\mathcal{H} = \mathbb{R}^d$.

A final overview of the discussed proper scores is given in Table~\ref{tab:proper_scores}, which compares classification and kernel-based cases.
To illustrate their connection further, we also include versions of each proper score for density forecasts based on the $L^2$-inner product $\left\langle p, q \right\rangle_{L^2} \coloneqq \int_{\mathbb{R}} p \left( x \right) q \left( x \right) \mathrm{d} \lambda \left( x \right)$ for densities $p$ and $q$, and Lebesgue measure $\lambda$ \citep{gneiting2014probabilistic}.
The respective $L^2$-norm and $L^2$-cosine similarity are defined via $\left\lVert p \right\rVert_{L^2} \coloneqq \sqrt{\left\langle p, p \right\rangle_{L^2}}$ and $\operatorname{cos}_{L^2} \left( p, q \right) \coloneqq \frac{\left\langle p, q \right\rangle_{L^2}}{\left\lVert p \right\rVert_{L^2} \left\lVert q \right\rVert_{L^2}}$.


\newpage
\begin{landscape}

\renewcommand{\arraystretch}{1.5}
\begin{table}[!h]
    \centering
    \rotatebox{0}{
    \begin{minipage}{\textwidth}
    \centering
    \caption{
    Common examples of proper scores and their entropy and divergence functions across three types of applications.
    Classification is the most prevalent use case for considering proper scores.
    However, they may as well be used for densities or sample-based distributions via kernels.
    The latter refers to distributions, which are not available in practice but are only expressed via a finite set of samples.
    We denote with $p$ and $q$ the densities for respective distributions $P$ and $Q$.
    }
    \label{tab:proper_scores}
    \vskip 0.2in
    \begin{tabular}{c|c|c|c|c|c}
        Score & $\mathcal{Y}$ & $\mathcal{P}$ & $S \left( P, y \right)$ & $H \left( Q \right)$ & $D \left( P, Q \right)$ \\ [0.5ex]
        \hline
        \hline
        Brier & $\left\{1, \dots d \right\}$ & $\Delta^d$ & $\sum_{i=1}^d P_i^2 - 2 P_y$ & $-\sum_{i=1}^d Q_i^2$ & $\left\lVert P - Q \right\rVert_2^2$ \\
        Log & $\left\{1, \dots d \right\}$ & $\Delta^d$ & $- \log P_y$ & $-\sum_{i=1}^d Q_i \log Q_i$ & $\sum_{i=1}^d Q_i \log \frac{Q_i}{P_i}$ \\
        Spherical & $\left\{1, \dots d \right\}$ & $\Delta^d$ & $- \frac{P_y}{\left\lVert P \right\rVert_2}$ & $-\left\lVert Q \right\rVert_2$ & $\left( 1 - \operatorname{cos} \left( P, Q \right) \right) \left\lVert Q \right\rVert_2$ \\
        \hline
        Quadratic & $\mathbb{R}$ & Densities & $\left\lVert p \right\rVert^2_{L^2} - 2 p \left(y \right)$ & $- \left\lVert q \right\rVert^2_{L^2}$ & $\left\lVert p - q \right\rVert^2_{L^2}$ \\
        Log & $\mathbb{R}$ & Densities & $- \log p \left( y \right)$ & $-\int q \left( y \right) \log q \left( y \right) \mathrm{d}y$ & $\int  q \left( y \right) \log \frac{q \left( y \right)}{p \left( y \right)} \mathrm{d}y$ \\
        Spherical & $\mathbb{R}$ & Densities & $- \frac{p \left( y \right)}{\left\lVert p \right\rVert_{L^2}}$ & $-\left\lVert q \right\rVert_{L^2}$ & $\left( 1 - \operatorname{cos}_{L^2} \left( p, q \right) \right) \left\lVert q \right\rVert_{L^2}$ \\
        \hline
        Kernel & $\mathbb{R}^d$ & General & $\left\lVert \mu_P \right\rVert_{\mathcal{H}}^2 - 2 \left\langle \mu_P, \phi \left( y \right) \right\rangle_{\mathcal{H}}$ & $-\left\lVert \mu_Q \right\rVert_{\mathcal{H}}^2$ & $\left\lVert \mu_P - \mu_Q \right\rVert_{\mathcal{H}}^2$ \\
        Kernel Log & $\mathbb{R}^d$ & General & $- \left\langle \phi \left( y \right), \left( \log \Sigma_P \right) \phi \left( y \right) \right\rangle_{\mathcal{H}}$ & $-\operatorname{tr} \Sigma_Q \log \Sigma_Q$ & $\operatorname{tr} \Sigma_Q \left( \log \Sigma_Q - \log \Sigma_P \right)$ \\
        Kernel Spherical & $\mathbb{R}^d$ & General & $- \frac{\left\langle \mu_P, \phi \left( y \right) \right\rangle_{\mathcal{H}}}{\left\lVert \mu_P \right\rVert_{\mathcal{H}}}$ & $-\left\lVert \mu_Q \right\rVert_{\mathcal{H}}$ & $\left( 1 - \operatorname{cos}_{\mathcal{H}} \left( \mu_P, \mu_Q \right) \right) \left\lVert \mu_Q \right\rVert_{\mathcal{H}}$ \\
    \end{tabular}
    \end{minipage}
    }
\end{table}
\end{landscape}

\newpage

\subsubsection{Kernel-based Independence Criterion}

So far, we have used kernels to define special cases of proper scores applicable to a wide range of practical scenarios.
Kernels may also be used for other applications, for example, for quantifying the dependence of random variables.
In this section, we introduce central kernel alignment \citep{cortes2012algorithms}, which is a kernel-based generalization of the Pearson correlation coefficient \citep{cohen2009pearson}.
This quantity plays a central role in our contributions towards disentangling the kernel spherical score in Section~\ref{sec:tmlr_24} and Chapter~\ref{ch:disentangling_mean_embeddings}.

Let $X$ and $Y$ be random variables with outcomes in some sets $\mathcal{X}$ and $\mathcal{Y}$ respectively.
Further, assume two p.s.d. kernels $k \colon \mathcal{X} \times \mathcal{X} \to \mathbb{R}$ and $\Tilde{k} \colon \mathcal{Y} \times \mathcal{Y} \to \mathbb{R}$ with respective RKHS $\mathcal{H}$ and $\Tilde{\mathcal{H}}$, as well as feature maps $\phi \colon \mathcal{X} \to \mathcal{H}$ and $\Tilde{\phi} \colon \mathcal{Y} \to \Tilde{\mathcal{H}}$.
Then, the cross-covariance operator $\mathcal{C}_{XY} \colon \Tilde{\mathcal{H}} \to \mathcal{H}$ is defined via
\begin{equation}
\begin{split}
    \mathcal{C}_{XY}
    & \coloneqq \mathbb{E} \left[ \left( \phi \left( X \right) - \mathbb{E} \left[ \phi \left( X \right) \right] \right) \otimes \left( \Tilde{\phi} \left( Y \right) - \mathbb{E} \left[ \Tilde{\phi} \left( Y \right) \right] \right) \right].
\end{split}
\end{equation}
It is related to the (non-central) covariance operator in Equation~\eqref{eq:non_c_cov_op} via
\begin{equation}
    \mathcal{C}_{XX} = \Sigma_{\mathbb{P}_X} - \mu_{\mathbb{P}_X} \otimes \mu_{\mathbb{P}_X}
\end{equation}
assuming $k = \Tilde{k}$.
The Hilbert-Schmidt norm of an operator $\mathcal{C} \colon \Tilde{\mathcal{H}} \to \mathcal{H}$ of Hilbert spaces $\mathcal{H}$ and $\Tilde{\mathcal{H}}$ with orthonormal bases $a_1, a_2, \dots$ and $b_1, b_2, \dots$ is defined via $\left\lVert \mathcal{C} \right\rVert_{\mathrm{HS}} = \sum_{ij} \left\langle a_i, \mathcal{C} b_j \right\rangle_{\mathcal{H}}$ \citep{gretton2005measuring}.
It reduces to the Frobenius norm if both spaces are finite-dimensional Euclidean spaces.
For $g \in \mathcal{H}$ and $h \in \mathcal{H}^\prime$ it holds $\left\lVert g \otimes h \right\rVert_{\mathrm{HS}} = \left\lVert g \right\rVert_{\mathcal{H}} \left\lVert h \right\rVert_{\Tilde{\mathcal{H}}}$, from which follows that 
\begin{equation}
\begin{split}
    \left\lVert \mathcal{C}_{XY} \right\rVert_{\mathrm{HS}}^2 & = \mathbb{E}_{X,Y,X^\mathrm{c},Y^\mathrm{c}} \left[ k \left( X, X^\mathrm{c} \right) \Tilde{k} \left( Y, Y^\mathrm{c} \right) \right] \\
    & \quad - \mathbb{E}_{X,X^\mathrm{c},Y^\mathrm{c}} \left[ k \left( X, X^\mathrm{c} \right) \mathbb{E}_Y \left[ \Tilde{k} \left( Y, Y^\mathrm{c} \right) \right] \right] \\
    & \quad - \mathbb{E}_{Y,X^\mathrm{c},Y^\mathrm{c}} \left[ \mathbb{E}_X \left[ k \left( X, X^\mathrm{c} \right) \right] \Tilde{k} \left( Y, Y^\mathrm{c} \right) \right] \\
    & \quad + \mathbb{E}_{X,X^\mathrm{c}} \left[ k \left( X, X^\mathrm{c} \right) \right]\mathbb{E}_{Y,Y^\mathrm{c}} \left[ \Tilde{k} \left( Y, Y^\mathrm{c} \right) \right],
\label{eq:HSIC_as_kernel}
\end{split}
\end{equation}
where $\left(X^\mathrm{c}, Y^\mathrm{c} \right)$ is an i.i.d. copy of $\left(X, Y \right)$ \citep{gretton2005measuring}.
Equation~\eqref{eq:HSIC_as_kernel} indicates that we can estimate $\left\lVert \mathcal{C}_{XY} \right\rVert_{\mathrm{HS}}^2$ in practice via the given kernels, even when $\mathcal{H}$ or $\Tilde{\mathcal{H}}$ are infinite-dimensional.
However, the estimated values will never be precisely zero.
This makes comparing multiple random variables problematic since their ranges may have different magnitudes.
Consequently, we use the normalized version referred to as \textbf{central kernel alignment} (CKA) \citep{cortes2012algorithms, chang2013canonical}, which is defined by
\begin{equation}
\begin{split}
    \operatorname{CKA}_{k,\Tilde{k}} \left( \mathbb{P}_{XY} \right) \coloneqq \frac{\left\lVert \mathcal{C}_{XY} \right\rVert_{\mathrm{HS}}^2}{\left\lVert \mathcal{C}_{XX} \right\rVert_{\mathrm{HS}}\left\lVert \mathcal{C}_{YY} \right\rVert_{\mathrm{HS}}}.
\end{split}
\end{equation}
It has the form of a squared correlation coefficient and by the Cauchy-Schwartz inequality lies within $\left[0, 1 \right]$ \citep{chang2013canonical}.
It holds
\begin{equation}
    \operatorname{CKA}_{k,\Tilde{k}} \left( \mathbb{P}_{XY} \right) = 0 \iff \mathbb{E}_{X,Y \sim \mathbb{P}_{XY}} \left[ \phi \left( X \right) \otimes \Tilde{\phi} \left( Y \right) \right] = \mu_{\mathbb{P}_X} \otimes \Tilde{\mu}_{\mathbb{P}_Y} 
\label{eq:CKA0_implies_disentangl}
\end{equation}
with $\Tilde{\mu}_{\mathbb{P}_Y} \coloneqq \mathbb{E}_{Y \sim \mathbb{P}_Y} \left[ \Tilde{\phi} \left( Y \right) \right]$.
Consequently, the $\operatorname{CKE}$ can be used as a normalized independence criterion.
The squared Pearson correlation is recovered as a special case for $\mathcal{X} = \mathbb{R} = \mathcal{Y}$ and $k \left( x, y \right) = xy = \Tilde{k} \left( x, y \right)$.
We use the CKA in our contributions for a more fine-grained model evaluation via kernel spherical score in Section~\ref{sec:summary_neurips_ws_24}.

\subsection{Uncertainty Calibration of Predictions}
\label{sec:bg_cal}

When assessing model performance, risk minimization offers an approach to identify what given model is closest to the Bayes predictor, i.e. the ideal predictor \citep{bishop2006pattern}.
In a noisy world, even the ideal predictor may never precisely predict the target outcome with 100\% accuracy.
Especially in sensitive applications, it is therefore not only important to aim for predicting the right events but also to correctly assign probability mass to target outcomes.
This predicted probability mass, under the assumption it is approximately correct, may then be used as a quantity of uncertainty in practice for sensitive decision-making.
Model (uncertainty) calibration has the purpose of assessing the correctness of such predicted probabilities to increase the trustworthiness of model uncertainty \citep{vaicenavicius2019evaluating}.
Specifically, when the model predicts a probability for a given input, then the predicted probability should match the randomness of the target variable.
A more specific example in practice can be given via classification tasks, the most prominent task to consider calibration for.

\subsubsection{Classification}
\label{sec:bg_cal_classif}

Even though uncertainty calibration is the easiest to explain for classification, it is still a theoretically challenging concept, and it is often not straightforward to transfer the mathematical notation to intuitive understanding.
An illustrative example is the following.
Assume a classifier predicts the probability of a disease for each of 10,000 people.
Now, we select all people, for whom the prediction assumes a specific probability, for example, "disease=30\%".
For a calibrated classifier, the probability of the disease occurring among the \emph{selected} people would be 30\%.

In classification, when we assess the calibration of a classifier assigning probabilities to a finite set of distinct outcomes, we are interested in the following question, which generalizes the previous example.

\begin{displayquote}
Does the probability of observing the target given a prediction match the prediction?
\end{displayquote}

In the following, we express this more technically and rigorously along the statement
\begin{equation}
    \operatorname{Prob} \left( \operatorname{target} \mid \operatorname{pred} \right) = \operatorname{pred},
\end{equation}
which we refer to as the calibration condition.
It is important to note that this condition is easily confused with the stricter condition for a Bayes classifier, which would be
\begin{equation}
    \operatorname{Prob} \left( \operatorname{target} \mid \operatorname{input} \right) = \operatorname{pred} \left( \operatorname{input} \right).
\end{equation}
This condition is what we optimize for via risk minimization, and is substantially more difficult to achieve than the calibration condition.
On the other hand, the calibration condition allows us to interpret the predicted probabilities of the classifier but might be insufficient for decision-making when the probabilities are too uniform.

To put this into a mathematical frame, we require the following setup.
In classification, a model $f \colon \mathcal{X} \to \Delta^d$ maps inputs from a feature space $\mathcal{X}$ to the probability simplex $\Delta^d$ for $d$ number of classes.
More technically, we assume the input for the model is a random variable $X$ with outcomes in $\mathcal{X}$ and the target is a random variable $Y$ with outcomes in $\mathcal{Y} = \left\{1, \dots, d \right\}$, and that these random variables follow a joint distribution $\left(X, Y \right) \sim \mathbb{P}_{XY}$.
For easier notation, we treat the conditional target distribution $\mathbb{P}_{Y \mid X}$ as a function $\mathcal{X} \to \Delta^d$ and the marginal target distribution $\mathbb{P}_{Y}$ as a vector in $\Delta^d$.
We can now define the following.
\begin{definition}[\citep{kull2019beyond}]
\label{def:top_label_calib}
    A classifier $f$ is defined to be \textbf{top-label confidence calibrated} if and only if for all $p \in \left[ 0, 1 \right]$ it holds
    \begin{equation}
        \mathbb{P} \left( Y = \arg\max_j f_j \left( X \right) \mid \max_j f_j \left( X \right) = p \right) = p.
        \label{eq:top_label_calib}
    \end{equation}
\end{definition}
Calibration errors, which quantify to what extend Equation~\eqref{eq:top_label_calib} is violated, are often of the form $\mathbb{E} \left[ D \left( \max_j f_j \left( X \right), \mathbb{P} \left( Y = \arg\max_j f_j \left( X \right) \mid \max_j f_j \left( X \right) \right) \right) \right]$ with some divergence or distance measure $D \colon \mathbb{R}_{\geq 0} \times \mathbb{R}_{\geq 0} \to \mathbb{R}$ \citep{vaicenavicius2019evaluating}.

Definition~\ref{def:top_label_calib} is in current literature the most prominent approach to consider classifier calibration \citep{naeini2015, guo2017calibration, minderer2021revisiting, chang2024survey}.
However, in this thesis, we are interested in deriving quantities of model uncertainty via first principles, which are, according to the current state of the literature, not defined for top-label confidence calibration.
In contrast, it is possible to derive a more general definition of calibration from risk minimization based on proper scores according to Equation~\eqref{eq:risk_min}, which we state in the following.
\begin{definition}[\citep{vaicenavicius2019evaluating}]
\label{def:canonical_calib}
    A classifier $f$ is defined to be \textbf{canonically calibrated} if and only if for all $p \in \Delta^d$ and $i = 1, \dots, d$ it holds
    \begin{equation}
        \mathbb{P} \left( Y = i \mid f \left( X \right) = p \right) = p_i.
        \label{eq:canonical_calib}
    \end{equation}
\end{definition}
Canonical calibration is a stronger condition than top-label confidence calibration, and a model that satisfies canonical calibration also satisfies top-label confidence calibration but not vice versa \citep{vaicenavicius2019evaluating}.
Calibration errors for Definition~\ref{def:canonical_calib} can be derived according to the so-called calibration-sharpness decomposition of proper scores \citep{Br_cker_2009, kull2015novel}.
The decomposition splits up the expected risk based on a proper score $S$ of a classifier into three distinct terms according to
\begin{equation}
\label{eq:cal_sharp_decomp_classif}
    \mathbb{E} \left[ S \left( f \left( X \right), Y \right) \right] = \mathbb{E} \left[ D \left( f \left( X \right), \mathbb{P}_{Y \mid f \left( X \right)} \right) \right] - \mathbb{E} \left[ D \left( \mathbb{P}_{Y}, \mathbb{P}_{Y \mid f \left( X \right)} \right) \right] + G \left( \mathbb{P}_{Y} \right),
\end{equation}
where $D$ and $G$ are the associated divergence and negative entropy of the proper score.
The first term is referred to as the calibration term as it is an expected divergence of the calibration condition in Definition~\ref{def:canonical_calib} \citep{Br_cker_2009}.
The second term quantifies the independence of the random variables $f \left( X \right)$ and $Y$.
It may be seen as a general measure of mutual information and is coined as model sharpness \citep{huszar2013scoring}.
Last, the third term only depends on the marginal distribution of $Y$, and is, thus, unaffected by our modeling choices for $f$.

This decomposition was first shown for the Brier score and used to originally derive calibration \citep{ANewVectorPartitionoftheProbabilityScore}.
As part of our contributions, we show that the sharpness term is equal to an f-Divergence and implies information monotonicity in neural networks in Section~\ref{sec:summary_aistats_24} and Section~\ref{sec:aistats_24}.

\subsubsection{General Case}
\label{sec:bg_cal_general}

A generalization of canonical calibration is, from a mathematical perspective, straightforward.
However, the lack of probabilities for individual outcomes in the general case makes the concept of calibration more difficult to interpret.
Based on \cite{widmann2021calibration}, we use the following definition.
We assume the target random variable $Y$ has an associated measurable space $\left( \mathcal{Y}, \mathcal{F}_{\mathcal{Y}} \right)$ and a set $\mathcal{P}$ of distributions defined on this measurable space.
\begin{definition}
\label{def:canonical_cal_general}
    A model $f \colon \mathcal{X} \to \mathcal{P}$ is defined to be \textbf{canonically calibrated} if and only if for all $P \in \mathcal{P}$ and $A \in \mathcal{F}_{\mathcal{Y}}$ it holds
    \begin{equation}
        \mathbb{P} \left( Y \in A \mid f \left( X \right) = P \right) = P \left( A \right).
    \end{equation}
\end{definition}
The classification case is recovered for $\mathcal{P} = \Delta^d$ with $\mathcal{Y} = \left\{ 1, \dots, d \right\}$.

As part of our contributions in this thesis, we show in Section~\ref{sec:neurips_22} that the calibration-sharpness decomposition of Equation~\eqref{eq:cal_sharp_decomp_classif} also holds beyond classification.
Further, Definition~\ref{def:canonical_cal_general} occurs implicitly in the decomposition mentioned above.

\subsubsection{Calibration Error Estimators For Classification}
\label{sec:bg_cal_estimators_classif}

Defining calibration errors is straight-forward since any divergence or distance defined for distributions may be used to compare the expected divergence between the prediction $f \left( X \right)$ and the conditional target $\mathbb{P} \left( Y \mid f \left( X \right) \right)$.
However, evaluating calibration errors in practice is notoriously difficult.
The difficulty lies in the term $\mathbb{P} \left( Y \mid f \left( X \right) \right)$ since it is a conditional distribution for a non-discrete random variable $f \left( X \right)$.

We assume a dataset of i.i.d. samples $\left( X_1, Y_1 \right), \dots, \left( X_n, Y_n \right) \sim \mathbb{P}_{XY}$ to estimate the calibration of a given classifier $f \colon \mathcal{X} \to \Delta^d$.
The most common approach to estimate calibration errors based on scalar conditionals is using a binning scheme \citep{naeini2015, guo2017calibration, minderer2021revisiting, Detlefsen2022}. 
A prominent estimator is the so-called \emph{expected calibration error} (ECE), which is a binning-based estimator based on the absolute distance \citep{guo2017calibration}.
In essence, the conditional target distribution $\mathbb{P} \left( Y = \arg\max_i f_i \left( X \right) \mid \max_i f_i \left( X \right) \right)$ is estimated via a histogram binning scheme, which places all top-label confidence predictions into mutually distinct bins $B_m \coloneqq \left\{ i \mid \arg\max_j f_j \left( X_i \right) \in I_m \right\}$, $m=1, \dots M$, based on a partition $\overset{}{\bigcup}_m I_m = [0,1]$.
The corresponding $L^p$ estimator is given by
\begin{equation}
\begin{split}
\label{def:tcep_est}
    & \operatorname{TCE}_p^{\operatorname{bin}} \coloneqq \left( \sum_{m=1}^M \frac{\lvert B_m \rvert}{n} \left\lvert \operatorname{acc} \left( B_m \right) - \operatorname{conf} \left( B_m \right) \right\rvert^p \right)^{\frac{1}{p}}
\end{split}
\end{equation}
with $\operatorname{acc} \left( B \right) = \frac{1}{\lvert B \rvert} \sum_{i \in B} \mathbf{1}_{Y_i = \arg\max_j f_j \left( X_i \right)}$ and $\operatorname{conf} \left( B \right) = \frac{1}{\lvert B \rvert} \sum_{i \in B} \arg\max_j f_j \left( X_i \right)$ \citep{kumar2019verified}.
This estimator is primarily suitable for target distributions conditioned on a scalar random variable.
The choice of bin intervals $I_1, \dots, I_M$ is user-defined and the estimator only converges to $\operatorname{TCE}_p$ for an adaptive scheme \citep{vaicenavicius2019evaluating}.
\cite{patel2021multiclass} and \cite{roelofs2022mitigating} propose approaches to automatically select appropriate bins.

Estimating canonical calibration is more difficult than top-label confidence calibration due to the target distribution $\mathbb{P} \left( Y \mid f \left( X \right) \right)$ being conditioned on a vector-valued random variable.
\cite{popordanoska2022} propose to use a kernel density ratio approach based on the Nadaraya-Watson-estimator \citep{bierens1996topics}.
The estimator for the $L^p$ canonical calibration error $\operatorname{CCE}_p$ is given by
\begin{equation}
\label{def:cce_kde_est}
\begin{split}
    & \operatorname{CCE}_2^{\operatorname{kde}} \coloneqq \left(\frac{1}{n} \sum_{i=1}^n \left\lVert f \left( X_i \right) - \frac{\sum_j k_{\operatorname{dir}} \left( f \left( X_j \right); f \left( X_i \right) \right) e_{Y_j}}{\sum_j k_{\operatorname{dir}} \left( f \left( X_j \right); f \left( X_i \right) \right)} \right\rVert^p_p \right)^{\frac{1}{p}},
\end{split}
\end{equation}
where $e_i$ refers to the unit vector with a "$1$" at index $i$ and $k_{\operatorname{dir}}$ is chosen to be the Dirichlet kernel, which is specifically suited for the simplex space \citep{ouimet2022asymptotic, popordanoska2022}.

\subsection{Aleatoric and Epistemic Uncertainty}
\label{sec:bg_aleatoric_epistemic}

The uncertainty of a prediction refers to the randomness or lack of knowledge, we have about the predicted target variable \citep{hullermeier2021aleatoric}.
In this section, we give an overview of two common types of uncertainty.
Aleatoric uncertainty refers to the irreducible randomness in our target variable \citep{hullermeier2021aleatoric}.
An example is the randomness of a coin flip.
We may determine the probabilities for Heads and Tails but we may never be able to predict the outcome of the coin toss with 100\% accuracy.
Furthermore, epistemic uncertainty refers to the lack of knowledge in constructing a prediction due to limited data \citep{hullermeier2021aleatoric}.
In the coin flip example, this refers to the uncertainty in the predicted probabilities for head and tails due to the finite number of observations we base our prediction on.
Both types of uncertainties are inevitable in practice.
For aleatoric uncertainty, we assume there is a noise source involved in affecting the target outcome, i.e., the same input may have different expressions of the target outcome.
It is assumed that the aleatoric uncertainty of a predictive task is independent of the chosen model.
For epistemic uncertainty, we assume there are only finite observations of the real-world, which are by themselves noisy, and, thus, lead to an uncertain prediction relative to the amount of data.
Specifically, it is assumed that epistemic uncertainty vanishes the more data we collect.

To turn the previous example numeric, assume the coin flip is expressed via the random variable $Y$ with \emph{Heads} encoded as "$1$" and \emph{Tails} as "$0$".
Then, $Y$ is following a Bernoulli distribution with $\mathbb{P} \left( Y = 1 \right) = p$ for some $p \in \left[ 0, 1 \right]$.
Further, assume we have a dataset of i.i.d. random variables $Y_1, \dots, Y_n \sim \mathbb{P}_Y$, which are previous observations of the coin flip.
We use $\hat{p}_n \coloneqq \frac{1}{n} \sum_{i=1}^n Y_i$ as the prediction for the unknown $p$ and the expected mean-squared error to assess the prediction defined by
\begin{equation}
    \operatorname{Error} \left( \hat{p}_n \right) \coloneqq \mathbb{E} \left[ \left( \hat{p}_n - Y \right)^2 \right].
\end{equation}
Then, we can use the bias-variance decomposition \citep{hastie2009elements} giving
\begin{equation}
\begin{split}
    \operatorname{Error} \left( \hat{p}_n \right) & = \underbrace{\left( \mathbb{E} \left[ \hat{p}_n \right] - p \right)^2}_{\text{Bias}} + \underbrace{\operatorname{Var} \left( \hat{p}_n \right)}_{\text{Variance}} + \underbrace{p \left( 1 - p \right)}_{\text{Noise}}. \\
\end{split}
\end{equation}
The bias term is zero since $\mathbb{E} \left[ \hat{p}_n \right] = p$.
The variance term is equal to $\frac{1}{n} p \left( 1 - p \right)$, which means the variance term vanishes towards zero with an increasing number of observations $n$.
This matches the description of epistemic uncertainty.
The noise term is independent of our prediction and of any noise source unaffiliated with our target variable.
If $p$ goes towards one or zero, i.e., the outcome becomes less noisy, then the noise term also vanishes towards zero.
This matches the description of aleatoric uncertainty.
This example illustrates how the variance term can be assigned to epistemic uncertainty and the noise term to aleatoric uncertainty.
In the following of this thesis, we use these concepts to distinguish various types of uncertainty.
This is especially relevant when summarizing our contributions regarding bias-variance decompositions in Section~\ref{sec:summary_uncertainties_via_bvd}.
Note that the presented quantities for aleatoric and epistemic uncertainty based on the bias-variance decomposition are not unique approaches.
Alternatively, aleatoric and epistemic uncertainties are also present in Bayesian modeling \citep{kendall2017uncertainties, depeweg2018decomposition, abdar2021review}, in systems engineering \citep{der2009aleatory, aldemir2013survey}, or when a hypothesis class of models is formally introduced in the risk-assessment \citep{hullermeier2021aleatoric}.

\begin{figure}
    \centering
    \includegraphics[width=\columnwidth]{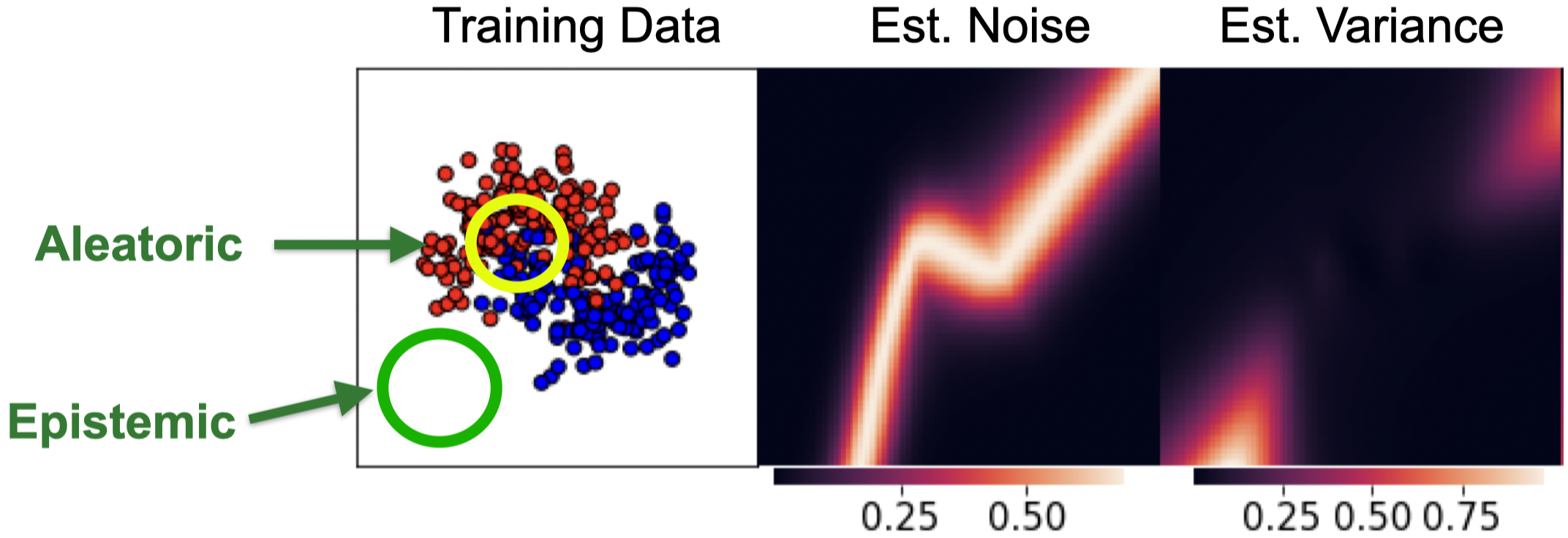}
\caption{
Illustration of aleatoric and epistemic uncertainty quantification in binary classification (red versus blue).
Aleatoric uncertainty refers to the irreducible noise in the data labeling process (areas where blue and red classes are overlapping).
Epistemic uncertainty refers to a lack of knowledge due to finite data (areas where no class instances are available).
The estimated noise and estimated variance of a classifier (here: neural network) offer practically feasible approximations.
}
\label{fig:basic_UQ_illus}
\end{figure}

In practice, we do not have access to the unknown ground truth value of $p$.
Consequently, we may use the prediction $\hat{p}_n$ instead as a workaround.
Following \cite{hullermeier2021aleatoric}, we can extend the concept from predicting a constant probability to binary classification by replacing $\hat{p}_n$ with a model $\hat{f} \colon \mathcal{X} \to \left[ 0, 1 \right]$ receiving inputs from a set $\mathcal{X}$ and trained on a dataset $\mathcal{D} \coloneqq \left\{ \left( X_1, Y_1 \right), \dots, \left( X_n, Y_n \right) \right\}$ of i.i.d. tuples of input-target pairs.
If we use the model predictions as estimates of aleatoric uncertainty and an ensemble of models trained on simulated datasets $\mathcal{D}_1, \dots, \mathcal{D}_n$, then we can estimate the aleatoric and epistemic uncertainties across the entire input space as in Figure~\ref{fig:basic_UQ_illus}.
However, for more complex prediction scenarios, it is not clear if an approximation is appropriate as it depends on the model, and we also have to approximate the variance term via an ensemble prediction.
Throughout our contributions, it is up to each application how to evaluate the performance of different approximations.

In consequence, we can see how the bias-variance decomposition of the mean-squared error offers a tool to assess statistically well-defined quantities of the more informal notions of aleatoric and epistemic uncertainty.
Other bias-variance decompositions also exist, e.g., 0-1 loss \citep{domingos2000unified}, or finite Bregman divergences \citep{pfau2013generalized}.
Based on these insights, we contribute a bias-variance decomposition of strictly proper scores to offer novel quantities of aleatoric and epistemic uncertainty.
Specifically, this approach reduces the search of aleatoric and epistemic uncertainty measures in theory to the search for a proper score and appropriate approximations in practice.
This is an integral concept of the framework for uncertainty quantification we discuss in this thesis.

\section{Uncertainties via Bias-Variance Decompositions}
\label{sec:summary_uncertainties_via_bvd}

In this section, we summarize our contributions regarding novel bias-variance decompositions presented in Chapter~\ref{ch:uncertainties_via_bvd}, which includes the works \emph{Uncertainty Estimates of Predictions via a General Bias-Variance Decomposition} \citep[Published at AISTATS]{gruber2023uncertainty} and \emph{A Bias-Variance-Covariance Decomposition of Kernel Scores for Generative Models} \citep[Published at ICML]{gruber2024biasvariancecovariance}.
As we have seen in Section~\ref{sec:bg_aleatoric_epistemic}, we can associate the components of the mean-squared error bias-variance decomposition with aleatoric and epistemic uncertainty.
We extend this concept by introducing a bias-variance decomposition for strictly proper scores, which we summarize in Section~\ref{sec:summary_aistats_23}.
The individual components will be represented via the proper score's associated entropy, Bregman divergence, and Bregman Information.
Further, in Section~\ref{sec:summary_icml_24} we discuss our contributions regarding a bias-variance-covariance decomposition of kernel scores, which arises when an ensemble of predictions has non-independent individual predictions.
Kernels are of particular practical interest since we can evaluate all derived quantities for sample-based generative models, like diffusion models or large language models, for which we offer state-of-the-art experimental results.

\subsection{Uncertainty Estimates of Predictions via a General Bias-Variance Decomposition}
\label{sec:summary_aistats_23}

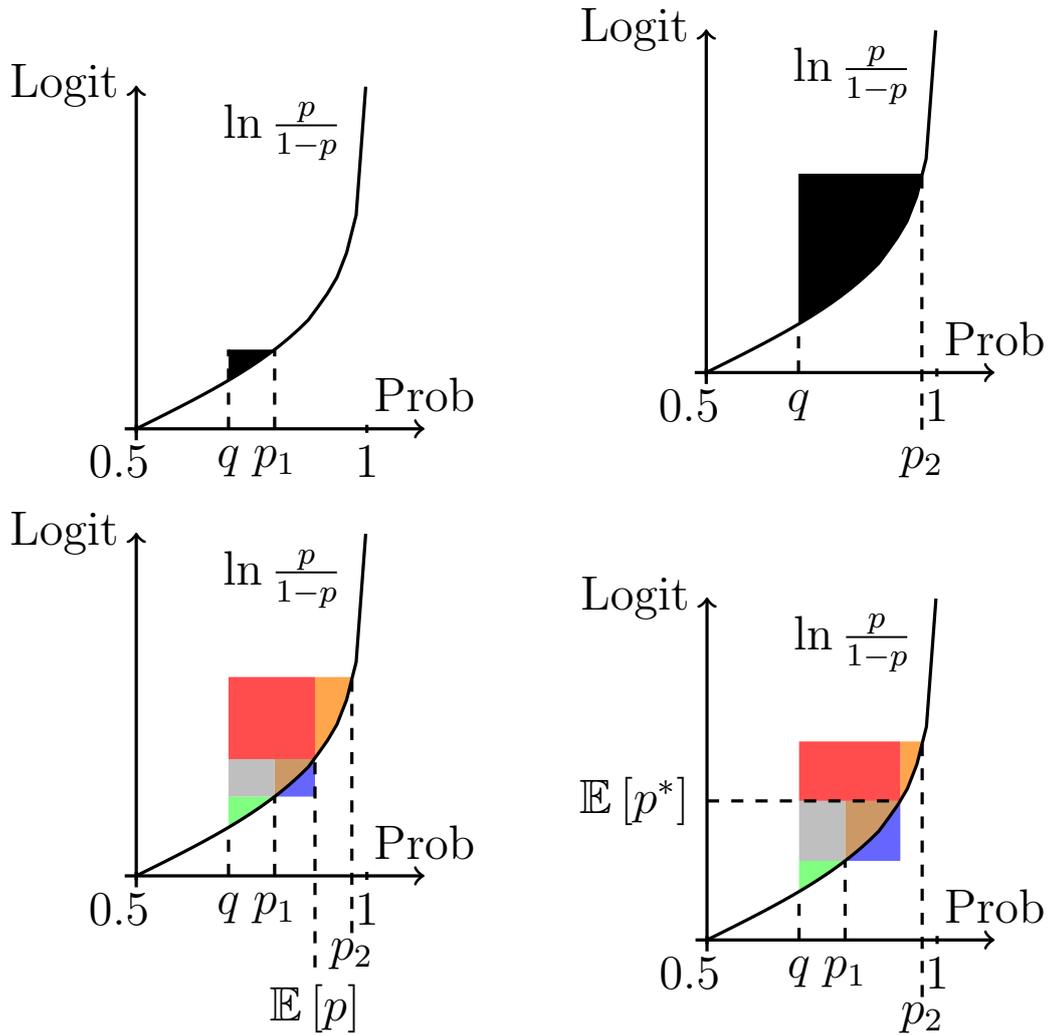
\begin{figure}
    \centering
    \begin{subfigure}{.5\textwidth}
    \centering
    \resizebox{0.9\columnwidth}{!}{%
    \begin{tikzpicture}
        \fill [black, domain=2.8:3.2, variable=\x]
          (2.8, {0.5*(ln(0.8)-ln(0.2))})
          -- plot ({\x}, {0.5*(ln(\x/4)-ln(1 - \x/4))})
          -- (3.2, {0.5*(ln(0.8)-ln(0.2))})
          -- cycle;
        \draw [thick] [->] (1.9,0)--(4.5,0) node[right, above] {Prob};
        \foreach \x in {0.5,1}
           \draw[xshift=4*\x cm, thick] (0pt,-1pt)--(0pt,1pt);
       \draw[thick] node[below] at (1.85,0) {$0.5$};
       \draw[thick] node[below] at (4,0) {$1$};
        
        \draw [thick] [->] (2,-0.1)--(2,3) node[above, left] {Logit};
        
        \draw [thick, domain=2:3.99, variable=\x]
          plot ({\x}, {0.5*(ln(\x/4) - ln(1 - \x/4))}) node[right] at (2.6,2.6) {$\ln \frac{p}{1-p}$};

        \draw[thick, black, dashed] (2.8, 0)--(2.8, {0.5*(ln(0.8)-ln(0.2))}) node[below] at (2.8, 0) {$q$};
        \draw[thick, black, dashed] (3.2, 0)--(3.2, {0.5*(ln(0.8)-ln(0.2))}) node[below] at (3.2, 0) {$p_1$};

    \end{tikzpicture}
    }
    \end{subfigure}%
        \begin{subfigure}{.5\textwidth}
    \centering
    \resizebox{0.9\columnwidth}{!}{%
    \begin{tikzpicture}
        \fill [black, domain=2.8:3.87, variable=\x]
          (2.8, {0.5*(ln(0.97)-ln(0.03))})
          -- plot ({\x}, {0.5*(ln(\x/4)-ln(1 - \x/4))})
          -- (3.87, {0.5*(ln(0.97)-ln(0.03))})
          -- cycle;
        \draw [thick] [->] (1.9,0)--(4.5,0) node[right, above] {Prob};
        \foreach \x in {0.5,1}
           \draw[xshift=4*\x cm, thick] (0pt,-1pt)--(0pt,1pt);
       \draw[thick] node[below] at (1.85,0) {$0.5$};
       \draw[thick] node[below] at (4,0) {$1$};
        
        \draw [thick] [->] (2,-0.1)--(2,3) node[above, left] {Logit};
        
        \draw [thick, domain=2:3.99, variable=\x]
          plot ({\x}, {0.5*(ln(\x/4) - ln(1 - \x/4))}) node[right] at (2.6,2.6) {$\ln \frac{p}{1-p}$};

        \draw[thick, black, dashed] (2.8, 0)--(2.8, {0.5*(ln(0.7)-ln(0.3))}) node[below] at (2.8, 0) {$q$};
        \draw[thick, black, dashed] (3.87, -0.5)--(3.87, {0.5*(ln(0.97)-ln(0.03))}) node[below] at (3.87, -0.5) {$p_2$};

    \end{tikzpicture}
    }
    \end{subfigure} \\
    \begin{subfigure}{.5\textwidth}
    \centering
    \resizebox{0.9\columnwidth}{!}{%
    \begin{tikzpicture}
        \fill [orange!70, domain=3.55:3.87, variable=\x]
          (3.55, {0.5*(ln(0.97)-ln(0.03))})
          -- plot ({\x}, {0.5*(ln(\x/4)-ln(1 - \x/4))})
          -- (3.87, {0.5*(ln(0.97)-ln(0.03))})
          -- cycle;
        \fill [blue!60, domain=3.2:3.55, variable=\x]
          (3.2, {0.5*(ln(0.8)-ln(0.2))})
          -- plot ({\x}, {0.5*(ln(\x/4)-ln(1 - \x/4))})
          -- (3.55, {0.5*(ln(0.8)-ln(0.2))})
          -- cycle;
        \fill [brown!80, domain=3.2:3.55, variable=\x]
          (3.2, {0.5*(ln(0.885)-ln(0.115))})
          -- plot ({\x}, {0.5*(ln(\x/4)-ln(1 - \x/4))})
          -- (3.55, {0.5*(ln(0.885)-ln(0.115))})
          -- cycle;
        \fill [red!70, domain=2.8:3.55, variable=\x]
          (2.8, {0.5*(ln(0.885)-ln(0.115))})
          -- plot ({\x}, {0.5*(ln(0.97)-ln(0.03))})
          -- (3.55, {0.5*(ln(0.885)-ln(0.115))})
          -- cycle;
        \fill [green!50, domain=2.8:3.2, variable=\x]
          (2.8, {0.5*(ln(0.8)-ln(0.2))})
          -- plot ({\x}, {0.5*(ln(\x/4)-ln(1 - \x/4))})
          -- (3.2, {0.5*(ln(0.8)-ln(0.2))})
          -- cycle;
        \fill [gray!50, domain=2.8:3.2, variable=\x]
          (2.8, {0.5*(ln(0.8)-ln(0.2))})
          -- plot ({\x}, {0.5*(ln(0.885)-ln(0.115))})
          -- (3.2, {0.5*(ln(0.8)-ln(0.2))})
          -- cycle;
        \draw [thick] [->] (1.9,0)--(4.5,0) node[right, above] {Prob};
        \foreach \x in {0.5,1}
           \draw[xshift=4*\x cm, thick] (0pt,-1pt)--(0pt,1pt);
       \draw[thick] node[below] at (1.85,0) {$0.5$};
       \draw[thick] node[below] at (4,0) {$1$};
        
        \draw [thick] [->] (2,-0.1)--(2,3) node[above, left] {Logit};
        
        \draw [thick, domain=2:3.99, variable=\x]
          plot ({\x}, {0.5*(ln(\x/4) - ln(1 - \x/4))}) node[right] at (2.6,2.6) {$\ln \frac{p}{1-p}$};

        \draw[thick, black, dashed] (2.8, 0)--(2.8, {0.5*(ln(0.7)-ln(0.3))}) node[below] at (2.8, 0) {$q$};
        \draw[thick, black, dashed] (3.55, -0.8)--(3.55, {0.5*(ln(0.885)-ln(0.115))}) node[below] at (3.55, -0.8) {$\mathbb{E} \left[ p \right]$};
        \draw[thick, black, dashed] (3.2, 0)--(3.2, {0.5*(ln(0.8)-ln(0.2))}) node[below] at (3.2, 0) {$p_1$};
        \draw[thick, black, dashed] (3.87, -0.5)--(3.87, {0.5*(ln(0.97)-ln(0.03))}) node[below] at (3.87, -0.4) {$p_2$};

    \end{tikzpicture}
    }
    \end{subfigure}%
    \begin{subfigure}{.5\textwidth}
    \centering
    \resizebox{0.9\columnwidth}{!}{%
    \begin{tikzpicture}
        \fill [orange!70, domain=3.68:3.87, variable=\x]
          (3.68, {0.5*(ln(0.97)-ln(0.03))})
          -- plot ({\x}, {0.5*(ln(\x/4)-ln(1 - \x/4))})
          -- (3.87, {0.5*(ln(0.97)-ln(0.03))})
          -- cycle;
        \fill [blue!60, domain=3.2:3.68, variable=\x]
          (3.2, {0.5*(ln(0.8)-ln(0.2))})
          -- plot ({\x}, {0.5*(ln(\x/4)-ln(1 - \x/4))})
          -- (3.68, {0.5*(ln(0.8)-ln(0.2))})
          -- cycle;
        \fill [brown!80, domain=3.2:3.68, variable=\x]
          (3.2, {0.5*(ln(0.92)-ln(0.08))})
          -- plot ({\x}, {0.5*(ln(\x/4)-ln(1 - \x/4))})
          -- (3.68, {0.5*(ln(0.92)-ln(0.08))})
          -- cycle;
        \fill [red!70, domain=2.8:3.68, variable=\x]
          (2.8, {0.5*(ln(0.92)-ln(0.08))})
          -- plot ({\x}, {0.5*(ln(0.97)-ln(0.03))})
          -- (3.68, {0.5*(ln(0.92)-ln(0.08))})
          -- cycle;
        \fill [green!50, domain=2.8:3.2, variable=\x]
          (2.8, {0.5*(ln(0.8)-ln(0.2))})
          -- plot ({\x}, {0.5*(ln(\x/4)-ln(1 - \x/4))})
          -- (3.2, {0.5*(ln(0.8)-ln(0.2))})
          -- cycle;
        \fill [gray!50, domain=2.8:3.2, variable=\x]
          (2.8, {0.5*(ln(0.8)-ln(0.2))})
          -- plot ({\x}, {0.5*(ln(0.92)-ln(0.08))})
          -- (3.2, {0.5*(ln(0.8)-ln(0.2))})
          -- cycle;
        \draw [thick] [->] (1.9,0)--(4.5,0) node[right, above] {Prob};
        \foreach \x in {0.5,1}
           \draw[xshift=4*\x cm, thick] (0pt,-1pt)--(0pt,1pt);
       \draw[thick] node[below] at (1.85,0) {$0.5$};
       \draw[thick] node[below] at (4,0) {$1$};
        
        \draw [thick] [->] (2,-0.1)--(2,3) node[above, left] {Logit};
        
        \draw [thick, domain=2:3.99, variable=\x]
          plot ({\x}, {0.5*(ln(\x/4) - ln(1 - \x/4))}) node[right] at (2.6,2.6) {$\ln \frac{p}{1-p}$};

        \draw[thick, black, dashed] (2.8, 0)--(2.8, {0.5*(ln(0.7)-ln(0.3))}) node[below] at (2.8, 0) {$q$};
        \draw[thick, black, dashed] (2, {0.5*(ln(0.92)-ln(0.08))})--(3.68, {0.5*(ln(0.92)-ln(0.08))}) node[left] at (2, {0.5*(ln(0.92)-ln(0.08))}) {$\mathbb{E} \left[ p^* \right]$};
        \draw[thick, black, dashed] (3.2, 0)--(3.2, {0.5*(ln(0.8)-ln(0.2))}) node[below] at (3.2, 0) {$p_1$};
        \draw[thick, black, dashed] (3.87, -0.5)--(3.87, {0.5*(ln(0.97)-ln(0.03))}) node[below] at (3.87, -0.4) {$p_2$};
    \end{tikzpicture}
    }
    \end{subfigure}
\caption{
The geometry of the bias-variance decomposition for the Kullback-Leibler divergence in binary classification. \textbf{Top:} The KL divergences of predictions $p_1$ and $p_2$ given $q$ are the black areas. A bias-variance decomposition requires a mean prediction $\bar{p}$ such that the average of the black areas is equal to areas, which a) only depend on $\bar{p}$ and $q$ (bias), or b) only depend on $\bar{p}$ and $p_1$ or $p_2$ (variance).
\textbf{Bottom-Left:} If we pick $\bar{p} = \mathbb{E} \left[ p \right] = \frac{1}{2} p_1 + \frac{1}{2} p_2$, we have $\text{red} \neq \text{grey} + \text{brown} + \text{blue}$. Thus, we cannot represent the size of the red area via other areas and cannot remove it. However, we need to remove it since it depends on $q$ \emph{and} $p_2$.
\textbf{Bottom-Right:} If we pick $\bar{p} = \mathbb{E} \left[ p^* \right] = \frac{1}{2} \ln \frac{p_1}{1-p_1} + \frac{1}{2} \ln \frac{p_2}{1-p_2}$, we have $\text{red} = \text{grey} + \text{brown} + \text{blue}$. This allows us to represent the size of the red area by adding the blue area. The bias term is then $\text{green} + \text{grey} + \text{brown}$, which does in its sum not depend on $p_1$ or $p_2$. And the variance term is $\frac{1}{2} \text{blue} + \frac{1}{2} \text{orange}$, which does not depend on $q$.
}
\label{fig:bvd_visual}
\end{figure}

In this section, we highlight the core contributions of Section~\ref{sec:aistats_23}, which incorporates \emph{Uncertainty Estimates of Predictions via a General Bias-Variance Decomposition} \citep[Published at AISTATS]{gruber2023uncertainty}, and embed these into the here presented framework by creating a link with aleatoric and epistemic uncertainty.

The work tackles the challenge of reliably estimating the uncertainty of a prediction in machine learning models, which is particularly motivated for safety-critical applications.
A common way to measure uncertainty in classification is through predicted confidence, but this can be unreliable under domain drift and is limited to classification tasks.
Here, we offer an alternative for measuring uncertainty across various predictive tasks via a bias-variance decomposition for proper scores, which gives measures of uncertainty as discussed in Section~\ref{sec:bg_aleatoric_epistemic}.
Specifically, we introduce a general bias-variance decomposition for strictly proper scores, identifying the Bregman Information as the variance term and the entropy function as the noise term.
To derive the decomposition, we generalize relevant properties of Bregman divergences presented in Section~\ref{sec:bg_breg_div} to functional Bregman divergences.
This includes a generalization of Lemma~\ref{le:breg_arg_flip} and Lemma~\ref{le:breg_decomp} to the functional case.
To highlight the core contribution, we repeat it in the following.
Assume a strictly proper score $S \colon \mathcal{P} \times \mathcal{Y} \to \mathbb{R}$ based on a convex set of distributions $\mathcal{P}$ defined on a targeted sample set $\mathcal{Y}$.
Further, let $\hat{P}$ be a random variable, representing a noisy prediction, with outcomes in $\mathcal{P}$, and $Y \sim Q$ the target random variable with outcomes in $\mathcal{Y}$ and target distribution $Q \in \mathcal{P}$.
We also need the typical assumption that $\hat{P}$ and $Y$ are independent \citep{hastie2009elements}.
This assumption is not problematic in practice since any possible input for the prediction is considered to be a constant.
Denote with $H \colon \mathcal{P} \to \mathbb{R}$ the associated entropy, and define the subgradient function $\mathbf{S} \colon \mathcal{P} \to \mathcal{L} \left( \mathcal{P} \right)$ of $-H$ based on $S$ by $\mathbf{S} \left( P \right) \left( y \right) \coloneqq -S \left( P, y \right)$ for $P \in \mathcal{P}$ and $y \in \mathcal{Y}$.
Then, in Section~\ref{sec:aistats_23} we state that
\begin{equation}
    \underbrace{\mathbb{E} \left[ S \left( \hat{P}, Y \right) \right]}_{\text{Generalization Error}} = \underbrace{D_{G^*,\mathbf{S}^{-1}} \left( \mathbb{E} \left[ \mathbf{S} \left( \hat{P} \right) \right], \mathbf{S} \left( Q \right) \right)}_{\text{Bias}} + \underbrace{\mathbb{B}_{G^*} \left[ \mathbf{S} \left( \hat{P} \right) \right]}_{\text{Variance}} + \underbrace{H \left( Q \right)}_{\text{Noise}}.
\label{eq:general_bvd}
\end{equation}
Note that the inverse $\mathbf{S}^{-1}$ can always be defined since by assumption $S$ is strictly proper, which makes $-H$ strictly convex and $\mathbf{S}$ injective (c.f. Section~\ref{sec:aistats_23} for an extensive discussion).
A geometric visualization of why the bias and variance are based on the mean in the dual space $\mathcal{L} \left( P \right)$ is given in Figure~\ref{fig:bvd_visual} for the Kullback-Leibler divergence, which is equal to the expected log score plus the noise term.
As discussed in Section~\ref{sec:bg_breg_div}, the (functional) Bregman Information generalizes the variance of a random variable based on a generating function.
We can recover the mean-squared error bias-variance decomposition via the Brier score and $\mathcal{P} = \Delta^d$, since then we have $G \left( P \right) = \sum_{i=1}^d P_i^2 = G^* \left( P \right)$ with $\mathbf{S} \left( P \right) = P$.
Thus, the Bregman Information provides a principled way to analyze the generalized variance of a predictor.
The decomposition generalizes existing decompositions in the literature, for example \citep{pfau2013generalized} and \citep{hansen2000general}, and provides closed-form solutions for specific cases.
As special cases, we establish new formulations of the bias-variance decomposition for the log-likelihood of exponential families and the classification log-likelihood.
Specifically, in these cases the dual space is reduced to a finite subspace of $\mathbb{R}^d$, further simplifying the expressions.
Further, we highlight that the Bregman Information generated by the log score measures model variability in the logit space, which is convenient for deep learning applications, as it avoids the need for transforming logit predictions.
We also generalize the law of total variance to Bregman Information, which offers theoretical performance guarantees for ensemble predictions, and gives a general approach for constructing confidence regions for predictions based on the Bregman Information.
Given the background in Section~\ref{sec:bg_aleatoric_epistemic}, the ensemble-based uncertainty estimations are types of epistemic uncertainty.
This can be seen via the properties of Bregman Information presented in Proposition~3.8 of Section~\ref{sec:aistats_23}, which shows when the Bregman Information diminishes for growing ensemble sizes.
If we associate an ensemble size with available data size, more data translates to more ensemble members, reducing the Bregman Information towards zero.
This property of Bregman Information matches precisely the requirements for epistemic uncertainty described in Section~\ref{sec:bg_aleatoric_epistemic}.
Equation~\eqref{eq:general_bvd} allows us to associate the notion of aleatoric and epistemic uncertainty with the noise and variance terms as in Section~\ref{sec:bg_aleatoric_epistemic}.
Specifically, we may use the predictive entropy $H \left( \hat{P} \right)$ as an estimate of aleatoric uncertainty, and the empirical Bregman Information 
\begin{equation}
    \hat{\mathbb{B}}_{G^*} \coloneqq \frac{1}{m} \sum_{i=1}^m G^* \left( \mathbf{S} \left( \hat{P}_i \right) \right) - G^* \left( \frac{1}{m} \sum_{i=1}^m \mathbf{S} \left( \hat{P}_i \right) \right)
\end{equation}
for an ensemble of $m$ predictions $\hat{P}_1, \dots, \hat{P}_m$ as an estimate of epistemic uncertainty.
In Figure~\ref{fig:basic_UQ_illus}, we plot the predictive entropy (middle) and the empirical Bregman Information (right) associated with the log score of a neural network classifier on a toy task (c.f. Section~\ref{sec:aistats_23} for details).
In practice, we propose to use Deep Ensembles \citep{lakshminarayanan2017simple}, test-time augmentation \citep{ayhan2018test}, Monte-Carlo Dropout \citep{gal2016dropout}, and bootstrapping \citep{efron1994introduction} to approximate the Bregman Information.
\cite{kahl2024values} show that test-time augmentation approximates a type of epistemic uncertainty, which is consistent with our approach.

The confidence score $\operatorname{CS} \colon \Delta^d \to [0,1]$ of a predicted probability vector $P \in \Delta^d$ is the predicted probability of the predicted class label defined as $\operatorname{CS} \left( P \right) \coloneqq \max_i P_i$.
Confidence scores are connected to the negative Shannon entropy \\$G_{\log} \left( P \right) \coloneqq \sum_{i=1}^d P_i \log P_i$ of the predicted probability vector since both assign their maximum and minimum value to the same probability vectors, i.e., for all $P\in \Delta^d$ holds 
\begin{equation}
    \operatorname{CS} \left( P \right) = \max_{Q \in \Delta^d} \operatorname{CS} \left( Q \right) \iff G_{\log} \left( P \right) = \max_{Q \in \Delta^d} G_{\log} \left( Q \right)
\end{equation}
as well as 
\begin{equation}
    \operatorname{CS} \left( P \right) = \min_{Q \in \Delta^d} \operatorname{CS} \left( Q \right) \iff G_{\log} \left( P \right) = \min_{Q \in \Delta^d} G_{\log} \left( Q \right).
\end{equation}
Consequently, confidence scores of individual predictions can be interpreted as a measure of aleatoric uncertainty similar to the Shannon entropy.
However, unlike entropy functions, confidence scores are not connected to the bias-variance decomposition.
In Section~\ref{sec:aistats_23}, we show that the Bregman Information can be a more meaningful measure of out-of-domain uncertainty compared to confidence scores, particularly in cases of corrupted data for neural networks in image classification.
Specifically, the Bregman Information is utilized for out-of-distribution detection, showing robustness to different degrees of domain drift.
This highlights the requirement of having a nuanced understanding of various measures of uncertainty, and to classify them into measures of aleatoric or epistemic uncertainty.
As a rule of thumb, within our framework empirical estimates of aleatoric uncertainty are usually based on individual probabilistic predictions of a model, while empirical estimates of epistemic uncertainty usually require an ensemble of predictions.

The previously discussed theoretical contribution is abstract and general, yet, we only offer empirical evaluation for classification tasks in Section~\ref{sec:aistats_23}.
In the next section, we focus on a subclass of proper scores, introduced as kernel scores in Section~\ref{sec:bg_ps_kernel_examples}, which results in an entropy and bias-variance decomposition empirically applicable to sample-based generative models.

\subsection{A Bias-Variance-Covariance Decomposition of Kernel Scores for Generative Models}
\label{sec:summary_icml_24}


In this section, we summarize the contributions of the work \emph{A Bias-Variance-Covariance Decomposition of Kernel Scores for Generative Models} \citep[Published at ICML]{gruber2024biasvariancecovariance}, which is located in Section~\ref{sec:icml_24}.
The backbone of the following contributions are kernels, which allow quantifying differences between distributions only based on their samples without requiring access to these distributions.
The major theoretical contributions here are twofold.
First, we prove a bias-variance-covariance decomposition for kernel scores, which solves important limitations of the more general decomposition in the last section.
Specifically, the decomposition holds for kernel scores that are non-strictly proper, i.e., their minimizer is not required to be unique.
Another solved limitation is that non-independent predictions within an ensemble allow a bias-variance-covariance decomposition, which is in general not possible for other proper scores.
Second, we formulate estimators of all involved quantities, which do not require explicit distributions and allow implicit distributions of sample-based generative models.
Such models have gained significant traction in recent years due to major advances in model architectures and training procedures.
Examples are diffusion models \citep{ho2020denoising} and large language models \citep{openai2023gpt4, touvron2023llama, team2023gemini}.

We state the decomposition according to the RKHS $\mathcal{H}$ of a p.s.d. kernel $k$ similar to Section~\ref{sec:icml_24} in the following.
We use the same notation and assumptions as in Section~\ref{sec:bg_ps_kernels} and Section~\ref{sec:aistats_23}.
The bias-variance decomposition for a kernel score $S_k$ is given by
\begin{equation}
    \underbrace{\mathbb{E} \left[ S_k \left( \hat{P}, Y \right) \right]}_{\text{Generalization Error}} = \underbrace{\left\lVert \mathbb{E} \left[ \mu_{\hat{P}} \right] - \mu_Q \right\rVert_{\mathcal{H}}^2}_{\text{Bias}} + \underbrace{\mathbb{E} \left[ \left\lVert \mu_{\hat{P}} - \mathbb{E} \left[ \mu_{\hat{P}} \right] \right\rVert_{\mathcal{H}}^2 \right]}_{\text{Variance}} \underbrace{ - \left\lVert \mu_Q \right\rVert_{\mathcal{H}}^2}_{\text{Noise}}.
\label{eq:ks_bvd}
\end{equation}
A covariance term appears if we assume $\hat{P}$ is the mean of non-independent predictions (c.f. Section~\ref{sec:icml_24}).
We can show that Equation~\eqref{eq:ks_bvd} is a special case of Equation~\eqref{eq:general_bvd}, which we omitted in the original work presented in Section~\ref{sec:icml_24}.
For completeness in the context of this thesis, we offer proof in the following.
First, note that the associated negative entropy of the kernel score $S_k$ is given by $G_k \left( P \right) = \left\lVert \mu_P \right\rVert_{\mathcal{H}}^2$ with subgradient $\mathbf{S}_k \left( P \right) \left( y \right) = 2 \left\langle \mu_P, \phi \left( y \right) \right\rangle_{\mathcal{H}} - \left\lVert \mu_P \right\rVert_{\mathcal{H}}^2$.
Then, we have for the convex conjugate that
\begin{equation}
    G_k^* \left( \mathbf{S}_k \left( Q \right) \right) = \sup_{P \in \mathcal{P}} 2 \left\langle \mu_Q, \mu_P \right\rangle_{\mathcal{H}} - \left\lVert \mu_Q \right\rVert_{\mathcal{H}}^2 - \left\lVert \mu_P \right\rVert_{\mathcal{H}}^2 = 0
\end{equation}
for all $Q \in \mathcal{P}$, and
\begin{equation}
\begin{split}
    G_k^* \left( \mathbb{E} \left[ \mathbf{S}_k \left( \hat{P} \right) \right] \right) & = \sup_{P \in \mathcal{P}} 2 \left\langle \mathbb{E} \left[ \mu_{\hat{P}} \right], \mu_P \right\rangle_{\mathcal{H}} - \mathbb{E} \left[ \left\lVert \mu_{\hat{P}} \right\rVert_{\mathcal{H}}^2 \right] - \left\lVert \mu_P \right\rVert_{\mathcal{H}}^2 \\
    & = \left\lVert \mathbb{E} \left[ \mu_{\hat{P}} \right] \right\rVert_{\mathcal{H}}^2 - \mathbb{E} \left[ \left\lVert \mu_{\hat{P}} \right\rVert_{\mathcal{H}}^2 \right],
\end{split}
\end{equation}
since $\left\langle \mu_P, \mu_{.} \right\rangle_{\mathcal{H}}$ is another subgradient of $\left\lVert \mu_P \right\rVert_{\mathcal{H}}^2$.
It also holds that
\begin{equation}
\begin{split}
    & \left\langle Q, \mathbb{E} \left[ \mathbf{S}_k \left( P \right) \right] - \mathbf{S}_k \left( Q \right) \right\rangle_{V^*} \\
    & = 2 \left\langle \mu_Q, \mathbb{E} \left[ \mu_{\hat{P}} \right] - \mu_Q \right\rangle_{\mathcal{H}} - \mathbb{E} \left[ \left\lVert \mu_{\hat{P}} \right\rVert_{\mathcal{H}}^2 \right] + \left\lVert \mu_Q \right\rVert_{\mathcal{H}}^2.
\end{split}
\end{equation}
Using these equations with Definition~\ref{def:breg_div} of Bregman divergences it follows for the bias term in Equation~\eqref{eq:general_bvd} that
\begin{equation}
\begin{split}
\label{eq:ks_bvd_bias}
    & D_{G^*,\mathbf{S}_k^{-1}} \left( \mathbf{S}_k \left( Q \right), \mathbb{E} \left[ \mathbf{S}_k \left( P \right) \right] \right) \\
    & = \left\lVert \mathbb{E} \left[ \mu_{\hat{P}} \right] \right\rVert_{\mathcal{H}}^2 - \mathbb{E} \left[ \left\lVert \mu_{\hat{P}} \right\rVert_{\mathcal{H}}^2 \right] - 2 \left\langle \mu_Q, \mathbb{E} \left[ \mu_{\hat{P}} \right] - \mu_Q \right\rangle_{\mathcal{H}} + \mathbb{E} \left[ \left\lVert \mu_{\hat{P}} \right\rVert_{\mathcal{H}}^2 \right] - \left\lVert \mu_Q \right\rVert_{\mathcal{H}}^2 \\
    & = \left\lVert \mu_P - \mu_Q \right\rVert_{\mathcal{H}}^2. \\
\end{split}
\end{equation}
Further, we get for the Bregman Information in Equation~\eqref{eq:general_bvd} that
\begin{equation}
    \mathbb{B}_{G^*} \left[ \mathbf{S}_k \left( \hat{P} \right) \right] = \mathbb{E} \left[ \left\lVert \mu_{\hat{P}} \right\rVert_{\mathcal{H}}^2 \right] - \left\lVert \mathbb{E} \left[ \mu_{\hat{P}} \right] \right\rVert_{\mathcal{H}}^2 = \mathbb{E} \left[ \left\lVert \mu_{\hat{P}} - \mathbb{E} \left[ \mu_{\hat{P}} \right] \right\rVert_{\mathcal{H}^2} \right].
\label{eq:ks_bvd_var}
\end{equation}
Together, Equation~\eqref{eq:ks_bvd_bias} and Equation~\eqref{eq:ks_bvd_bias} prove the connection between the decompositions in Equation~\eqref{eq:ks_bvd} and Equation~\eqref{eq:general_bvd}.

Consequently, we can apply the concepts of aleatoric and epistemic uncertainty as discussed in the last section here as well.
The decomposition in Equation~\eqref{eq:ks_bvd} provides a theoretical framework to understand the generalization behavior and uncertainty of generative models and their ensembles.
This presents the first extension of the bias-variance-covariance decomposition beyond the mean squared error introduced in \citep{ueda1996generalization}.
We also derive estimators for the kernel-based variance and entropy for uncertainty estimation, which can be estimated based on i.i.d. samples $\hat{Y}_1, \dots, \hat{Y}_m \sim \hat{P}$ of the implicitly predicted distribution $\hat{P}$.
We prove that the estimators are unbiased and consistent, and specify their convergence rate.
This approach requires only generated samples, not the underlying model itself, making it applicable to closed-source models, like OpenAI's GPT4 \citep{openai2023gpt4} or Google's Gemini \citep{team2023gemini}.
In summary, this results in measures of aleatoric and epistemic uncertainty, which can be estimated in a sample-only setting.

To demonstrate the wide applicability of the decomposition, we evaluate the related quantities on image, audio, and language generation tasks.
For image generation, we use diffusion models \citep{ho2020denoising} trained on synthetic MNIST images \citep{loosli-canu-bottou-2006}, for audio generation generative flow models \citep{kim2020glow} trained on the text-to-speech dataset LJSpeech \citep{ljspeech17}, and for language generation open pretrained transformer models \citep{zhang2022opt} performing question-answering tasks in CoQA \citep{reddy-etal-2019-coqa} and TriviaQA \citep{joshi2017triviaqa}.
We examine the generalization behavior of these models and investigate how bias, variance, and kernel entropy relate to the generalization error.
For example, we discover that the mode collapse of underrepresented minority classes may be expressed purely in the bias.
We require a measure of dependence to quantify how strongly entropy and variance are related to the generalization error.
A strong relation indicates an effective measure of uncertainty.
For the image and audio generation tasks, we use the kernel score as the generalization error, which has a continuous scale for typical kernel choices, like the RBF or Laplacian kernel.
Consequently, we use the Pearson correlation to assess how well entropy and variance predict the kernel score.
We discover a strong absolute correlation between entropy and kernel score ($\geq 0.9$) and a less strong correlation for the variance.
For language generation, we use a binary error based on the lexical similarity for generalization behavior according to \citep{kuhn2022semantic}.
\cite{kuhn2022semantic} proceed to use the area under receiver operator characteristic (AUROC) to quantify the relation between error and uncertainty measure, which we also adopt in our evaluations.
Further, for language generation, we use a pretrained embedding model, which maps text into a semantic vector space which we then use as kernel input.
The kernel-based entropy, which we interpret as estimated aleatoric uncertainty based on Section~\ref{sec:bg_aleatoric_epistemic}, combined with such embedding models shows superior performance in predicting the correctness of large language models for question-answering tasks.
Specifically, our approach outperforms the state-of-the-art baselines across almost all settings.

\medskip

In conclusion, the proposed bias-variance decomposition for proper scores offers a novel approach to evaluate the generalization behavior of models across a variety if tasks.
Specifically, the associated predictive entropy presents a viable approach for estimating aleatoric uncertainty and the associated variance of a predictive ensemble to estimate the epistemic uncertainty.
We successfully translate this result to kernel scores, which allows the evaluation of sample-based generative models for better uncertainty quantification in practice.
While aleatoric and epistemic uncertainty are quantities we may understand on the instance level, calibration represents how well the estimated aleatoric uncertainty matches the ground-truth target on a dataset level, as we will see in the next section.

\section{Improving Uncertainty Estimates via Calibration}
\label{sec:summary_uncertainty_cal_sharp}

In this section, we summarize our contributions regarding calibration of model uncertainties presented in Chapter~\ref{ch:uncertainties_via_cal_sharp}, which includes the works \emph{Better Uncertainty Calibration via Proper Scores for Classification and Beyond} \citep[Published at NeurIPS]{gruber2022better}, \emph{Consistent and
Asymptotically Unbiased Estimation of Proper Calibration Errors} \citep[Published at AISTATS]{gruber2024consistent}, and \emph{Optimizing Estimators of Squared Calibration Errors in Classification} \citep[In Submission at TMLR]{gruber2024optimizingestimatorssquaredcalibration}.
In the context of probability forecasts, calibration addresses the issue of quantifying the correctness of predictive uncertainty.
For example, overconfidence in modern neural networks is a common problem that can lead to misleading uncertainties.
Our works contribute to the broader goal of making machine learning models more trustworthy and reliable, which is crucial for their deployment in real-world applications
The findings have practical implications for practitioners working in sensitive domains where accurate uncertainty quantification is essential.
To relate the contributions summarized in this section to uncertainty quantification, as we introduced in the previous parts of this thesis, we require the following result.
Note that aleatoric uncertainty in our framework is a proper score's entropy function of the target distribution, i.e., let $H$ be an entropy function and $\mathbb{P}_{Y \mid X=x}$ the target distribution for a given feature random variable $X$ expressed as value $x \in \mathcal{X}$ for an input set $\mathcal{X}$.
Then, the quantity $H \left( \mathbb{P}_{Y \mid X = x} \right)$ represents the noise term in the respective bias-variance decomposition and a measure of aleatoric uncertainty.
However, we do not have access to $\mathbb{P}_{Y \mid X = x}$ in practice but have a prediction of a model $f \colon \mathcal{X} \to \mathcal{P}$ for a set $\mathcal{P}$ of possible target distributions.
Consequently, the quantity $H \left( f \left( x \right) \right)$ is what may be used in practice to estimate the aleatoric uncertainty $H \left( \mathbb{P}_{Y \mid X = x} \right)$, as discussed in Section~\ref{sec:summary_uncertainties_via_bvd}.
Now, if the model $f$ is canonically calibrated, we can make the statement that whenever it predicts a value $p \in \mathcal{P}$, the target distribution $\mathbb{P}_{Y \mid f \left( X \right) = p}$ conditional on this prediction is the same.
To connect calibration with aleatoric uncertainty estimation, we may ask ourselves how a predicted aleatoric uncertainty value $H \left( p \right)$ is connected to the unknown ground-truth aleatoric uncertainty given this prediction $H \left( \mathbb{P}_{Y \mid H \left( f \left( X \right) \right) = H \left( p \right)} \right)$.
This question is answered in the following theorem.
\begin{theorem}
\label{th:connection_aleatoric_calibration}
    If a model $f \colon \mathcal{X} \to \mathcal{P}$ is canonically calibrated with respect to an input variable $X$ and a target variable $Y$, then for any concave $H \colon \mathcal{P} \to \mathbb{R}$ and $p \in \mathcal{P}$ it holds
    \begin{equation}
        H \left( p \right) \leq H \left( \mathbb{P}_{Y \mid H \left( f \left( X \right) \right) = H \left( p \right)} \right).
    \end{equation}
\end{theorem}
The proof is fairly short and presented in the following.
It uses Jensen's inequality and integration rules.
\begin{proof}
We write $P = f \left( X \right)$ for shorter expressions. Starting from the definition of canonical calibration in Equation~\eqref{def:canonical_cal_general}, for all $p \in \mathcal{P}$ it holds
\begin{equation}
\begin{split}
    & p = \mathbb{P}_{Y \mid P=p} = \mathbb{P}_{Y \mid P=p, H \left( P \right)=H \left( p \right)} \\
    & \implies \mathbb{E} \left[ P \mid H \left( P \right) = H \left( p \right) \right] = \mathbb{P}_{Y \mid H \left( P \right)=H \left( p \right)} \\
    & \implies H \left( \mathbb{E} \left[ P \mid H \left( P \right) = H \left( p \right) \right] \right) = H \left( \mathbb{P}_{Y \mid H \left( P \right)=H \left( p \right)} \right) \\
    \overset{H\text{ concave}}&{\implies} \mathbb{E} \left[ H \left( P \right) \mid H \left( P \right) = H \left( p \right) \right] \leq H \left( \mathbb{P}_{Y \mid H \left( P \right)=H \left( p \right)} \right) \\
    & \implies H \left( p \right) \leq H \left( \mathbb{P}_{Y \mid H \left( P \right)=H \left( p \right)} \right).
\end{split}
\end{equation}
\end{proof}

In other terms, under the assumption that our model is (canonically) calibrated, Theorem~\ref{th:connection_aleatoric_calibration} states that whenever our model predicts an aleatoric uncertainty the ground truth aleatoric uncertainty of our target distribution given this prediction is equal or larger.
Initially, this might be unsatisfying since the original calibration property is an equality, however, we emphasize that this statement holds for any choice of entropy function $H$ and does not require further recalibration.
Further, the inequality in the presented direction allows a conservative risk-mitigation approach in practice.
Whenever we design an uncertainty threshold and our calibrated model exceeds this threshold, then Theorem~\ref{th:connection_aleatoric_calibration} states that the unknown ground truth aleatoric uncertainty also exceeds the threshold.
In other words, we can identify instances that are too uncertain to be safe no matter the choice of entropy function $H$.
Consequently, the following contributions regarding canonical calibration are well-connected to the previously discussed contributions in Section~\ref{sec:summary_uncertainties_via_bvd} via Theorem~\ref{th:connection_aleatoric_calibration}.

In Section~\ref{sec:summary_neurips_22}, we introduce the notion of proper calibration errors by showing how the calibration-sharpness decomposition known in classification extends to general proper scores beyond classification.
A proper calibration error represents a canonical calibration error directly derived from a proper score.
Further, we show how an injective transformation is a practical approach for improving the proper calibration error of a given model.
Next, in Section~\ref{sec:summary_aistats_24}, we summarize a novel estimator for proper calibration errors in classification, which was missing in the original work on proper calibration errors.
Last, we give a summary of a novel approach for optimizing estimators of squared calibration errors in practice in Section~\ref{sec:summary_tmlr_24}.
This is important since the squared canonical calibration error is a prominent proper calibration error but there is currently no approach in the literature to assess and compare different estimators or hyperparameters of a chosen estimator for this and similar calibration errors.

\subsection{Better Uncertainty Calibration via Proper Scores for Classification and Beyond}
\label{sec:summary_neurips_22}

In this section, we present the summary of Section~\ref{sec:neurips_22}, which incorporates the work \emph{Better Uncertainty Calibration via Proper Scores for Classification and Beyond} \citep[Published at NeurIPS]{gruber2022better}.
The work highlights the need for trustworthy uncertainty estimates in machine learning models, especially in critical applications.
We point out the limitations of existing calibration error estimators, which are usually biased and inconsistent, making it difficult to assess the true reliability of a model's predicted probabilities.
Further, we introduce the concept of proper calibration errors, which are directly linked to proper scores, and assess a model's canonical calibration.
We do this by showing that the calibration-sharpness decomposition of proper scores for classification \citep{Br_cker_2009} can be generalized to general, non-discrete distributions.
Assume a model $f \colon \mathcal{X} \to \mathcal{P}$ and a proper score $S \colon \mathcal{P} \to \mathcal{Y}$ with associated negative entropy $G$ and divergence $D$, where $\mathcal{P}$ is a convex set of general distributions defined for some measurable space $\left( \mathcal{Y}, \mathcal{F}_{\mathcal{Y}} \right)$, input variable $X$ with outcomes in $\mathcal{X}$, target variable $Y$ with outcomes in $\mathcal{Y}$.
Then, in Section~\ref{sec:neurips_22} we show that
\begin{equation}
\begin{split}
    \underbrace{\mathbb{E} \left[ S \left( f \left( X \right), Y \right) \right]}_{\text{Expected Error}} & = \underbrace{\mathbb{E} \left[ D \left( f \left( X \right), \mathbb{P}_{Y \mid f \left( X \right)} \right) \right]}_{\text{Calibration}} \\
    & \quad - \underbrace{\mathbb{E} \left[ D \left( \mathbb{P}_{Y}, \mathbb{P}_{Y \mid f \left( X \right)} \right) \right]}_{\text{Sharpness}} + \underbrace{G \left( \mathbb{P}_{Y} \right).}_{\text{Marginal Noise}}
\end{split}
\end{equation}
We refer to a calibration error defined as $\operatorname{CE} \left( f \right) \coloneqq \mathbb{E} \left[ D \left( \mathbb{P}_{Y}, \mathbb{P}_{Y \mid f \left( X \right)} \right) \right]$ as \textbf{proper calibration error} to highlight the connection to proper scores.
By extending the decomposition beyond classification, we can transfer the concept of canonical calibration in a principled manner to other probabilistic tasks like variance regression.
Another theoretical contribution is to provide a taxonomy of existing calibration errors in classification, aiding in understanding their relationships and limitations and contributing to a clearer overview of the literature.

Estimating a calibration error in practice for a given dataset is notoriously difficult.
We provide an analysis to highlight the limitations of the most prominent estimator in current literature, which is based on histogram binning \citep{naeini2015}.
To circumvent the difficulty of calibration error estimation, we propose to use injective transformations for calibration improvement, which preserve model sharpness and ensure meaningful uncertainty adjustments.
Let $h \colon \mathcal{P} \to \mathcal{P}$ be an injective function, then we show that
\begin{equation}
    \underbrace{\mathbb{E} \left[ S \left( h \circ f \left( X \right), Y \right) \right] - \mathbb{E} \left[ S \left( f \left( X \right), Y \right) \right]}_{\text{Error Improvement}} = \underbrace{\operatorname{CE} \left( h \circ f \right) - \operatorname{CE} \left( f \right)}_{\text{Calibration Improvement}}.
\label{eq:ps_cal_diff}
\end{equation}
This is practically relevant, since the right-hand side of Equation~\eqref{eq:ps_cal_diff} is difficult to estimate, while the left-hand side is straightforward.
A possible choice for $h$ is temperature scaling, which uses a scalar parameter $\alpha \in \mathbb{R}_{>0}$, defined for $A \in \mathcal{F}_{\mathcal{Y}}$ by
\begin{equation}
    h_{\operatorname{TS}} \left( P \right) \left( A \right) \coloneqq \frac{1}{\int_{\mathcal{Y}} \left( P \left( y \right) \right)^\alpha \mathrm{d}\mu \left( y \right)} \left( P \left( A \right) \right)^\alpha,
\end{equation}
where $\int_{\mathcal{Y}} \left( P \left( y \right) \right)^\alpha \mathrm{d}\mu \left( y \right)$ is a normalization constant based on a given base measure $\mu$ of distributions in $\mathcal{P}$.
Informally, Temperature scaling makes likely events likelier and unlikely events unlikelier, or vice versa depending if $\alpha$ is larger or smaller than one.
If we are in classification, then $\mathcal{P} = \Delta^d$, which reduces Temperature scaling to $h_{\operatorname{TS}} \left( P \right) = \frac{1}{\sum_{i=1}^d P_i^\alpha} \left(P_1^\alpha, \dots, P_d^\alpha \right)^\intercal$ \citep{guo2017calibration}.
Injective transformations, like Temperature scaling, usually preserve the accuracy of the model $f$ in classification, which highlights the practicality of such approaches.

A significant part of our empirical contribution in Section~\ref{sec:neurips_22} is to compare a difference related to Equation~\eqref{eq:ps_cal_diff} with current calibration error estimators used in the literature.
We do so across various deep learning architectures and real-world image classification datasets.
The proposed approach is shown to be more robust and reliable than existing methods, particularly in scenarios with limited test data, addressing a common practical challenge.
We also offer evaluations for the calibration of variance regression models to highlight the extension of calibration beyond classification.

In this section, we introduced proper calibration errors.
However, such calibration errors are difficult to estimate due to the target term $\mathbb{P}_{Y \mid f \left( X 
\right)}$.
In the next section, we summarize our contribution in the form of an estimator for proper calibration errors in classification.

\subsection{Consistent and Asymptotically Unbiased Estimation of Proper Calibration Errors}
\label{sec:summary_aistats_24}

In this section, we summarize Section~\ref{sec:aistats_24}, which is about the work \emph{Consistent and Asymptotically Unbiased Estimation of Proper Calibration Errors} \citep[Published at AISTATS]{gruber2024consistent}.
In this work, we address the lack of a general estimator for proper calibration errors in the current literature.
As in the other works, the core of our contribution is based on proper scores, which evaluate the quality of probabilistic predictions and play a crucial role in developing accurate and well-calibrated models.
In the following, we focus on proper scores and proper calibration errors for classification, i.e., $\mathcal{P} = \Delta^d$ in the notation of previous sections.
Other cases are discussed in Section~\ref{sec:open_challenges} as future work.
While an estimator for the proper calibration error based on the Brier score exists (c.f. \citep{popordanoska2022}), an estimator for other cases has not been proposed so far.
The Dirichlet kernel $k_{\operatorname{Dir}} \colon \Delta^d \times \Delta^d \to \mathbb{R}$ is defined by $k_{\operatorname{Dir}} \left( x, y \right) \coloneqq \frac{\Gamma \left( \sum_{i=1}^d \alpha_i \right)}{\prod_{i=1}^d \Gamma \left( \alpha_i \right)} \prod_{i=1}^d y^{\alpha_i-1}_i$ with $\alpha = \frac{x}{h} + 1$ and bandwidth parameter $h$ \citep{ouimet2022asymptotic}.
For a given classifier $f \colon \mathcal{X} \to \Delta^d$ with input random variable $X$ and target random variable $Y$, we propose to estimate $\mathbb{P} \left( Y = y \mid f \left( X \right) = p \right)$ for $y \in \mathcal{Y}$ and $p \in \Delta^d$ via the kernel density ratio
\begin{equation}
\label{eq:kd_ratio_estimator}
    \hat{\mathbb{P}} \left( Y = y \mid f \left( X \right) = p \right) \coloneqq \frac{\sum_{i=1}^n k_{\operatorname{Dir}} \left( p, f \left( X_i \right) \right) \mathbf{1}_{Y_i=y}}{\sum_{i=1}^n k_{\operatorname{Dir}} \left( p, f \left( X_i \right) \right)}
\end{equation}
for a given dataset of $n$ i.i.d. input-target pairs $\mathcal{D} = \left\{ \left(X_1, Y_1 \right), \dots, \left(X_n, Y_n \right) \right\}$.
Then, for a model $f$ and a given proper calibration error $\operatorname{CE}$ induced by a proper score with associated divergence $D$, we propose the estimator
\begin{equation}
    \hat{\operatorname{CE}} \left( f \right) \coloneqq \frac{1}{n} \sum_{i=1}^n D \left( f \left( X_i \right), \hat{\mathbb{P}}_{Y \mid f \left( X \right) = f \left( X_i \right)} \right)
\end{equation}
with $\hat{\mathbb{P}}_{Y \mid f \left( X \right) = p} \coloneqq \left( \hat{\mathbb{P}} \left( Y = 1 \mid f \left( X \right) = p \right), \dots, \hat{\mathbb{P}} \left( Y = d \mid f \left( X \right) = p \right) \right)^\intercal \in \Delta^d$.
In a successive analysis, we show that this estimator is consistent and asymptotically unbiased with $n \to \infty$ under the assumption that $D$ is differentiable.
In mathematical terms, consistency means
\begin{equation}
    \lim_{n \to \infty} \mathbb{P}_{\mathcal{D}} \left( \left\lvert \hat{\operatorname{CE}} \left( f \right) - \operatorname{CE} \left( f \right) \right\rvert > \epsilon \right) = 0
\end{equation}
for all $\epsilon > 0$, and asymptotically unbiasedness that
\begin{equation}
    \lim_{n \to \infty} \mathbb{E}_{\mathcal{D}} \left[ \hat{\operatorname{CE}} \left( f \right) \right] = \operatorname{CE} \left( f \right).
\end{equation}
We also introduce a similar estimator based on $\hat{\mathbb{P}} \left( Y = y \mid f \left( X \right) = p \right)$ for the sharpness term with similar properties.

As an additional theoretical result, we establish a connection between sharpness and information monotonicity, providing insights into the workings of neural networks.
Specifically, we show how the sharpness term is equal to a multi-distribution f-Divergence (c.f. \citep{duchi2018multiclass} for a definition) and implies a generalized definition of mutual information between the model output $f \left( X \right)$ as a random variable and the target random variable $Y$.
The form of the general mutual information is only dependent on the proper score $S$, and choosing the log score implies the mutual information as defined in information theory \citep{mackay2003information}.
We then prove that the general mutual information between the values in an intermediate layer of a neural network and the target variable is bounded by the general mutual information of the previous layer.

As experimental results, we evaluate our estimator on common neural network architectures and image classification tasks.
In these evaluations, we emphasize the proper calibration error based on the Kullback-Leibler divergence, which is induced by the log score.
In the literature, this is the first evaluation of the Kullback-Leibler calibration error, which is more principled according to information theory than the squared calibration error based on the Brier score.
Specifically, the Kullback-Leibler calibration error is induced by the log score, which is also known as cross-entropy loss and is a more common training objective for neural networks than the Brier score \citep{goodfellow2016deep}.
The experiments validate the claimed properties of the proposed estimator and suggest that the selection of a calibration method should depend on the specific calibration error of interest.

To summarize the core contribution in the context of this thesis, we proposed a general estimator for proper calibration errors in classification and offered extensive evaluations with the novel Kullback-Leibler calibration error.
In Section~\ref{sec:open_challenges}, we will discuss how our estimator may be generalized beyond classification.
In general, the estimation of calibration errors is a challenging problem and multiple estimators with a variety of hyperparameters have been proposed in the literature.
Yet, it remains an open question of how to pick an estimator for a given dataset and model combination in practice.
In the next section, we propose a novel procedure to tackle this research problem for squared calibration errors.

\subsection{Optimizing Estimators of Squared Calibration Errors in Classification}
\label{sec:summary_tmlr_24}


In this section, we summarize our contributions of Section~\ref{sec:tmlr_24}, which incorporates the work \emph{Optimizing Estimators of Squared Calibration Errors in Classification} \citep[In Submission at TMLR]{gruber2024optimizingestimatorssquaredcalibration}.
As we have discussed in previous sections, calibration errors in classification quantify the alignment between predicted probabilities and true likelihoods given the prediction.
This is crucial for trustworthy and interpretable machine learning models.
However, existing literature does not guide how to select an estimator and its hyperparameter for a calibration error.
In this work, we propose a mean-squared error-based risk objective for comparing and optimizing calibration estimators.
The critical advantage is that the proposed risk is applicable in practical settings and works for all notions of calibration, including canonical calibration as implied by the calibration-sharpness decomposition in Equation~\eqref{eq:cal_sharp_decomp_classif}.
We achieve this by reformulating common estimators of squared calibration errors as a regression problem with i.i.d. input pairs.
For this, assume a dataset $\mathcal{D} = \left\{ \left( X_1, Y_1 \right), \dots, \left( X_n, Y_n \right) \right\}$ of $n$ i.i.d. random variables of input-target pairs, which we want to use to estimate the calibration of a classifier $f \colon \mathcal{X} \to \Delta^d$.
Then, we show that the binning-based estimator in \citep{naeini2015} and the kernel density-based estimator in \citep{popordanoska2022} can be formulated as
\begin{equation}
    \hat{\operatorname{CE}}_2 \left( f \right) = \frac{1}{n} \sum_{i=1}^n h \left( f \left( X_i \right), f \left( X_i \right) \right)
\end{equation}
for some function $h$ depending on the estimator and the notion of calibration.
We refer to such an $h$ as \textbf{calibration estimation function}.
The empirical risk we propose for a calibration estimation function $h$ is then defined via
\begin{equation}
\begin{split}
    \hat{\mathcal{R}}_{\operatorname{CE}} \left( h \right) & \coloneqq \frac{1}{n\left(n-1\right)} \cdot \\
    & \quad \sum_{i=1}^n \sum_{\substack{j=1 \\ j \neq i}}^n \left( \left\langle f \left( X_i \right) - e_{Y_i}, f \left( X_j \right) - e_{Y_j} \right\rangle_{\mathbb{R}^d} - h \left( f \left( X_i \right), f \left( X_j \right) \right) \right)^2,
\label{eq:emp_risk_ce}
\end{split}
\end{equation}
where we assume $h$ is an estimator for canonical calibration with $d$ classes.
An analogous risk for other notions of calibration, like top-label confidence calibration can also be defined.
We provide a rigorous measure-theoretic foundation for the proposed risk metric, which shows under which conditions the expected risk identifies a calibration estimation function such that $\mathbb{E}_{\mathcal{D}} \left[ \hat{\operatorname{CE}}_2 \left( f \right) \right] = \operatorname{CE}_2 \left( f \right)$.
We advocate the following training-validation-testing pipeline when optimizing a calibration error estimator since our final estimate should ideally not depend on the dataset we used for optimizing the estimator.
Similar to conventional machine learning procedure, we require a split of the original dataset $\mathcal{D}$ into a training set $\mathcal{D}_{\mathrm{tr}}$, a validation set  $\mathcal{D}_{\mathrm{val}}$, and a test set  $\mathcal{D}_{\mathrm{te}}$.
Denote with $h_{\mathcal{D}_{\mathrm{tr}}, \mathcal{D}_{\mathrm{val}}}$ the calibration estimation function fitted on $\mathcal{D}_{\mathrm{tr}}$ and with optimized hyperparameters based on $\mathcal{D}_{\mathrm{val}}$ and according to the empirical risk in Equation~\eqref{eq:emp_risk_ce}.
Then, we propose to estimate the calibration error via
\begin{equation}
    \hat{\operatorname{CE}}_2 \left( f \right) = \frac{1}{\left\lvert \mathcal{D}_{\mathrm{te}} \right\rvert} \sum_{\left(X, Y \right) \in \mathcal{D}_{\mathrm{te}}} h_{\mathcal{D}_{\mathrm{tr}}, \mathcal{D}_{\mathrm{val}}} \left( f \left( X \right), f \left( X \right) \right).
\end{equation}
We also propose novel calibration estimators based on kernel ridge regression, which are closed-form solutions of the risk objective according to certain assumptions.
The effectiveness of the proposed pipeline is demonstrated by optimizing existing estimators and comparing them with our novel estimators on common image classification tasks.
The experiments show that no single calibration estimator outperforms others across all settings.
This result emphasizes the importance of using the proposed risk to select an appropriate estimator for new settings.

In conclusion, we contributed to the advancement of better calibration estimation techniques in practice, which promotes the development of more trustworthy machine learning models.
The contributions in this section were mostly for classification, and future work may emphasis more novel tasks as well (c.f. Section~\ref{sec:open_challenges} for a discussion).
In the following, we shift our focus once more to generative models and emphasize the balance between well-established tasks, like classification, and more recent tasks, like data generation, throughout this thesis.
Specifically, we propose a novel approach to evaluate generative models in a more fine-grained and more interpretable way based on a unique property of the kernel spherical score.

\section{Disentangling Mean Embeddings for Better Diagnostics of Image Generators}
\label{sec:summary_neurips_ws_24}

In previous sections, we contributed toward an evaluation framework based on proper scores for uncertainty quantification across a multitude of different tasks.
So far, we always assumed a proper score is given and then derived all successive quantities.
In this section, we offer an orthogonal contribution, which allows a more fine-grained evaluation of the previous contributions under specific circumstances.
In the following, we summarize Chapter~\ref{ch:disentangling_mean_embeddings}, which presents the work \emph{Disentangling Mean Embeddings for Better Diagnostics of Image Generators} \citep[Accepted at IAI Workshop @ NeurIPS]{gruber2024disentanglingmeanembeddingsbetter}.
There we contribute how the cosine similarity of mean embeddings in an RKHS may be disentangled into the product of cosine similarities of smaller RKHS.
We apply this technique in the context of image generation and discuss how this enhances the diagnostics during model training.
Note that the cosine similarity of mean embeddings is up to a target-dependent factor proportional to the expected kernel spherical score.
Consequently, in the following, we state the core contribution via the kernel spherical score to highlight the connection with previous chapters of this thesis.

The evaluation of image generators is challenging due to the limitations of traditional image evaluation metrics.
One such limitation is the inability to determine the contribution of different image regions to the overall model error.
In Chapter~\ref{ch:disentangling_mean_embeddings}, we propose a novel approach to disentangle the cosine similarity of mean embeddings into the product of cosine similarities for individual pixel clusters, which represent image regions.
This allows for the evaluation and interpretation of the model performance of each cluster in isolation.
The approach enhances the explainability of image generation models and the likelihood of identifying the source pixel region of model misbehavior.
The method is based on central kernel alignment to determine the degree of independence between pixels, which is then used in hierarchical clustering to form pixel clusters as follows.
Let us denote the expected kernel spherical score (EKS) for a prediction $P \in \mathcal{P}$ and a target random variable $Y \sim Q \in \mathcal{P}$ with
\begin{equation}
    \operatorname{EKS}_k \left( P, Q \right) \coloneqq \mathbb{E}_{Y \sim Q} \left[ S_{k\text{-spherical}} \left( P, Y \right) \right].
\end{equation}
Then, the EKS is related to the cosine similarity between the mean embeddings $\mu_P$ and $\mu_Q$ of prediction $P$ and target distributions $Q$ via
\begin{equation}
    \frac{\operatorname{EKS}_k \left( P, Q \right)}{\left\lVert \mu_Q \right\rVert_{\mathcal{H}}} = \frac{\left\langle \mu_P, \mu_Q \right\rangle_{\mathcal{H}}}{\left\lVert \mu_P \right\rVert_{\mathcal{H}} \left\lVert \mu_Q \right\rVert_{\mathcal{H}}},
\end{equation}
where the right-hand side is the generalization of the cosine similarity to a RKHS $\mathcal{H}$.

Now, we translate the theoretical contribution of Chapter~\ref{ch:disentangling_mean_embeddings} from the cosine similarity to the expected kernel spherical score.
Assume we have random variables $X = \left( X_1, \dots, X_d \right)^\intercal$ and $Y = \left( Y_1, \dots, Y_d \right)^\intercal$ with outcomes in a space $\mathcal{X}^d$ and for a p.s.d. kernel $k \colon \mathcal{X} \times \mathcal{X} \to \mathbb{R}$, 
there exists a partition $\mathbf{I}$ of the indices $\left\{ 1, \dots, d \right\}$ such that for all $I,I^\prime \in \mathbf{I}$ it holds $\operatorname{CKE}_{k^{\otimes \lvert I \rvert},k^{\otimes \lvert I^\prime \rvert}} \left( \mathbb{P}_{X_I X_{I^\prime}} \right) = 0 = \operatorname{CKE}_{k^{\otimes \lvert I \rvert},k^{\otimes \lvert I^\prime \rvert}} \left( \mathbb{P}_{Y_{I} Y_{I^\prime}} \right)$ with ${X_{I}} \coloneqq \left( X_i \right)_{i \in I}^\intercal$ and ${Y_{I^\prime}} \coloneqq \left( Y_i \right)_{i \in I^\prime}^\intercal$. Then, we have
\begin{equation}
    \operatorname{EKS}_{k^{\otimes d}} \left( \mathbb{P}_X, \mathbb{P}_{Y} \right) = \prod_{I \in \mathbf{I}} \operatorname{EKS}_{k^{\otimes d}} \left( \mathbb{P}_{X_I}, \mathbb{P}_{Y_I} \right).
\label{eq:eks_disentanglement}
\end{equation}
The proof for Equation~\eqref{eq:eks_disentanglement} is analogous to the one for cosine similarity in Chapter~\ref{ch:disentangling_mean_embeddings} by removing the factor $\left\lVert \mu_Q \right\rVert_{\mathcal{H}}$ in all equations.

In practice, we may assume the random variable $X$ represents generated images and the random variable $Y$ the images from a training or test set.
Then, the random variables $X_1, \dots, X_d$ (and $Y_1, \dots, Y_d$ respectively) represent the pixels in the images.
The central kernel alignment is used to compute a pixel-by-pixel correlation matrix based on the training set.
We then use hierarchical clustering on this correlation matrix to find independent pixel clusters.
The approach is demonstrated on various real-world use cases, like image generation of celebrity faces or Chest X-Ray scans.
The findings demonstrate how we improve the interpretability of the performance evaluation for models trained to generate such images.
Specifically, we can identify model misbehavior more easily and more transparently.
For example, we successfully assign changes of the overall error to individual pixel regions.

The core assumption of our approach is that, first, mean embeddings are a meaningful representation of the images. This assumption is also shared with the maximum mean discrepancy, i.e., the associated divergence of the kernel score, used in the image processing literature (c.f. \citep{scholkopf1997support}).
Second, the image distribution of the training set is structured well enough to allow independent pixel regions.
This case is not always given, for example, a dataset of images of all human organs may have organs appear in different shapes in any region of the image.
Consequently, our approach may not find perfectly independent clusters in all practical cases.
However, we also discuss how violated assumptions can be verified simply.

In summary, our approach offers a valuable tool for evaluating image generators based on a rigorous theory applicable to cosine similarity and the kernel spherical score.
Specifically, we make it possible to detect and explain the performance of image generators across various image regions, which is an important aspect of safe applications.
In Section~\ref{sec:future_directions}, we discuss how this may be linked more directly with uncertainty quantification.
This is in principle possible since the individual clusters are assessed via a proper score again.

    \chapter{Uncertainties induced by the Bias-Variance Decomposition of Proper Scores}
\label{ch:uncertainties_via_bvd}

\section{Uncertainty Estimates of Predictions via a General Bias-Variance Decomposition}
\label{sec:aistats_23}
\ifnum\renderpapers=1
    {\includepdf[pages=-, pagecommand={}]{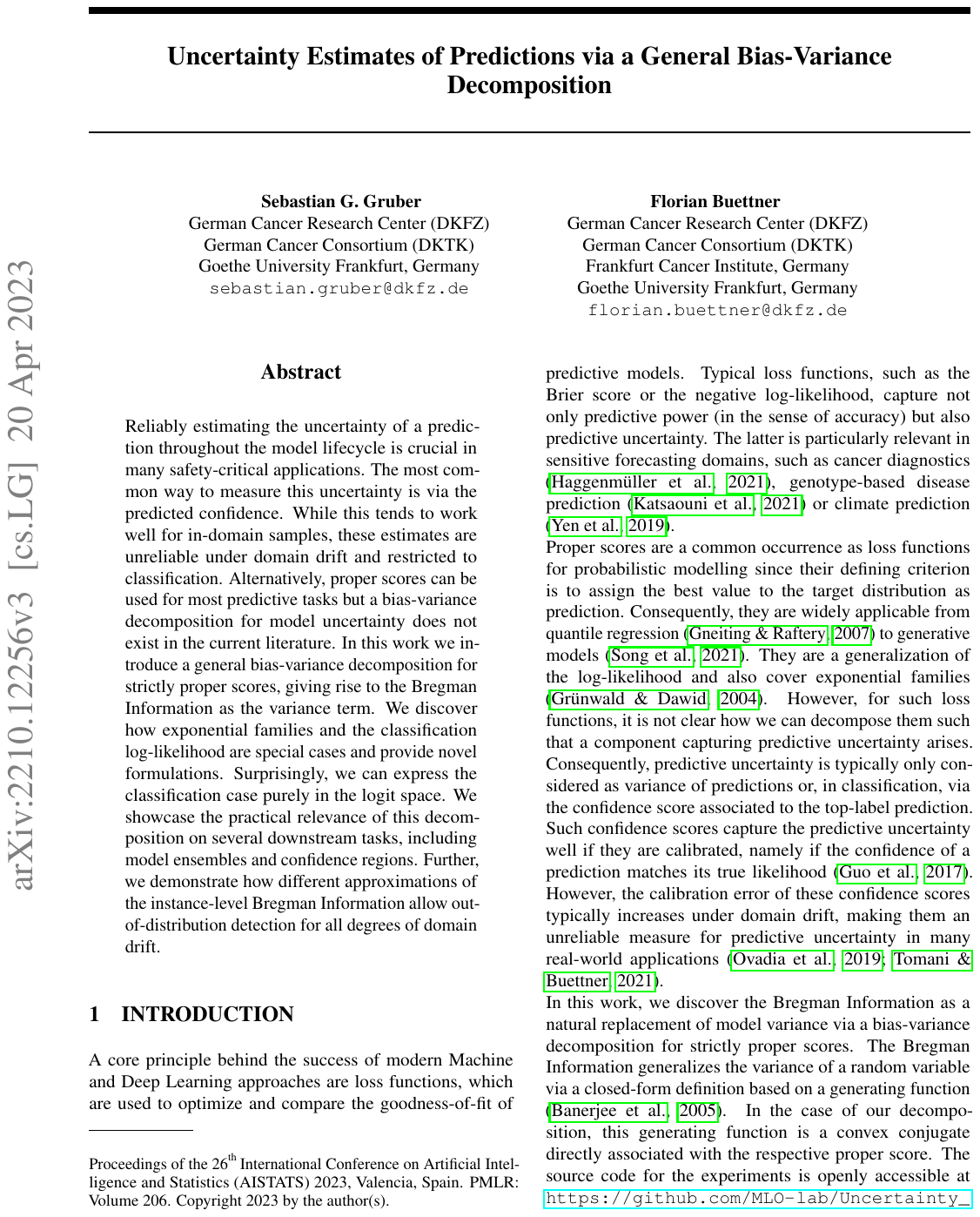}} 
\fi

\section{A Bias-Variance-Covariance Decomposition of Kernel Scores for Generative Models}
\label{sec:icml_24}
\ifnum\renderpapers=1
    {\includepdf[pages=-, pagecommand={}]{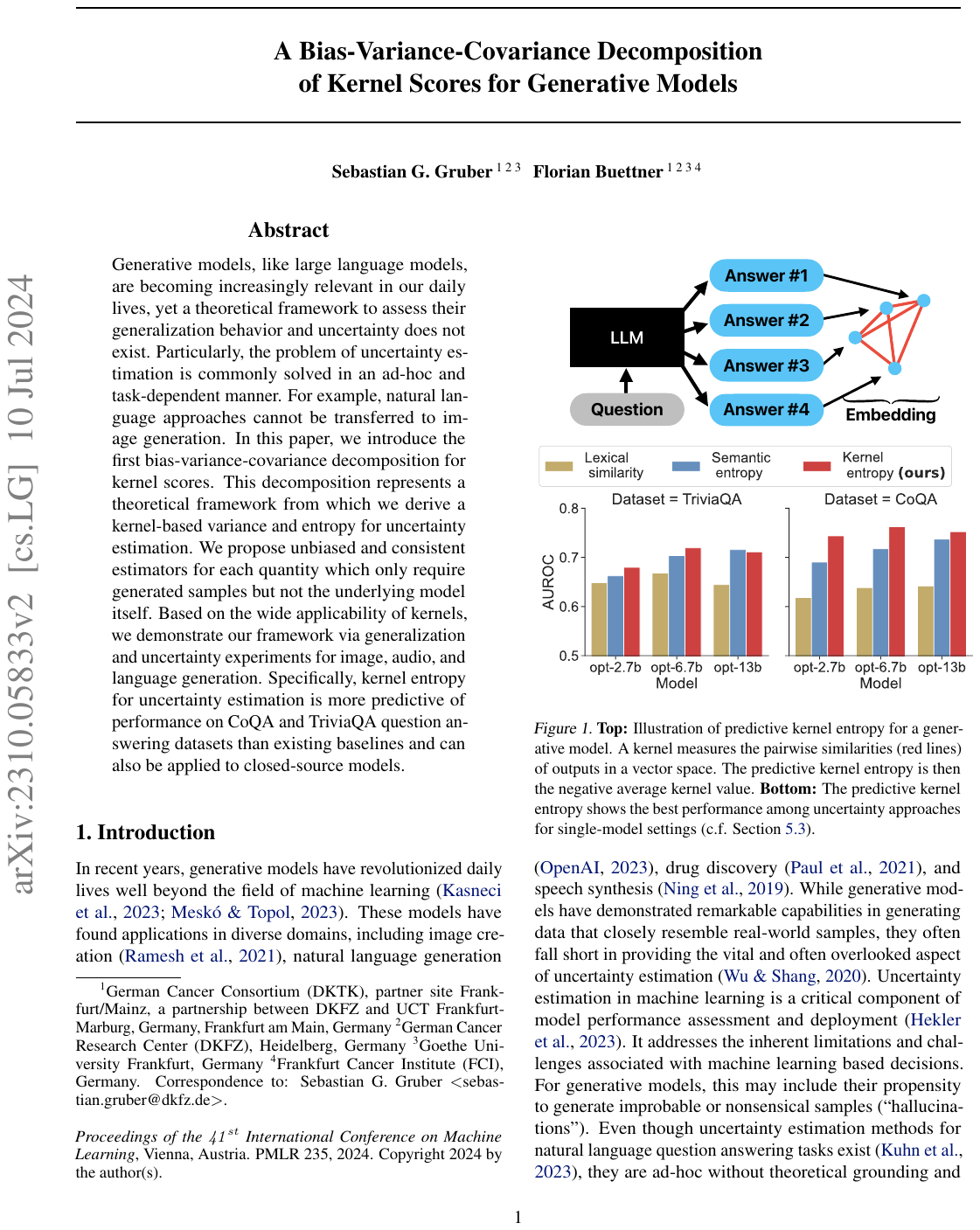}} 
\fi


    \chapter{Improving Uncertainties via Calibration-Sharpness Decomposition of Proper Scores}
\label{ch:uncertainties_via_cal_sharp}

\section{Better Uncertainty Calibration via Proper Scores for Classification and Beyond
}
\label{sec:neurips_22}
\ifnum\renderpapers=1
    {\includepdf[pages=-,  pagecommand={}]{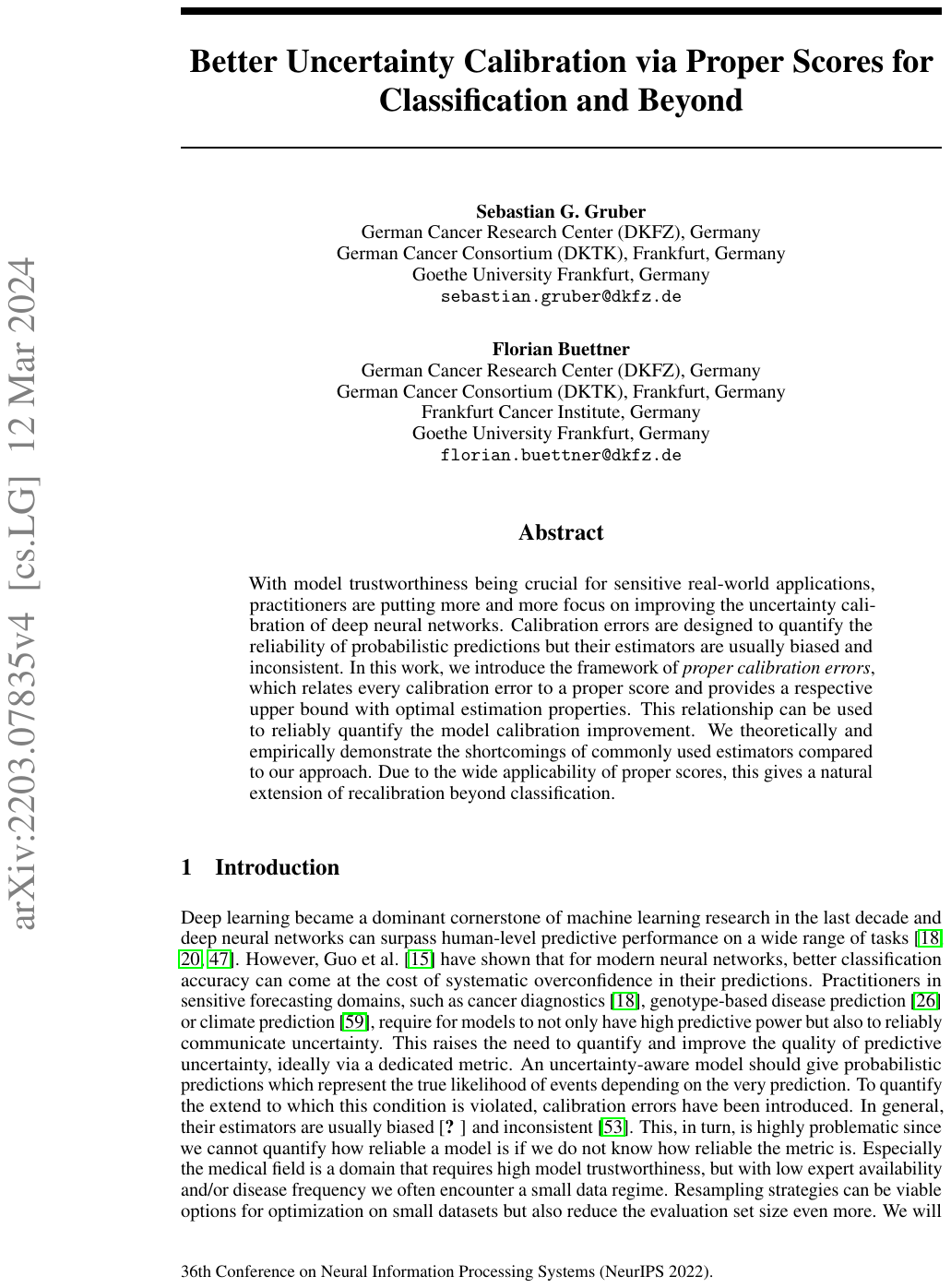}} 
\fi

\section{Consistent and Asymptotically Unbiased Estimation of Proper Calibration Errors}
\label{sec:aistats_24}
\ifnum\renderpapers=1
    {\includepdf[pages=-,  pagecommand={}]{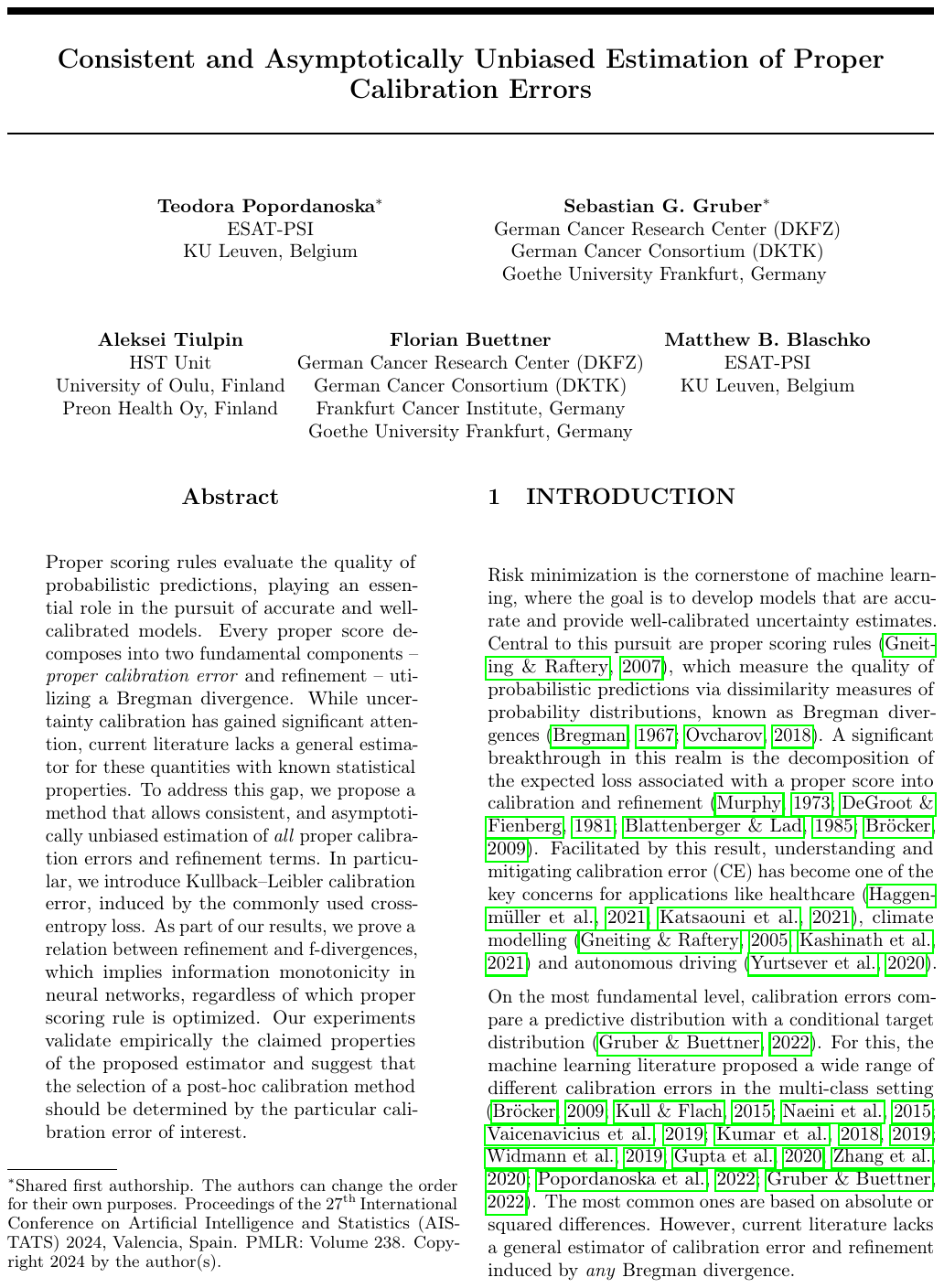}} 
\fi
\section{Optimizing Estimators of Squared Calibration Errors in Classification}
\label{sec:tmlr_24}
\ifnum\renderpapers=1
    {\includepdf[pages=-,  pagecommand={}]{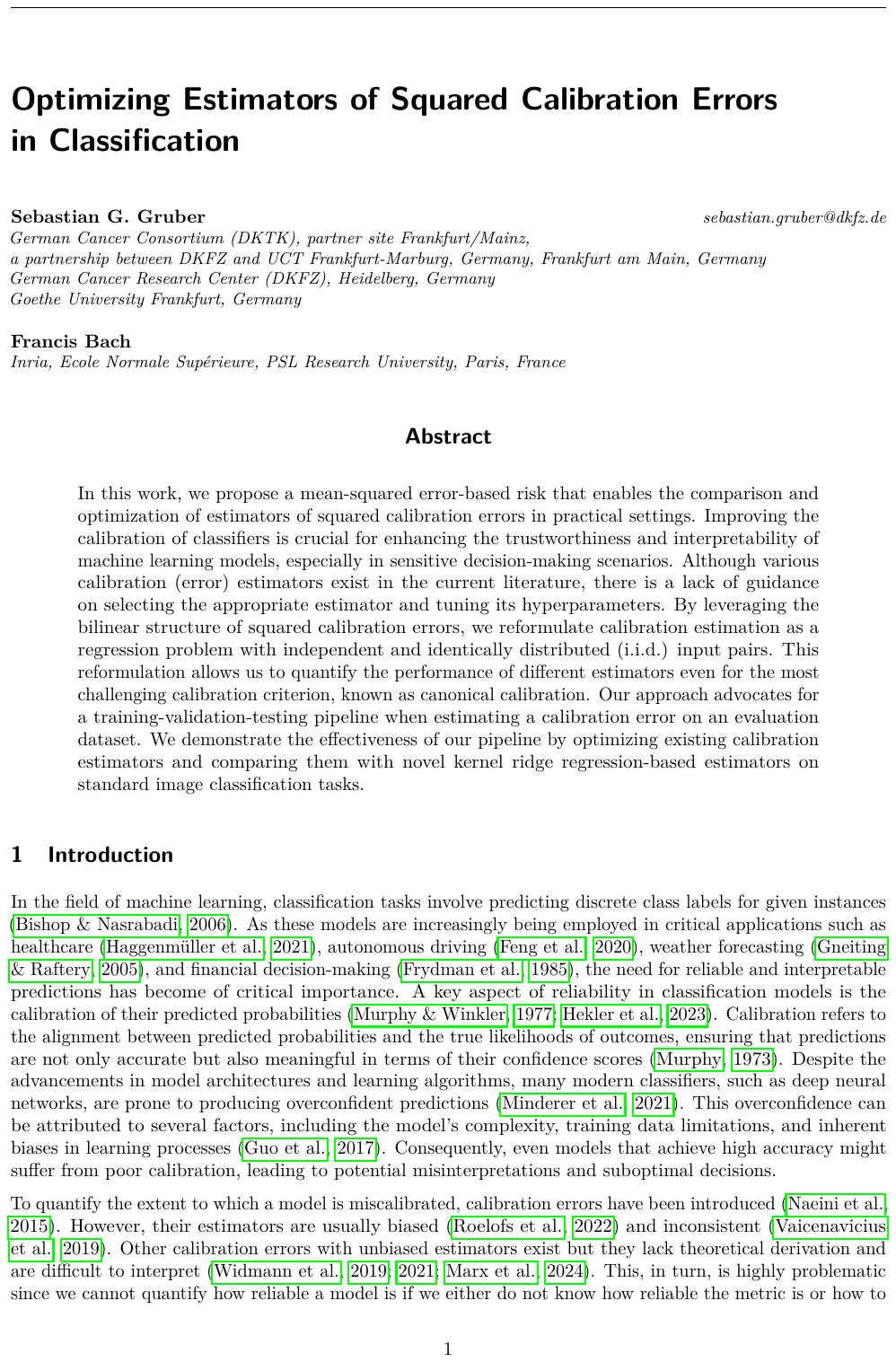}} 
\fi

    \chapter{Disentangling Mean Embeddings for Better Diagnostics of Image Generators}
\label{ch:disentangling_mean_embeddings}
\ifnum\renderpapers=1
    {\includepdf[pages=-, pagecommand={}]{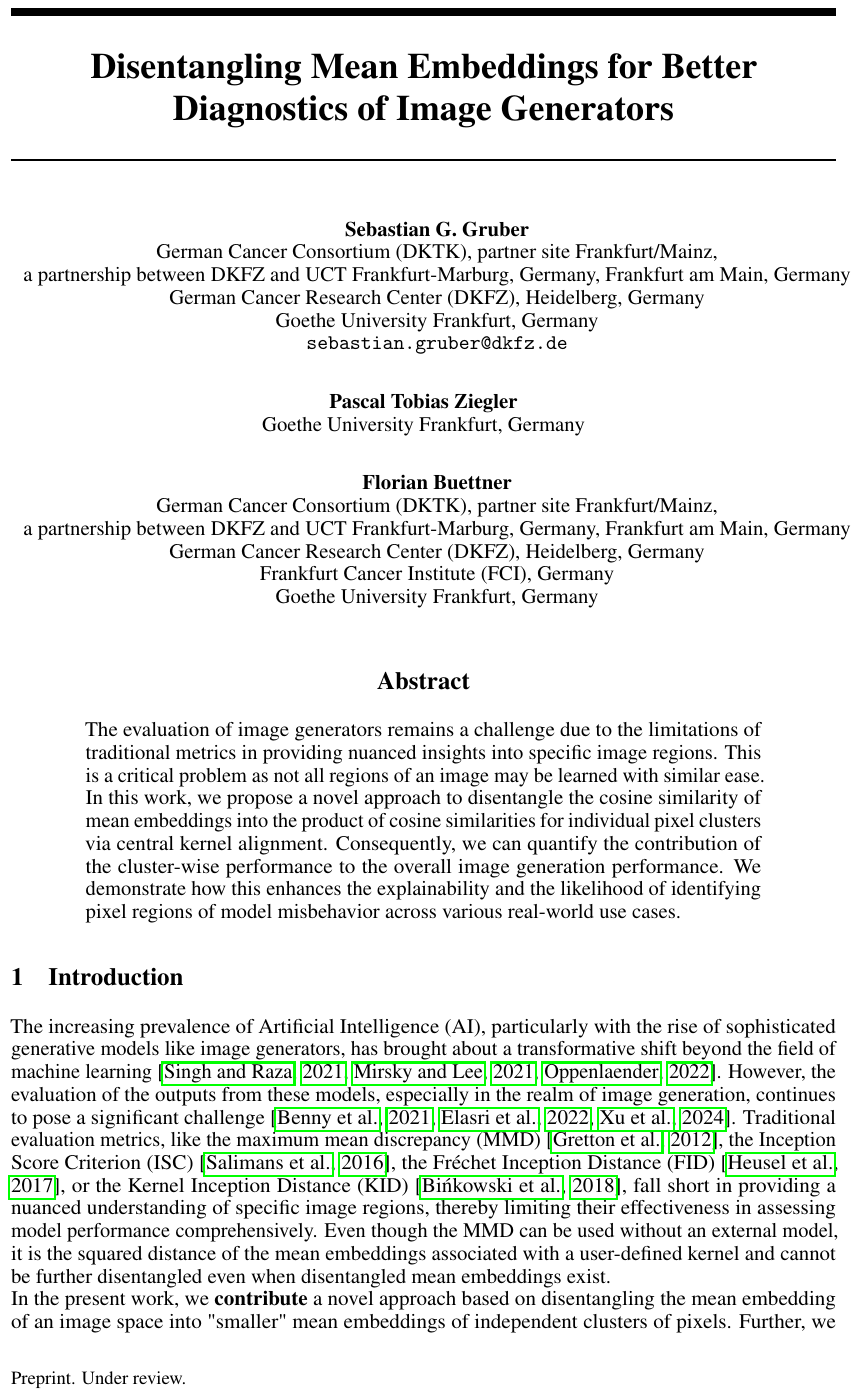}} 
\fi
    \chapter{Conclusion and Outlook}
\label{ch:conclusion_outlook}

\section{Summary}
\label{sec:thesis_summary}

This thesis established a novel framework for uncertainty quantification in machine learning via proper scores.
Proper scores are a class of loss functions for probabilistic predictions, which are by definition minimized in expectation by predicting the true target distribution.
The core contributions are new theoretical results for proper scores and their associated entropy and divergence functions.
These theoretical results are coupled with empirical evaluations, which underline the relevance of the presented theory.
Specifically, we showed that Bregman Information can be a more informative measure of uncertainty than confidence scores for out-of-distribution scenarios.
Further, the kernel score entropy function (called kernel entropy in Section~\ref{sec:icml_24}) outperforms state-of-the-art baselines for uncertainty estimation of large language models.
Even though we offered strong empirical results, the purpose of our framework is to generalize the ideas and intuition we develop from one application to new and unexplored machine learning tasks.
For example, the bias-variance decomposition allows to generalize concepts developed in regression and classification to more general tasks, like generative modeling.
Such an abstraction of results may seem unnecessarily complicated when looking at each task in isolation.
However, we argue that the scientific field of machine learning develops novel approaches for new tasks regularly, which makes it inefficient to invent methods for uncertainty quantification in each new task from scratch.
In consequence, our contributions are not theory for advancing theory but rather theory for a better understanding of empirical approaches and evaluations in existing and newly given tasks.
Via this approach, we successfully presented novel actionable insights for machine learning practitioners.
We summarize the results in the individual chapters to support our claims further.

In Chapter~\ref{ch:uncertainties_via_bvd}, we approached uncertainty quantification via the bias-variance decomposition of proper scores, with a particular focus on generative models.
We showed how general variance and bias terms are expressed as Bregman Information and Bregman divergence in a dual space.
We offered subsequent results for ensemble predictions.
Further, in Section~\ref{sec:summary_uncertainties_via_bvd}, we connected the notions of aleatoric and epistemic uncertainty with quantities arising in the decomposition.
Specifically, we suggested that the general variance term can be seen as a measure of epistemic uncertainty, while the noise term provides insights into aleatoric uncertainty.
This is already well-established for classification and regression, and our results allow a translation to generative models.
Our empirical results support the claim that such a translation of concepts offers novel solutions for generative models.
Our approach of using the entropy of the kernel score combined with semantic embeddings outperforms other state-of-the-art baselines across a variety of question-answering tasks with large language models.
Other applications we assessed with our framework are image generation and text-to-speech audio generation.
In total, the empirical evaluations support the claim that our framework can be used to derive state-of-the-art solutions in practice.

In Chapter~\ref{ch:uncertainties_via_cal_sharp}, we contributed new insights into classification and generalized the theory regarding calibration-sharpness beyond classification.
Our generalized theory includes the formulation of proper calibration errors, i.e., canonical calibration errors induced by proper scores.
Specifically, the recommendations of \cite{maier2024metrics} for practitioners are based on our contributions towards calibration.
Further, we proposed a general estimator for proper calibration errors in classification and analyzed its theoretical properties.
We offered insights into the taxonomy behind various calibration errors in classification.
We also analyzed and empirically demonstrated shortcomings of binning-based calibration error estimators used in the literature, and proved the existence of information monotonicity in neural networks for any proper score in classification.
Last, we introduced a mean-squared error-based risk for optimizing estimators of squared calibration errors in classification.
This allows us to compare different estimators and tune their hyperparameters.
This risk-based approach implies a training-validation-testing pipeline whenever we estimate the calibration error of a classifier.
Experimental results show that no calibration error estimator dominates other estimators, which advocates the usage of our proposed risk in practice.
In Section~\ref{sec:summary_uncertainty_cal_sharp}, we described how our results regarding canonical calibration are related to aleatoric uncertainty estimates.

In Chapter~\ref{ch:disentangling_mean_embeddings}, we offered a novel approach to disentangling the cosine similarity of mean embeddings.
The disentanglement is based on a kernel-based independence measure between sub-domains of the original target domain.
Specifically, we proved under which conditions we can express the cosine similarity of the whole target domain as the product of similarities for each sub-domain.
In our experiments, we used image generation as a task with the image space as the target domain and clusters of pixels as sub-domains.
We then disentangled the image-wise error of the image generator into cluster-wise errors, which offered novel diagnostics of model misbehavior.
Specifically, we highlighted the practical relevancy of the theoretical contributions along various real-world use cases.
In Section~\ref{sec:summary_neurips_ws_24}, we proved how our theoretical results regarding the cosine similarity of mean embeddings can be readily translated to the kernel spherical score.

In summary, we provided multiple stand-alone works regarding uncertainty quantification and model evaluation for classification and beyond.
These can be tied together into a single framework via proper scores, which are the foundation of each work.
The notions of aleatoric and epistemic uncertainty help to translate well-established intuition to abstract statistical quantities, which apply to novel tasks, like generative modeling.
The practical algorithms derived within this framework can offer novel insights into model behavior and outperform state-of-the-art baselines, which are often ad-hoc and limited to a specific setup.
This implies that transferring these algorithms to new tasks, like graph or video generation, may result in equally strong performance.
In the following, we discuss important open challenges within the framework and future research directions regarding model evaluation.

\section{Open Challenges}
\label{sec:open_challenges}

Our theoretical contributions are based on proper scores, which are a broad class of loss functions.
In this thesis, we combined these contributions towards a novel framework for uncertainty quantification.
However, not all our contributions apply to all proper scores.
Further, the experimental evaluations are also only conducted for a small selection of proper scores.
In this section, we discuss some missing theoretical and empirical pieces in the presented framework, which require additional proofs and evaluations in future work.

In Chapter~\ref{ch:uncertainties_via_bvd}, we provided a bias-variance decomposition for strictly proper scores.
However, it is not clear if the same holds for all non-strictly proper scores.
The problem lies in the fact that a non-strictly proper score has a non-strictly concave entropy function.
The subgradient of a non-strictly concave function is in general not invertible, which makes the used technique for exchanging the arguments in the Bregman divergence infeasible.
It is up to future research to explore if there is a workaround for this restriction and to prove the bias-variance decomposition for non-strictly proper scores.
Further, we showed strong empirical performance of the entropy function associated with the kernel score for uncertainty estimation.
However, a theoretical explanation is missing, which is required for performance guarantees.
For image and audio generation, we could even detect a strong linear correlation between the evaluated kernel score and associated entropy.
We hypothesize that linearity is not a coincidence and see it as future research to offer a rigorous theory under which conditions the entropy and kernel score align.
Such a result may also offer a general explanation about the reliability of entropy functions as uncertainty estimates.

In Chapter~\ref{ch:uncertainties_via_cal_sharp}, we generalized the calibration-sharpness decomposition beyond classification and subsequently introduced proper calibration errors and how to optimize them via injective transformations.
However, we do not offer evaluations beyond classification and variance regression.
In future research, we aim to translate the concept of calibration to generative models according to the calibration-sharpness decomposition and to optimize model outputs via injective transformations as suggested by our theoretical contributions.
Further, the estimator for proper calibration errors was only defined and analyzed in the classification case.
Another future contribution is to develop estimators for general proper calibration errors.
For example, for a calibration error $\operatorname{CE}_k$, which can be expressed as
\begin{equation}
    \operatorname{CE}_k \left( f \right) = \mathbb{E} \left[ D \left( \mu_{f \left( X \right)}, \mu_{\mathbb{P}_{Y \mid f \left( X \right)}} \right) \right]
\end{equation}
based on a divergence $D$ and mean embeddings associated with a kernel $k$,
we may generalize the estimator in Equation~\eqref{eq:kd_ratio_estimator} to
\begin{equation}
\label{eq:general_kd_ratio_estimator}
    \left\langle \hat{\mu}_{\mathbb{P}_{Y \mid f \left( X \right) = p}}, \phi \left( y \right) \right\rangle_{\mathcal{H}} \coloneqq \frac{\sum_{i=1}^n k_{\operatorname{Dir}} \left( p, f \left( X_i \right) \right) k \left( Y_i, y \right)}{\sum_{i=1}^n k_{\operatorname{Dir}} \left( p, f \left( X_i \right) \right)},
\end{equation}
where $\mathcal{H}$ is the associated RKHS and $\hat{\mu}_{\mathbb{P}_{Y \mid f \left( X \right) = p}}$ is an estimator for $\mu_{\mathbb{P}_{Y \mid f \left( X \right) = p}}$.
Examples of such calibration errors are the proper calibration errors induced by the kernel score and the kernel spherical score.

However, estimating calibration errors beyond classification is at least as difficult as for classification.
Consequently, we also require tools to better assess calibration estimators in practice.
The risk-based optimization approach we developed is an important step in this direction and further research to generalize this concept beyond classification is required.

Further, the connection between calibration and the other types of uncertainty is also not well explored.
Offering more results regarding the link between model calibration and aleatoric uncertainty estimates of the same model would further unify Chapter~\ref{ch:uncertainties_via_bvd} and Chapter~\ref{ch:uncertainties_via_cal_sharp}.
The contributions within these chapters may currently be seen as orthogonal approaches in uncertainty quantification outside of this thesis.

In Chapter~\ref{ch:disentangling_mean_embeddings}, we disentangled the cosine similarity of mean embeddings for better model diagnostics.
The cosine similarity is closely related to the kernel spherical score and it is an open question if other kernel-based proper scores exist, which also allow such a disentanglement.
Furthermore, we only evaluated image generation models with the provided theory.
Other machine learning tasks may exist, which also benefit from such a disentanglement.
Additionally, the evaluation was only for the generalization error without offering experiments for uncertainty quantification.
The connection between performance disentanglement and uncertainty quantification may be further explored in the future.

\section{Future Directions}
\label{sec:future_directions}

In the last section, we discussed open challenges in the here presented framework for uncertainty quantification.
However, we may see uncertainty quantification as an application of a more general approach to offer contributions regarding proper scores.
In the most general sense, proper scores are simply an approach to evaluate the performance of machine learning models.
The definition of proper scores is based on practical assumptions, for example, we require a target sample and a distribution prediction as inputs.
However, the original definition of proper scores appeared when classification and regression were the most prominent prediction tasks.
In recent developments, generative modeling gained significant traction in academic and industrial applications.
It is therefore an open question if the original definition of proper scores is still appropriate for such modern approaches.
This is especially relevant since generative models are notoriously difficult to evaluate.
It might be a fruitful future research direction to refine or modify proper scores for generative models.
To the best of our knowledge, it is currently unknown if the resulting class of loss functions is then a subset, a superset, or simply a partially overlapping set to proper scores.
Regardless, such research may present valuable real-world contributions since generative modeling has an, in principle, endless variety of possible domains.
And, supported by the recent breakthrough in large language models, we claim the rise of new domains for generative modeling is unpredictable.
It is therefore a reasonable approach to offer a principled theory of how to evaluate generative models in an abstract manner, which significantly increases the likelihood that future applications will be covered by the theory.
Otherwise, we may encounter similar evaluation problems repeatedly whenever a new domain arises.
As a possible approach, we want to emphasize the large flexibility of kernels throughout our contributions.
For example, we successfully used the kernel score and associated quantities for image, audio, and language generation.
Consequently, more research into kernel methods for evaluation purposes could offer the required flexibility for new and unknown applications.
In general, we see it as an open challenge to explore kernel-based proper scores and their possible applications beyond the here presented cases.

On a general level, our approach to developing theory had the purpose of guiding the experimental evaluations for uncertainty quantification.
Specifically, our theoretical results improve the intuition and understanding behind the conducted experiments.
Within our results, theory and evaluations act as mutual reinforcements -- theory allows to transfer of informal concepts across seemingly disconnected applications in practice, while practical evaluations justify the unavoidable theoretical assumptions.
Based on the results presented in this thesis, it is a promising long-term goal to further develop theoretical contributions regarding evaluating the most modern machine learning approaches in an empirically testable and verifiable manner.
    
    \appendix
        
        
        \bibliography{main}{}

\begin{thebibliography}{105}
\providecommand{\natexlab}[1]{#1}
\providecommand{\url}[1]{\texttt{#1}}
\expandafter\ifx\csname urlstyle\endcsname\relax
  \providecommand{\doi}[1]{doi: #1}\else
  \providecommand{\doi}{doi: \begingroup \urlstyle{rm}\Url}\fi

\bibitem[Abdar et~al.(2021)Abdar, Pourpanah, Hussain, Rezazadegan, Liu, Ghavamzadeh, Fieguth, Cao, Khosravi, Acharya, et~al.]{abdar2021review}
Moloud Abdar, Farhad Pourpanah, Sadiq Hussain, Dana Rezazadegan, Li~Liu, Mohammad Ghavamzadeh, Paul Fieguth, Xiaochun Cao, Abbas Khosravi, U~Rajendra Acharya, et~al.
\newblock A review of uncertainty quantification in deep learning: Techniques, applications and challenges.
\newblock \emph{Information Fusion}, 76:\penalty0 243--297, 2021.

\bibitem[Aldemir(2013)]{aldemir2013survey}
Tunc Aldemir.
\newblock A survey of dynamic methodologies for probabilistic safety assessment of nuclear power plants.
\newblock \emph{Annals of Nuclear Energy}, 52:\penalty0 113--124, 2013.

\bibitem[Ayhan and Berens(2018)]{ayhan2018test}
Murat~Seckin Ayhan and Philipp Berens.
\newblock Test-time data augmentation for estimation of heteroscedastic aleatoric uncertainty in deep neural networks.
\newblock In \emph{Medical Imaging with Deep Learning}, 2018.

\bibitem[Bach(2022)]{bach2022information}
Francis Bach.
\newblock Information theory with kernel methods.
\newblock \emph{IEEE Transactions on Information Theory}, 69\penalty0 (2):\penalty0 752--775, 2022.

\bibitem[Bach(2024)]{bach2024sum}
Francis Bach.
\newblock Sum-of-squares relaxations for information theory and variational inference.
\newblock \emph{Foundations of Computational Mathematics}, pages 1--39, 2024.

\bibitem[Banerjee et~al.(2005)Banerjee, Merugu, Dhillon, Ghosh, and Lafferty]{banerjee2005clustering}
Arindam Banerjee, Srujana Merugu, Inderjit~S Dhillon, Joydeep Ghosh, and John Lafferty.
\newblock Clustering with bregman divergences.
\newblock \emph{Journal of machine learning research}, 6\penalty0 (10), 2005.

\bibitem[Bierens(1996)]{bierens1996topics}
Herman~J. Bierens.
\newblock \emph{Topics in Advanced Econometrics: Estimation, Testing, and Specification of Cross-section and Time Series Models}.
\newblock Cambridge University Press, 1996.

\bibitem[Bi{\'n}kowski et~al.(2018)Bi{\'n}kowski, Sutherland, Arbel, and Gretton]{binkowski2018demystifying}
Miko{\l}aj Bi{\'n}kowski, Danica~J Sutherland, Michael Arbel, and Arthur Gretton.
\newblock Demystifying {MMD} {GAN}s.
\newblock In \emph{International Conference on Learning Representations}, 2018.

\bibitem[Bishop and Nasrabadi(2006)]{bishop2006pattern}
Christopher~M. Bishop and Nasser~M. Nasrabadi.
\newblock \emph{Pattern Recognition and Machine Learning}, volume~4.
\newblock Springer, 2006.

\bibitem[Bregman(1967)]{bregman1967relaxation}
L.M. Bregman.
\newblock The relaxation method of finding the common point of convex sets and its application to the solution of problems in convex programming.
\newblock \emph{USSR Computational Mathematics and Mathematical Physics}, 7\penalty0 (3):\penalty0 200 -- 217, 1967.

\bibitem[Brier(1950)]{VERIFICATIONOFFORECASTSEXPRESSEDINTERMSOFPROBABILITY}
Glenn~W. Brier.
\newblock Verification of forecasts expressed in terms of probability.
\newblock \emph{Monthly Weather Review}, 78\penalty0 (1):\penalty0 1 -- 3, 1950.

\bibitem[Bröcker(2009)]{Br_cker_2009}
Jochen Bröcker.
\newblock Reliability, sufficiency, and the decomposition of proper scores.
\newblock \emph{Quarterly Journal of the Royal Meteorological Society}, 135\penalty0 (643):\penalty0 1512–1519, Jul 2009.

\bibitem[Capi{\'n}ski and Kopp(2004)]{capinski2004measure}
Marek Capi{\'n}ski and Peter~Ekkehard Kopp.
\newblock \emph{Measure, Integral and Probability}, volume~14.
\newblock Springer, 2004.

\bibitem[Chang et~al.(2013)Chang, Kruger, Kustra, and Zhang]{chang2013canonical}
Billy Chang, Uwe Kruger, Rafal Kustra, and Junping Zhang.
\newblock Canonical correlation analysis based on hilbert-schmidt independence criterion and centered kernel target alignment.
\newblock In \emph{International Conference on Machine Learning}, pages 316--324, 2013.

\bibitem[Chang et~al.(2024)Chang, Wang, Wang, Wu, Yang, Zhu, Chen, Yi, Wang, Wang, et~al.]{chang2024survey}
Yupeng Chang, Xu~Wang, Jindong Wang, Yuan Wu, Linyi Yang, Kaijie Zhu, Hao Chen, Xiaoyuan Yi, Cunxiang Wang, Yidong Wang, et~al.
\newblock A survey on evaluation of large language models.
\newblock \emph{ACM Transactions on Intelligent Systems and Technology}, 15\penalty0 (3):\penalty0 1--45, 2024.

\bibitem[Chazal et~al.(2024{\natexlab{a}})Chazal, Korba, and Bach]{chazal2024statistical}
Cl{\'e}mentine Chazal, Anna Korba, and Francis Bach.
\newblock Statistical and geometrical properties of regularized kernel kullback-leibler divergence.
\newblock \emph{arXiv preprint arXiv:2408.16543}, 2024{\natexlab{a}}.

\bibitem[Chazal et~al.(2024{\natexlab{b}})Chazal, Korba, and Bach]{chazal2024statisticalgeometricalpropertiesregularized}
Clémentine Chazal, Anna Korba, and Francis Bach.
\newblock Statistical and geometrical properties of regularized kernel kullback-leibler divergence, 2024{\natexlab{b}}.
\newblock URL \url{https://arxiv.org/abs/2408.16543}.

\bibitem[Cohen et~al.(2009)Cohen, Huang, Chen, Benesty, Benesty, Chen, Huang, and Cohen]{cohen2009pearson}
Israel Cohen, Yiteng Huang, Jingdong Chen, Jacob Benesty, Jacob Benesty, Jingdong Chen, Yiteng Huang, and Israel Cohen.
\newblock Pearson correlation coefficient.
\newblock \emph{Noise reduction in speech processing}, pages 1--4, 2009.

\bibitem[Cortes et~al.(2012)Cortes, Mohri, and Rostamizadeh]{cortes2012algorithms}
Corinna Cortes, Mehryar Mohri, and Afshin Rostamizadeh.
\newblock Algorithms for learning kernels based on centered alignment.
\newblock \emph{The Journal of Machine Learning Research}, 13:\penalty0 795--828, 2012.

\bibitem[Dawid(2007)]{dawid2007geometry}
A~Philip Dawid.
\newblock The geometry of proper scoring rules.
\newblock \emph{Annals of the Institute of Statistical Mathematics}, 59\penalty0 (1):\penalty0 77--93, 2007.

\bibitem[Depeweg et~al.(2018)Depeweg, Hernandez-Lobato, Doshi-Velez, and Udluft]{depeweg2018decomposition}
Stefan Depeweg, Jose-Miguel Hernandez-Lobato, Finale Doshi-Velez, and Steffen Udluft.
\newblock Decomposition of uncertainty in bayesian deep learning for efficient and risk-sensitive learning.
\newblock In \emph{International Conference on Machine Learning}, pages 1184--1193, 2018.

\bibitem[Der~Kiureghian and Ditlevsen(2009)]{der2009aleatory}
Armen Der~Kiureghian and Ove Ditlevsen.
\newblock Aleatory or epistemic? does it matter?
\newblock \emph{Structural safety}, 31\penalty0 (2):\penalty0 105--112, 2009.

\bibitem[Detlefsen et~al.(2022)Detlefsen, Borovec, Schock, Jha, Koker, Liello, Stancl, Quan, Grechkin, and Falcon]{Detlefsen2022}
Nicki~Skafte Detlefsen, Jiri Borovec, Justus Schock, Ananya~Harsh Jha, Teddy Koker, Luca~Di Liello, Daniel Stancl, Changsheng Quan, Maxim Grechkin, and William Falcon.
\newblock Torchmetrics - measuring reproducibility in pytorch.
\newblock \emph{Journal of Open Source Software}, 7\penalty0 (70):\penalty0 4101, 2022.

\bibitem[Domingos(2000)]{domingos2000unified}
Pedro Domingos.
\newblock A unified bias-variance decomposition.
\newblock In \emph{International Conference on Machine Learning}, pages 231--238, 2000.

\bibitem[Duchi et~al.(2018)Duchi, Khosravi, and Ruan]{duchi2018multiclass}
John Duchi, Khashayar Khosravi, and Feng Ruan.
\newblock Multiclass classification, information, divergence and surrogate risk.
\newblock \emph{The Annals of Statistics}, 46\penalty0 (6B):\penalty0 3246--3275, 2018.

\bibitem[Eaton(1981)]{eaton1981method}
Morris Eaton.
\newblock A method for evaluating improper prior distributions.
\newblock Technical report, University of Minnesota, 1981.

\bibitem[Efron(1994)]{efron1994introduction}
Bradley Efron.
\newblock \emph{An Introduction to the Bootstrap}.
\newblock CRC press, 1994.

\bibitem[Feng et~al.(2020)Feng, Haase-Sch{\"u}tz, Rosenbaum, Hertlein, Glaeser, Timm, Wiesbeck, and Dietmayer]{feng2020deep}
Di~Feng, Christian Haase-Sch{\"u}tz, Lars Rosenbaum, Heinz Hertlein, Claudius Glaeser, Fabian Timm, Werner Wiesbeck, and Klaus Dietmayer.
\newblock Deep multi-modal object detection and semantic segmentation for autonomous driving: Datasets, methods, and challenges.
\newblock \emph{IEEE Transactions on Intelligent Transportation Systems}, 22\penalty0 (3):\penalty0 1341--1360, 2020.

\bibitem[Friedman(2001)]{friedman2001greedy}
Jerome~H Friedman.
\newblock Greedy function approximation: a gradient boosting machine.
\newblock \emph{Annals of statistics}, pages 1189--1232, 2001.

\bibitem[Frigyik et~al.(2008)Frigyik, Srivastava, and Gupta]{frigyik2008functional}
B{\'e}la~A Frigyik, Santosh Srivastava, and Maya~R Gupta.
\newblock Functional bregman divergence and bayesian estimation of distributions.
\newblock \emph{IEEE Transactions on Information Theory}, 54\penalty0 (11):\penalty0 5130--5139, 2008.

\bibitem[Frydman et~al.(1985)Frydman, Altman, and Kao]{frydman1985introducing}
Halina Frydman, Edward~I. Altman, and Duen-Li Kao.
\newblock Introducing recursive partitioning for financial classification: the case of financial distress.
\newblock \emph{The Journal of Finance}, 40\penalty0 (1):\penalty0 269--291, 1985.

\bibitem[Gal and Ghahramani(2016)]{gal2016dropout}
Yarin Gal and Zoubin Ghahramani.
\newblock Dropout as a {B}ayesian approximation: Representing model uncertainty in deep learning.
\newblock In \emph{International Conference on Machine Learning}, pages 1050--1059, 2016.

\bibitem[Gemini et~al.(2023)Gemini, Anil, Borgeaud, Wu, Alayrac, Yu, Soricut, Schalkwyk, Dai, Hauth, et~al.]{team2023gemini}
Team Gemini, Rohan Anil, Sebastian Borgeaud, Yonghui Wu, Jean-Baptiste Alayrac, Jiahui Yu, Radu Soricut, Johan Schalkwyk, Andrew~M Dai, Anja Hauth, et~al.
\newblock Gemini: a family of highly capable multimodal models.
\newblock \emph{arXiv preprint arXiv:2312.11805}, 2023.

\bibitem[Gneiting and Katzfuss(2014)]{gneiting2014probabilistic}
Tilmann Gneiting and Matthias Katzfuss.
\newblock Probabilistic forecasting.
\newblock \emph{Annual Review of Statistics and Its Application}, 1:\penalty0 125--151, 2014.

\bibitem[Gneiting and Raftery(2005)]{Gneiting2005WeatherFW}
Tilmann Gneiting and Adrian~E. Raftery.
\newblock Weather forecasting with ensemble methods.
\newblock \emph{Science}, 310:\penalty0 248 -- 249, 2005.

\bibitem[Gneiting and Raftery(2007)]{gneitingscores}
Tilmann Gneiting and Adrian~E Raftery.
\newblock Strictly proper scoring rules, prediction, and estimation.
\newblock \emph{Journal of the American Statistical Association}, 102\penalty0 (477):\penalty0 359--378, 2007.

\bibitem[Goodfellow et~al.(2016)Goodfellow, Bengio, and Courville]{goodfellow2016deep}
Ian Goodfellow, Yoshua Bengio, and Aaron Courville.
\newblock \emph{Deep Learning}.
\newblock MIT Press, 2016.

\bibitem[Gretton et~al.(2005)Gretton, Bousquet, Smola, and Sch{\"o}lkopf]{gretton2005measuring}
Arthur Gretton, Olivier Bousquet, Alex Smola, and Bernhard Sch{\"o}lkopf.
\newblock Measuring statistical dependence with hilbert-schmidt norms.
\newblock In \emph{International Conference on Algorithmic Learning Theory}, pages 63--77, 2005.

\bibitem[Gretton et~al.(2012)Gretton, Borgwardt, Rasch, Sch{{\"o}}lkopf, and Smola]{JMLR:v13:gretton12a}
Arthur Gretton, Karsten~M. Borgwardt, Malte~J. Rasch, Bernhard Sch{{\"o}}lkopf, and Alexander Smola.
\newblock A kernel two-sample test.
\newblock \emph{Journal of Machine Learning Research}, 13\penalty0 (25):\penalty0 723--773, 2012.

\bibitem[Gruber and Bach(2024)]{gruber2024optimizingestimatorssquaredcalibration}
Sebastian~G. Gruber and Francis Bach.
\newblock Optimizing estimators of squared calibration errors in classification.
\newblock \emph{arXiv preprint arXiv:2410.07014}, 2024.

\bibitem[Gruber and Buettner(2022)]{gruber2022better}
Sebastian~G. Gruber and Florian Buettner.
\newblock Better uncertainty calibration via proper scores for classification and beyond.
\newblock In \emph{Advances in Neural Information Processing Systems}, 2022.

\bibitem[Gruber and Buettner(2023)]{gruber2023uncertainty}
Sebastian~G. Gruber and Florian Buettner.
\newblock Uncertainty estimates of predictions via a general bias-variance decomposition.
\newblock In \emph{International Conference on Artificial Intelligence and Statistics}, pages 11331--11354, 2023.

\bibitem[Gruber and Buettner(2024)]{gruber2024biasvariancecovariance}
Sebastian~G. Gruber and Florian Buettner.
\newblock A bias-variance-covariance decomposition of kernel scores for generative models.
\newblock In \emph{International Conference on Machine Learning}, pages 16460--16501, 2024.

\bibitem[Gruber et~al.(2024{\natexlab{a}})Gruber, Popordanoska, Tiulpin, Buettner, and Blaschko]{gruber2024consistent}
Sebastian~G. Gruber, Teodora Popordanoska, Aleksei Tiulpin, Florian Buettner, and Matthew~B. Blaschko.
\newblock Consistent and asymptotically unbiased estimation of proper calibration errors.
\newblock In \emph{International Conference on Artificial Intelligence and Statistics}, pages 3466--3474, 2024{\natexlab{a}}.

\bibitem[Gruber et~al.(2024{\natexlab{b}})Gruber, Ziegler, and Buettner]{gruber2024disentanglingmeanembeddingsbetter}
Sebastian~G. Gruber, Pascal~Tobias Ziegler, and Florian Buettner.
\newblock Disentangling mean embeddings for better diagnostics of image generators.
\newblock \emph{arXiv preprint arXiv:2409.01314}, 2024{\natexlab{b}}.

\bibitem[Guo et~al.(2017)Guo, Pleiss, Sun, and Weinberger]{guo2017calibration}
Chuan Guo, Geoff Pleiss, Yu~Sun, and Kilian~Q. Weinberger.
\newblock On calibration of modern neural networks.
\newblock In \emph{International Conference on Machine Learning}, pages 1321--1330, 2017.

\bibitem[Gupta et~al.(2022)Gupta, Smith, Adlam, and Mariet]{gupta2022ensembles}
Neha Gupta, Jamie Smith, Ben Adlam, and Zelda~E Mariet.
\newblock Ensembles of classifiers: a bias-variance perspective.
\newblock \emph{Transactions on Machine Learning Research}, 2022.

\bibitem[Haggenmüller et~al.(2021)Haggenmüller, Maron, Hekler, Utikal, Barata, Barnhill, Beltraminelli, Berking, Betz-Stablein, Blum, Braun, Carr, Combalia, Fernandez-Figueras, Ferrara, Fraitag, French, Gellrich, Ghoreschi, Goebeler, Guitera, Haenssle, Haferkamp, Heinzerling, Heppt, Hilke, Hobelsberger, Krahl, Kutzner, Lallas, Liopyris, Llamas-Velasco, Malvehy, Meier, Müller, Navarini, Navarrete-Dechent, Perasole, Poch, Podlipnik, Requena, Rotemberg, Saggini, Sangueza, Santonja, Schadendorf, Schilling, Schlaak, Schlager, Sergon, Sondermann, Soyer, Starz, Stolz, Vale, Weyers, Zink, Krieghoff-Henning, Kather, {von Kalle}, Lipka, Fröhling, Hauschild, Kittler, and Brinker]{HAGGENMULLER2021202}
Sarah Haggenmüller, Roman~C. Maron, Achim Hekler, Jochen~S. Utikal, Catarina Barata, Raymond~L. Barnhill, Helmut Beltraminelli, Carola Berking, Brigid Betz-Stablein, Andreas Blum, Stephan~A. Braun, Richard Carr, Marc Combalia, Maria-Teresa Fernandez-Figueras, Gerardo Ferrara, Sylvie Fraitag, Lars~E. French, Frank~F. Gellrich, Kamran Ghoreschi, Matthias Goebeler, Pascale Guitera, Holger~A. Haenssle, Sebastian Haferkamp, Lucie Heinzerling, Markus~V. Heppt, Franz~J. Hilke, Sarah Hobelsberger, Dieter Krahl, Heinz Kutzner, Aimilios Lallas, Konstantinos Liopyris, Mar Llamas-Velasco, Josep Malvehy, Friedegund Meier, Cornelia~S.L. Müller, Alexander~A. Navarini, Cristián Navarrete-Dechent, Antonio Perasole, Gabriela Poch, Sebastian Podlipnik, Luis Requena, Veronica~M. Rotemberg, Andrea Saggini, Omar~P. Sangueza, Carlos Santonja, Dirk Schadendorf, Bastian Schilling, Max Schlaak, Justin~G. Schlager, Mildred Sergon, Wiebke Sondermann, H.~Peter Soyer, Hans Starz, Wilhelm Stolz, Esmeralda Vale, Wolfgang Weyers,
  Alexander Zink, Eva Krieghoff-Henning, Jakob~N. Kather, Christof {von Kalle}, Daniel~B. Lipka, Stefan Fröhling, Axel Hauschild, Harald Kittler, and Titus~J. Brinker.
\newblock Skin cancer classification via convolutional neural networks: systematic review of studies involving human experts.
\newblock \emph{European Journal of Cancer}, 156:\penalty0 202--216, 2021.

\bibitem[Hansen and Heskes(2000)]{hansen2000general}
Jakob~Vogdrup Hansen and Tom Heskes.
\newblock General bias/variance decomposition with target independent variance of error functions derived from the exponential family of distributions.
\newblock In \emph{International Conference on Pattern Recognition}, pages 207--210, 2000.

\bibitem[Hastie et~al.(2009)Hastie, Tibshirani, Friedman, and Friedman]{hastie2009elements}
Trevor Hastie, Robert Tibshirani, Jerome~H Friedman, and Jerome~H Friedman.
\newblock \emph{The Elements of Statistical Learning: Data Mining, Inference, and Prediction}, volume~2.
\newblock Springer, 2009.

\bibitem[Hekler et~al.(2023)Hekler, Brinker, and Buettner]{hekler2023test}
Achim Hekler, Titus~J. Brinker, and Florian Buettner.
\newblock Test time augmentation meets post-hoc calibration: uncertainty quantification under real-world conditions.
\newblock In \emph{AAAI Conference on Artificial Intelligence}, pages 14856--14864, 2023.

\bibitem[Hendrickson and Buehler(1971)]{hendrickson1971proper}
Arlo~D Hendrickson and Robert~J Buehler.
\newblock Proper scores for probability forecasters.
\newblock \emph{The Annals of Mathematical Statistics}, 42\penalty0 (6):\penalty0 1916--1921, 1971.

\bibitem[Ho et~al.(2020)Ho, Jain, and Abbeel]{ho2020denoising}
Jonathan Ho, Ajay Jain, and Pieter Abbeel.
\newblock Denoising diffusion probabilistic models.
\newblock \emph{Advances in Neural Information Processing Systems}, 33:\penalty0 6840--6851, 2020.

\bibitem[H{\"u}llermeier and Waegeman(2021)]{hullermeier2021aleatoric}
Eyke H{\"u}llermeier and Willem Waegeman.
\newblock Aleatoric and epistemic uncertainty in machine learning: An introduction to concepts and methods.
\newblock \emph{Machine Learning}, 110\penalty0 (3):\penalty0 457--506, 2021.

\bibitem[Huszar(2013)]{huszar2013scoring}
Ferenc Huszar.
\newblock \emph{Scoring rules, divergences and information in {B}ayesian machine learning}.
\newblock PhD thesis, University of Cambridge, 2013.

\bibitem[Hyv{\"a}rinen and Dayan(2005)]{hyvarinen2005estimation}
Aapo Hyv{\"a}rinen and Peter Dayan.
\newblock Estimation of non-normalized statistical models by score matching.
\newblock \emph{Journal of Machine Learning Research}, 6\penalty0 (4), 2005.

\bibitem[Ito and Johnson(2017)]{ljspeech17}
Keith Ito and Linda Johnson.
\newblock The lj speech dataset.
\newblock \url{https://keithito.com/LJ-Speech-Dataset/}, 2017.

\bibitem[Joshi et~al.(2017)Joshi, Choi, Weld, and Zettlemoyer]{joshi2017triviaqa}
Mandar Joshi, Eunsol Choi, Daniel~S Weld, and Luke Zettlemoyer.
\newblock Triviaqa: A large scale distantly supervised challenge dataset for reading comprehension.
\newblock In \emph{Annual Meeting of the Association for Computational Linguistics (Volume 1: Long Papers)}, pages 1601--1611, 2017.

\bibitem[Kahl et~al.(2024)Kahl, L{\"u}th, Zenk, Maier-Hein, and Jaeger]{kahl2024values}
Kim-Celine Kahl, Carsten~T. L{\"u}th, Maximilian Zenk, Klaus Maier-Hein, and Paul~F Jaeger.
\newblock Val{UES}: A framework for systematic validation of uncertainty estimation in semantic segmentation.
\newblock In \emph{International Conference on Learning Representations}, 2024.

\bibitem[Kasneci et~al.(2023)Kasneci, Se{\ss}ler, K{\"u}chemann, Bannert, Dementieva, Fischer, Gasser, Groh, G{\"u}nnemann, H{\"u}llermeier, et~al.]{kasneci2023chatgpt}
Enkelejda Kasneci, Kathrin Se{\ss}ler, Stefan K{\"u}chemann, Maria Bannert, Daryna Dementieva, Frank Fischer, Urs Gasser, Georg Groh, Stephan G{\"u}nnemann, Eyke H{\"u}llermeier, et~al.
\newblock Chatgpt for good? on opportunities and challenges of large language models for education.
\newblock \emph{Learning and Individual Differences}, 103:\penalty0 102274, 2023.

\bibitem[Kelleher(2019)]{kelleher2019deep}
John~D Kelleher.
\newblock \emph{Deep Learning}.
\newblock MIT Press, 2019.

\bibitem[Kendall and Gal(2017)]{kendall2017uncertainties}
Alex Kendall and Yarin Gal.
\newblock What uncertainties do we need in {B}ayesian deep learning for computer vision?
\newblock \emph{Advances in Neural Information Processing Systems}, 30, 2017.

\bibitem[Kim et~al.(2020)Kim, Kim, Kong, and Yoon]{kim2020glow}
Jaehyeon Kim, Sungwon Kim, Jungil Kong, and Sungroh Yoon.
\newblock Glow-tts: A generative flow for text-to-speech via monotonic alignment search.
\newblock \emph{Advances in Neural Information Processing Systems}, 33:\penalty0 8067--8077, 2020.

\bibitem[Krizhevsky et~al.(2012)Krizhevsky, Sutskever, and Hinton]{krizhevsky2012imagenet}
Alex Krizhevsky, Ilya Sutskever, and Geoffrey~E Hinton.
\newblock Imagenet classification with deep convolutional neural networks.
\newblock \emph{Advances in Neural Information Processing Systems}, 25, 2012.

\bibitem[Kuhn et~al.(2023)Kuhn, Gal, and Farquhar]{kuhn2022semantic}
Lorenz Kuhn, Yarin Gal, and Sebastian Farquhar.
\newblock Semantic uncertainty: Linguistic invariances for uncertainty estimation in natural language generation.
\newblock In \emph{International Conference on Learning Representations}, 2023.

\bibitem[Kull and Flach(2015)]{kull2015novel}
Meelis Kull and Peter Flach.
\newblock Novel decompositions of proper scoring rules for classification: Score adjustment as precursor to calibration.
\newblock In \emph{Machine Learning and Knowledge Discovery in Databases: European Conference, ECML PKDD}, pages 68--85. Springer, 2015.

\bibitem[Kull et~al.(2019)Kull, Perello~Nieto, K{\"a}ngsepp, Silva~Filho, Song, and Flach]{kull2019beyond}
Meelis Kull, Miquel Perello~Nieto, Markus K{\"a}ngsepp, Telmo Silva~Filho, Hao Song, and Peter Flach.
\newblock Beyond temperature scaling: Obtaining well-calibrated multi-class probabilities with dirichlet calibration.
\newblock \emph{Advances in Neural Information Processing Systems}, 32:\penalty0 12316--12326, 2019.

\bibitem[Kumar et~al.(2019)Kumar, Liang, and Ma]{kumar2019verified}
Ananya Kumar, Percy Liang, and Tengyu Ma.
\newblock Verified uncertainty calibration.
\newblock In \emph{Advances on Neural Information Processing Systems}, pages 3792--3803, 2019.

\bibitem[Lakshminarayanan et~al.(2017)Lakshminarayanan, Pritzel, and Blundell]{lakshminarayanan2017simple}
Balaji Lakshminarayanan, Alexander Pritzel, and Charles Blundell.
\newblock Simple and scalable predictive uncertainty estimation using deep ensembles.
\newblock \emph{Advances in Neural Information Processing Systems}, 30, 2017.

\bibitem[Loosli et~al.(2007)Loosli, Canu, and Bottou]{loosli-canu-bottou-2006}
Ga\"{e}lle Loosli, St\'{e}phane Canu, and L\'{e}on Bottou.
\newblock Training invariant support vector machines using selective sampling.
\newblock In \emph{Large Scale Kernel Machines}, pages 301--320. MIT Press, Cambridge, MA., 2007.

\bibitem[MacKay(2003)]{mackay2003information}
David~JC MacKay.
\newblock \emph{Information Theory, Inference and Learning Algorithms}.
\newblock Cambridge University Press, 2003.

\bibitem[Maier-Hein et~al.(2024)Maier-Hein, Reinke, Godau, Tizabi, Buettner, Christodoulou, Glocker, Isensee, Kleesiek, Kozubek, et~al.]{maier2024metrics}
Lena Maier-Hein, Annika Reinke, Patrick Godau, Minu~D Tizabi, Florian Buettner, Evangelia Christodoulou, Ben Glocker, Fabian Isensee, Jens Kleesiek, Michal Kozubek, et~al.
\newblock Metrics reloaded: Recommendations for image analysis validation.
\newblock \emph{Nature methods}, 21\penalty0 (2):\penalty0 195--212, 2024.

\bibitem[McCarthy(1956)]{mccarthy1956measures}
John McCarthy.
\newblock Measures of the value of information.
\newblock \emph{Proceedings of the National Academy of Sciences}, 42\penalty0 (9):\penalty0 654--655, 1956.

\bibitem[Minderer et~al.(2021)Minderer, Djolonga, Romijnders, Hubis, Zhai, Houlsby, Tran, and Lucic]{minderer2021revisiting}
Matthias Minderer, Josip Djolonga, Rob Romijnders, Frances Hubis, Xiaohua Zhai, Neil Houlsby, Dustin Tran, and Mario Lucic.
\newblock Revisiting the calibration of modern neural networks.
\newblock \emph{Advances in Neural Information Processing Systems}, 34, 2021.

\bibitem[Murphy(1973)]{ANewVectorPartitionoftheProbabilityScore}
Allan~H. Murphy.
\newblock A new vector partition of the probability score.
\newblock \emph{Journal of Applied Meteorology and Climatology}, 12\penalty0 (4):\penalty0 595 -- 600, 1973.

\bibitem[Murphy and Winkler(1977)]{10.2307/2346866}
Allan~H. Murphy and Robert~L. Winkler.
\newblock Reliability of subjective probability forecasts of precipitation and temperature.
\newblock \emph{Journal of the Royal Statistical Society. Series C (Applied Statistics)}, 26\penalty0 (1):\penalty0 41--47, 1977.

\bibitem[Naeini et~al.(2015)Naeini, Cooper, and Hauskrecht]{naeini2015}
Mahdi~Pakdaman Naeini, Gregory~F. Cooper, and Milos Hauskrecht.
\newblock Obtaining well calibrated probabilities using {Bayesian} binning.
\newblock In \emph{Proceedings of the {{Twenty}}-{{Ninth AAAI Conference}} on {{Artificial Intelligence}}}, pages 2901--2907, 2015.

\bibitem[OpenAI(2023)]{openai2023gpt4}
OpenAI.
\newblock Gpt-4 technical report, 2023.

\bibitem[Ouimet and Tolosana-Delgado(2022)]{ouimet2022asymptotic}
Fr{\'e}d{\'e}ric Ouimet and Raimon Tolosana-Delgado.
\newblock Asymptotic properties of dirichlet kernel density estimators.
\newblock \emph{Journal of Multivariate Analysis}, 187:\penalty0 104832, 2022.

\bibitem[Ovcharov(2018)]{10.3150/16-BEJ857}
Evgeni~Y. Ovcharov.
\newblock {Proper scoring rules and Bregman divergence}.
\newblock \emph{Bernoulli}, 24\penalty0 (1):\penalty0 53 -- 79, 2018.

\bibitem[Patel et~al.(2021)Patel, Beluch, Yang, Pfeiffer, and Zhang]{patel2021multiclass}
Kanil Patel, William~H. Beluch, Bin Yang, Michael Pfeiffer, and Dan Zhang.
\newblock Multi-class uncertainty calibration via mutual information maximization-based binning.
\newblock In \emph{International Conference on Learning Representations}, 2021.

\bibitem[Pfau(2013)]{pfau2013generalized}
David Pfau.
\newblock A generalized bias-variance decomposition for bregman divergences.
\newblock \emph{Unpublished Manuscript}, 2013.

\bibitem[Popordanoska et~al.(2022)Popordanoska, Sayer, and Blaschko]{popordanoska2022}
Teodora Popordanoska, Raphael Sayer, and Matthew~B. Blaschko.
\newblock A consistent and differentiable {$L_p$} canonical calibration error estimator.
\newblock In \emph{Advances in Neural Information Processing Systems}, 2022.

\bibitem[Reddy et~al.(2019)Reddy, Chen, and Manning]{reddy-etal-2019-coqa}
Siva Reddy, Danqi Chen, and Christopher~D. Manning.
\newblock {C}o{QA}: A conversational question answering challenge.
\newblock \emph{Transactions of the Association for Computational Linguistics}, 7:\penalty0 249--266, 2019.

\bibitem[Ren et~al.(2021)Ren, Luo, and Zhu]{ren2021improving}
Yong Ren, Yucen Luo, and Jun Zhu.
\newblock Improving generative moment matching networks with distribution partition.
\newblock In \emph{AAAI Conference on Artificial Intelligence}, volume~35, pages 9403--9410, 2021.

\bibitem[Rockafellar(1970)]{rockafellar1970convex}
R~Tyrrell Rockafellar.
\newblock \emph{Convex Analysis}, volume~18.
\newblock Princeton University Press, 1970.

\bibitem[Roelofs et~al.(2022)Roelofs, Cain, Shlens, and Mozer]{roelofs2022mitigating}
Rebecca Roelofs, Nicholas Cain, Jonathon Shlens, and Michael~C. Mozer.
\newblock Mitigating bias in calibration error estimation.
\newblock In \emph{International Conference on Artificial Intelligence and Statistics}, pages 4036--4054, 2022.

\bibitem[Savage(1971)]{savage1971elicitation}
Leonard~J Savage.
\newblock Elicitation of personal probabilities and expectations.
\newblock \emph{Journal of the American Statistical Association}, 66\penalty0 (336):\penalty0 783--801, 1971.

\bibitem[Sch{\"o}lkopf(1997)]{scholkopf1997support}
Bernhard Sch{\"o}lkopf.
\newblock \emph{Support vector learning}.
\newblock PhD thesis, Oldenbourg M{\"u}nchen, Germany, 1997.

\bibitem[Sch{\"o}lkopf and Smola(2002)]{scholkopf2002learning}
Bernhard Sch{\"o}lkopf and Alexander~J. Smola.
\newblock \emph{Learning with Kernels: Support Vector Machines, Regularization, Optimization, and Beyond}.
\newblock MIT Press, 2002.

\bibitem[Si et~al.(2009)Si, Tao, and Geng]{si2009bregman}
Si~Si, Dacheng Tao, and Bo~Geng.
\newblock Bregman divergence-based regularization for transfer subspace learning.
\newblock \emph{IEEE Transactions on Knowledge and Data Engineering}, 22\penalty0 (7):\penalty0 929--942, 2009.

\bibitem[Steinbach et~al.(2000)Steinbach, Karypis, and Kumar]{steinbach2000comparison}
Michael Steinbach, George Karypis, and Vipin Kumar.
\newblock A comparison of document clustering techniques.
\newblock Technical report, Department of Computer Science and Engineering, University of Minnesota, 2000.

\bibitem[Steinwart and Ziegel(2021)]{steinwart2021strictly}
Ingo Steinwart and Johanna~F Ziegel.
\newblock Strictly proper kernel scores and characteristic kernels on compact spaces.
\newblock \emph{Applied and Computational Harmonic Analysis}, 51:\penalty0 510--542, 2021.

\bibitem[Touvron et~al.(2023)Touvron, Lavril, Izacard, Martinet, Lachaux, Lacroix, Rozi{\`e}re, Goyal, Hambro, Azhar, et~al.]{touvron2023llama}
Hugo Touvron, Thibaut Lavril, Gautier Izacard, Xavier Martinet, Marie-Anne Lachaux, Timoth{\'e}e Lacroix, Baptiste Rozi{\`e}re, Naman Goyal, Eric Hambro, Faisal Azhar, et~al.
\newblock Llama: Open and efficient foundation language models.
\newblock \emph{arXiv preprint arXiv:2302.13971}, 2023.

\bibitem[Ueda and Nakano(1996)]{ueda1996generalization}
Naonori Ueda and Ryohei Nakano.
\newblock Generalization error of ensemble estimators.
\newblock In \emph{Proceedings of International Conference on Neural Networks (ICNN'96)}, volume~1, pages 90--95. IEEE, 1996.

\bibitem[Vaicenavicius et~al.(2019)Vaicenavicius, Widmann, Andersson, Lindsten, Roll, and Sch{\"o}n]{vaicenavicius2019evaluating}
Juozas Vaicenavicius, David Widmann, Carl Andersson, Fredrik Lindsten, Jacob Roll, and Thomas Sch{\"o}n.
\newblock Evaluating model calibration in classification.
\newblock In \emph{International Conference on Artificial Intelligence and Statistics}, pages 3459--3467, 2019.

\bibitem[Watrous(2018)]{watrous2018theory}
John Watrous.
\newblock \emph{The Theory of Quantum Information}.
\newblock Cambridge University Press, 2018.

\bibitem[Widmann et~al.(2021)Widmann, Lindsten, and Zachariah]{widmann2021calibration}
David Widmann, Fredrik Lindsten, and Dave Zachariah.
\newblock Calibration tests beyond classification.
\newblock In \emph{International Conference on Learning Representations}, 2021.

\bibitem[Williamson(2014)]{williamson2014geometry}
Robert~C Williamson.
\newblock The geometry of losses.
\newblock In \emph{Conference on Learning Theory}, pages 1078--1108, 2014.

\bibitem[Williamson and Cranko(2023)]{williamson2023geometry}
Robert~C Williamson and Zac Cranko.
\newblock The geometry and calculus of losses.
\newblock \emph{Journal of Machine Learning Research}, 24\penalty0 (342):\penalty0 1--72, 2023.

\bibitem[Winkler(1969)]{winkler1969scoring}
Robert~L Winkler.
\newblock Scoring rules and the evaluation of probability assessors.
\newblock \emph{Journal of the American Statistical Association}, 64\penalty0 (327):\penalty0 1073--1078, 1969.

\bibitem[Yurtsever et~al.(2020)Yurtsever, Lambert, Carballo, and Takeda]{yurtsever2020survey}
Ekim Yurtsever, Jacob Lambert, Alexander Carballo, and Kazuya Takeda.
\newblock A survey of autonomous driving: Common practices and emerging technologies.
\newblock \emph{IEEE access}, 8:\penalty0 58443--58469, 2020.

\bibitem[Zalinescu(2002)]{zalinescu2002convex}
Constantin Zalinescu.
\newblock \emph{Convex Analysis in General Vector Spaces}.
\newblock World Scientific, 2002.

\bibitem[Zamo and Naveau(2018)]{zamo2018estimation}
Micha{\"e}l Zamo and Philippe Naveau.
\newblock Estimation of the continuous ranked probability score with limited information and applications to ensemble weather forecasts.
\newblock \emph{Mathematical Geosciences}, 50\penalty0 (2):\penalty0 209--234, 2018.

\bibitem[Zhang et~al.(2022)Zhang, Roller, Goyal, Artetxe, Chen, Chen, Dewan, Diab, Li, Lin, et~al.]{zhang2022opt}
Susan Zhang, Stephen Roller, Naman Goyal, Mikel Artetxe, Moya Chen, Shuohui Chen, Christopher Dewan, Mona Diab, Xian Li, Xi~Victoria Lin, et~al.
\newblock Opt: Open pre-trained transformer language models.
\newblock \emph{arXiv preprint arXiv:2205.01068}, 2022.

\end{thebibliography}
        \bibliographystyle{plainnat}

        \listoffigures
        \listoftables
        
    \addcontentsline{toc}{chapter}{Acknowledgements}
    \newpage
\section*{Acknowledgements}

First and foremost, I would like to provide my deepest gratitude to my supervisor, Prof. Dr. Florian B\"uttner, for this truly amazing time as a PhD student.
When I started my doctoral studies, I was not confident I could contribute scientifically valuable results through the upcoming years of hard work.
I only had a vision of the future and knew I at least needed to try turning it into reality, even at the risk of failure.
Florian's never-ending support, \textbf{optimism}, and trust allowed this thesis to grow in contributions and me to grow as a scientist.
After years of trial and error, I feel proud of the result and delighted with how it represents my original vision.
I don't think this would have been possible without Florian.

However, Florian wasn't the only scientist who supported the quality and quantity of this thesis.
I also want to thank Prof. Dr. David R\"ugamer and Prof. Dr. Matthias Kaschube for acting as thesis examiners, Dr. Paul J\"ager and Prof. Dr. Visvanathan Ramesh for being part of my thesis advisory committee, and the German Cancer Research Center (DKFZ) for providing an excellent environment for doctoral studies.
Further, I appreciate the productive collaborations with Prof. Dr. Francis Bach, Prof. Dr. Matthew Blaschko, Prof. Dr. Aleksei Tiulpin, Theodora Popordanoska, and Pascal Ziegler.

Francis deserves special gratitude for hosting me for three months in Paris, and again, Florian for making this possible.
Francis is a charismatic and funny host, and I am grateful for the scientific exchange during this time. 
Especially Dr. David Holzm\"uller deserves recognition for giving me a great time during and after Paris.
I hope we continue to have inspiring conversations about science in the future.

Last, and most importantly, I want to thank my family and friends for their support throughout the years.
Especially my parents for providing me with an environment without worry and sorrow.

    \includepdf[width=1.25\textwidth, pages=-,  pagecommand={}]{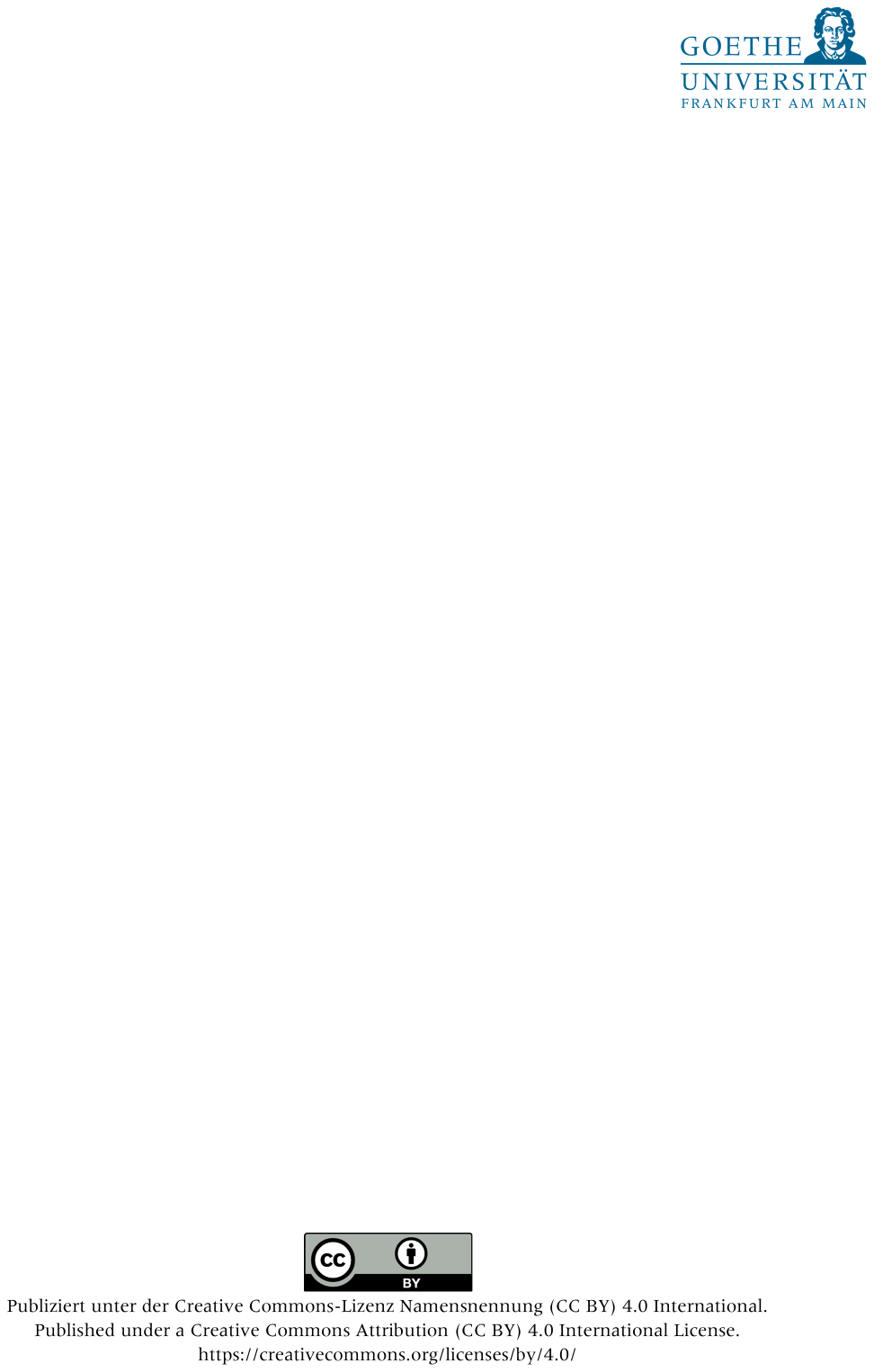}

\end{document}